\documentclass[a4paper,fleqn,longmktitle]{cas-dc}

\usepackage{multicol}

\ExplSyntaxOn
\cs_if_exist:NF \vbox_unpack_clear:N
  {
    \cs_new_protected:Npn \vbox_unpack_clear:N #1
      { \unvbox #1 }
  }

\RenewDocumentCommand \ProcessLongTitleBox { }
{
  \printFirstPageNotes
  \noindent \vbox_unpack_clear:N \g_stm_front_box
  \par \vskip 12pt
  \normalcolor \normalfont
  \begin{multicols}{2}
}
\ExplSyntaxOff

\AtEndDocument{\end{multicols}}

\usepackage[numbers]{natbib}

\usepackage{subcaption}

\usepackage{graphicx}%
\usepackage{multirow}%
\usepackage{amsmath,amssymb,amsfonts}%
\usepackage{amsthm}%
\usepackage{mathrsfs}%
\usepackage[title]{appendix}%
\usepackage{xcolor}%
\usepackage{textcomp}%
\usepackage{manyfoot}%
\usepackage{booktabs}%
\usepackage{algorithm}%
\usepackage{algorithmicx}%
\usepackage{algpseudocode}%
\usepackage{listings}%
\usepackage{rotating} 
\usepackage{caption}
\usepackage{pdfpages}
\usepackage{placeins}

\usepackage[separate-uncertainty=true]{siunitx}
\usepackage{pgf}
\usepackage{adjustbox}

\def\tsc#1{\csdef{#1}{\textsc{\lowercase{#1}}\xspace}}
\tsc{WGM}
\tsc{QE}

\begin{document}
\let\WriteBookmarks\relax
\def\floatpagepagefraction{1}
\def\textpagefraction{.001}

\shorttitle{}    

\shortauthors{}  

\title [mode = title]{OSS: Open Suturing Skills Vision-Based Assessment Challenge 2024-2025}  

\tnotemark[1] 

\tnotetext[1]{} 

\author[1,6]{Hanna Hoffmann}
\cormark[1]
\credit{Conceptualization, Data Curation, Investigation, Methodology, Project Administration, Software, Validation, Formal analysis, Investigation, Writing - Original Draft, Writing - Review \& Editing, Visualization}
\ead{hanna.hoffmann@nct-dresden.de}

\author[3,5]{Setareh Bady}
\credit{Data Curation}

\author[1,6]{Claas de Boer}
\credit{Investigation, Project Administration}

\author[1,7]{Max Kirchner}
\credit{Investigation, Project Administration, Writing - Review \& Editing}

\author[16]{Jan Egger}
\credit{Supervision}

\author[5]{Rainer Röhrig}

\author[3]{Frank Hölzle}

\author[4]{Lennart Johannes Gruber}
\credit{Data Curation}

\author[3]{Kunpeng Xie}
\credit{Data Curation}

\author[3]{Marlon Neuhaus}
\credit{Data Curation}

\author[13]{Victor Alves} 
\credit{Investigation, Methodology, Software, Validation}

\author[13]{Guilherme Barbosa} 
\credit{Investigation, Methodology, Software, Validation}

\author[13]{Leonardo Barroso} 
\credit{Investigation, Methodology, Software, Validation}

\author[13]{João Carvalho} 
\credit{Investigation, Methodology, Software, Validation}

\author[12]{Hao Chen} 
\credit{Investigation, Methodology, Software, Validation}

\author[20]{Gabriella d'Albenzio} 
\credit{Data Curation, Software, Validation, Writing - Original Draft}

\author[13,15,16]{André Ferreira} 
\credit{Investigation, Methodology, Software, Validation}

\author[13]{Nuno Gomes} 
\credit{Investigation, Methodology, Software, Validation}

\author[28]{Yuichiro Hayashi} 
\credit{Investigation, Methodology, Software, Validation}

\author[10]{Kousuke Hirasawa} 
\credit{Investigation, Methodology, Software, Validation}

\author[20]{Rebecca Hisey}
\credit{Investigation, Data Curation, Methodology, Software, Validation, Writing - Original Draft, Supervision}

\author[21]{Seungjae Hong} 
\credit{Methodology, Investigation, Validation, Software}

\author[21,23,24]{Seoi Jeong} 
\credit{Writing - Review \& Editing, Supervision}

\author[13,14,15]{Tiago Jesus} 
\credit{Investigation, Methodology, Software, Validation}

\author[21]{Daehong Kang} 
\credit{Conceptualization, Methodology, Investigation, Validation, Software}

\author[9]{Satoshi Kasai} 
\credit{Investigation, Methodology, Software, Validation}

\author[11,19]{Shunsuke Kikuchi} 
\credit{Investigation, Methodology, Software, Validation}

\author[30]{Takayuki Kitasaka} 
\credit{Investigation, Methodology, Software, Validation}

\author[8]{Satoshi Kondo} 
\credit{Investigation, Methodology, Software, Validation}

\author[21,24,26]{Hyoun-Joong Kong} 
\credit{Writing - Review \& Editing, Supervision, Project Administration, Funding Acquisition}

\author[21,22]{Youngbin Kong} 
\credit{Writing - Review \& Editing, Supervision}

\author[11]{Atsushi Kouno} 
\credit{Investigation, Methodology, Software, Validation}

\author[17]{Shlomi Laufer} 
\credit{Investigation, Methodology, Software, Validation}

\author[24,25]{Kyu Eun Lee} 
\credit{Project Administration, Funding Acquisition}

\author[20]{Bining Long} 
\credit{Data Curation, Methodology, Software, Writing - Original Draft}

\author[20]{Nooshin Maghsoodi} 
\credit{Data Curation, Methodology, Software, Writing - Original Draft}

\author[11]{Hiroki Matsuzaki} 
\credit{Investigation, Methodology, Software, Validation}

\author[18]{Evangelos Mazomenos} 
\credit{Investigation, Methodology, Software, Validation}

\author[17]{Ori Meiraz} 
\credit{Investigation, Methodology, Software, Validation}

\author[28,29,31]{Kensaku Mori} 
\credit{Investigation, Methodology, Software, Validation}

\author[20]{Marina Music} 
\credit{Data Curation, Methodology, Software, Writing - Original Draft}

\author[28,29]{Masahiro Oda} 
\credit{Investigation, Methodology, Software, Validation}

\author[17]{Roi Papo} 
\credit{Investigation, Methodology, Software, Validation}

\author[21,22]{Jieun Park} 
\credit{Conceptualization, Methodology, Investigation, Validation, Software, Writing - Original Draft}

\author[13]{Rafael Piexoto} 
\credit{Investigation, Methodology, Software, Validation}

\author[27]{Saeid Rezaei} 
\credit{Investigation, Methodology, Software, Validation}

\author[13]{Mariana Ribeiro} 
\credit{Investigation, Methodology, Software, Validation}

\author[21,22]{Soyeon Shin} 
\credit{Methodology, Investigation, Validation, Software, Writing - Original Draft}

\author[12]{Yang Shu} 
\credit{Investigation, Methodology, Software, Validation}

\author[17]{Idan Smoller} 
\credit{Investigation, Methodology, Software, Validation}

\author[18]{Danail Stoyanov} 
\credit{Investigation, Methodology, Software, Validation}

\author[12]{Yihui Wang} 
\credit{Investigation, Methodology, Software, Validation}

\author[28]{Xinkai Zhao} 
\credit{Investigation, Methodology, Software, Validation}

\author[1,2]{Sebastian Bodenstedt}
\credit{Conceptualization, Funding acquisition, Resources, Supervision}

\author[1,2,6,7]{Isabel Funke}
\credit{Conceptualization, Formal analysis, Investigation, Methodology, Software, Validation, Supervision, Writing - Review \& Editing}

\author[1,2,6]{Stefanie Speidel}
\fnmark[1]
\credit{Conceptualization, Funding acquisition, Resources, Supervision, Writing - Review \& Editing}

\author[3,5]{Behrus Hinrichs-Puladi}
\fnmark[1]
\credit{Conceptualization, Funding acquisition, Resources, Supervision, Writing - Review \& Editing}

\affiliation[1]{organization={Department of Translational Surgical Oncology, National Center for Tumor Diseases (NCT/UCC) Dresden},
            addressline={}, 
            city={Dresden},
            country={Germany}}

\affiliation[2]{organization={The Centre for Tactile Internet with Human-in-the-Loop (CeTI), TUD Dresden University of Technology},
            addressline={}, 
            city={Dresden},
            country={Germany}}

\affiliation[3]{organization={Department of Oral and Maxillofacial Surgery, University Hospital RWTH Aachen},
            addressline={}, 
            city={Aachen},
            country={Germany}}

\affiliation[4]{organization={Center for Tooth-, Mouth- and Jaw Medicine, University Göttingen},
            addressline={}, 
            city={Göttingen},
            country={Germany}}

\affiliation[5]{organization={Institute of Medical Informatics, University Hospital RWTH Aachen},
            addressline={}, 
            city={Aachen},
            country={Germany}}

\affiliation[6]{organization={Faculty of Medicine and University Hospital Carl Gustav Carus, TUD Dresden University of Technology},
            addressline={}, 
            city={Dresden},
            country={Germany}}

\affiliation[7]{organization={German Cancer Research Center (DKFZ)},
            addressline={}, 
            city={Heidelberg},
            country={Germany}}

\affiliation[8]{organization={Muroran Institute of Technology},
            city={},
            country={Japan}}

\affiliation[9]{organization={Niigata University of Health and Welfare},
            city={},
            country={Japan}}

\affiliation[10]{organization={Konica Minolta, Inc.},
            city={},
            country={Japan}}

\affiliation[11]{organization={Jmees, Inc.},
            city={},
            country={Japan}}

\affiliation[12]{organization={Department of Computer Science and Engineering, The Hong Kong University of Science and Technology},
            city={Hong Kong},
            country={China}}

\affiliation[13]{organization={Center Algoritmi/LASI, University of Minho},
            city={Braga},
            country={Portugal}}

\affiliation[14]{organization={Life and Health Sciences Research Institute (ICVS), School of Medicine, University of Minho},
            city={Braga},
            country={Portugal}}

\affiliation[15]{organization={ICVS/3B's - PT Government Associate Laboratory},
            city={Braga},
            country={Portugal}}

\affiliation[16]{organization={Institute for AI in Medicine (IKIM), University Medicine Essen},
            city={Essen},
            country={Germany}}

\affiliation[17]{organization={The Faculty of Data and Decisions Science, Technion - Israel Institute of Technology},
            city={Haifa},
            country={Israel}}

\affiliation[18]{organization={UCL Hawkes Institute, University College London},
            city={London},
            country={England}}

\affiliation[19]{organization={Computational and Systems Biology Program, UCLA},
            city={Los Angeles},
            country={USA}}

\affiliation[20]{organization={School of Computing, Queen's University},
            city={Kingston},
            country={Canada}}

\affiliation[21]{organization={Department of Transdisciplinary Medicine, Seoul National University Hospital},
            city={Seoul},
            country={South Korea}}

\affiliation[22]{organization={Interdisciplinary Program in Medical Informatics, Seoul National University},
            city={Seoul},
            country={South Korea}}

\affiliation[23]{organization={Department of Clinical Medical Sciences, Seoul National University},
            city={Seoul},
            country={South Korea}}

\affiliation[24]{organization={Institute of Convergence Medicine with Innovative Technology, Seoul National University Hospital},
            city={Seoul},
            country={South Korea}}

\affiliation[25]{organization={Department of Surgery, Seoul National University College of Medicine and Seoul National University Hospital},
            city={Seoul},
            country={South Korea}}

\affiliation[26]{organization={Department of Medicine, Seoul National University College of Medicine},
            city={Seoul},
            country={South Korea}}

\affiliation[27]{organization={University College},
            city={Cork},
            country={Ireland}}

\affiliation[28]{organization={Graduate School of Informatics, Nagoya University},
            city={Nagoya},
            country={Japan}}

\affiliation[29]{organization={Information Technology Center, Nagoya University},
            city={Nagoya},
            country={Japan}}

\affiliation[30]{organization={Department of Information Science, Aichi Institute of Technology},
            city={Nagoya},
            country={Japan}}

\affiliation[31]{organization={Research Center for Medical Bigdata, National Institute of Informatics},
            city={Tokyo},
            country={Japan}}

\fntext[1]{These authors jointly supervised this work.}


\begin{abstract}
Achieving high levels of surgical skill through effective training is essential for optimal patient outcomes. Automated, data-driven skill assessment holds significant potential to improve surgical training. While machine learning-based methods are increasingly popular for assessing skills in minimally invasive surgery, their application to open surgery remains limited. We present the results of a dedicated MICCAI challenge designed to benchmark and advance vision-based skill assessment in open surgery.

The challenge dataset comprises videos of an open suturing training task recorded with a static GoPro camera in a dry-lab setting, with instrument trajectories available in addition to the primary video modality. The OSS Challenge was hosted over two consecutive years, comprising two and three independent tasks, respectively: (1) classifying skill level into four classes, (2) predicting the full Objective Structured Assessment of Technical Skills across eight categories, and (3) tracking hands and surgical tools. Participants submitted diverse solutions including deep learning-based video models, tracking-driven methods, and hybrid approaches.

General-purpose spatiotemporal video models consistently achieved the strongest performance, though conceptually diverse approaches reached competitive levels when well-executed. Predicting fine-grained OSATS scores remains challenging but benefits substantially from increased training data. Keypoint tracking proves difficult given frequent occlusions and out-of-frame instances, limiting current applicability for motion-based skill analysis. This work benchmarks innovative and diverse solutions for surgical skill assessment, highlighting both the promise and current limitations of video-based evaluation in open surgery and identifying critical directions for advancing automated skill assessment toward clinical impact.
\end{abstract}

\begin{graphicalabstract}
\includegraphics[width=\linewidth,keepaspectratio]{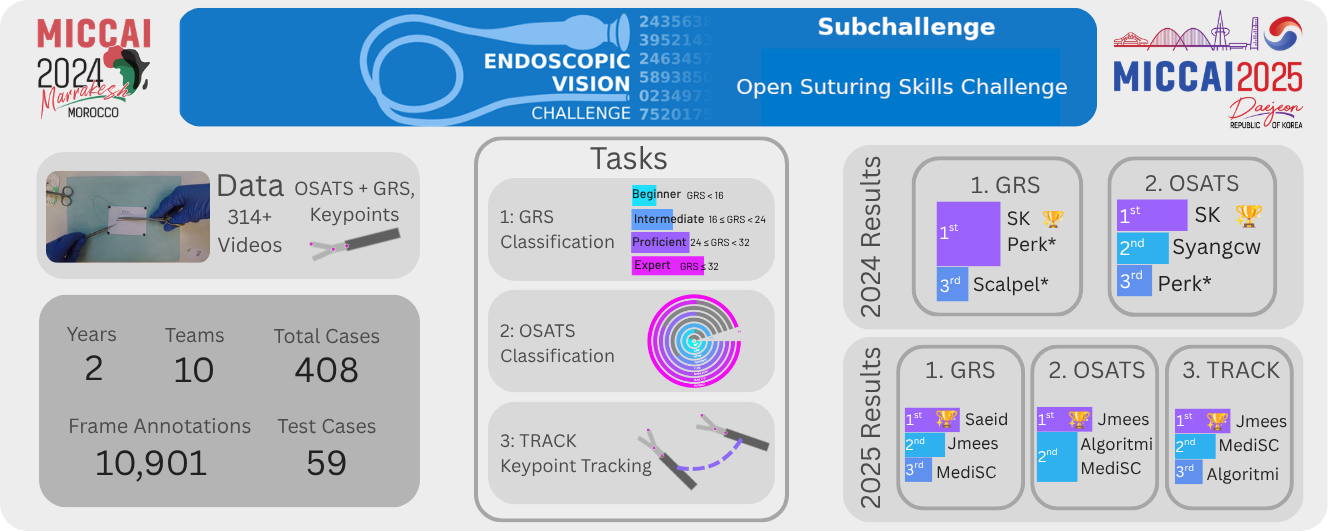}
\end{graphicalabstract}

\begin{highlights}
\item First vision-based skill assessment challenge for open surgery
\item First benchmark of diverse vision-based methods for surgical skill assessment on a dataset
\item Multi-year benchmark comprising classification, OSATS prediction, and keypoint tracking tasks
\item Deep learning video models outperformed tracking-based methods
\item Introduction of keypoint tracking task (2025) for skill assessment
\end{highlights}



\maketitle

\section{Introduction}\label{sec:intro}
Surgical training has undergone significant changes in recent years. Simulation-based practice has become a fundamental component of surgical education~\cite{Elendu2024}, and stronger surgical skills are directly linked to improved patient outcomes through fewer errors, greater procedural efficiency, and better postoperative recovery~\cite{Birkmeyer2013}. Traditionally, surgical education followed an apprenticeship model, with trainees learning through observation and hands-on practice under experienced surgeons~\cite{Seymour2002}. However, this approach suffers from variability in training quality and subjective skill assessment. Standardized rating scales such as OSATS and GEARS have been developed to address this~\cite{Ahmed2011,Goldbraikh2022,Goh2012}, yet they require significantly more time than in-situ feedback. This limitation, combined with the scarcity of surgical mentors, has driven the push for automated skill assessment methods.

Current automated approaches~\cite{Lam2022,Levin2019,Pedrett2023} most commonly use tool motion data to compute trajectory-based features and metrics indicative of skill. Tool motion is either extracted from video using tracking methods~\cite{Fathabadi2021,Lavanchy2021,Lazar2023} or derived from robot kinematics~\cite{Benmansour2023,Ogul2022}. Other methods leverage raw video data directly, using deep learning to extract visual and temporal features for skill prediction~\cite{Anastasiou2023,Hoffmann2024,Kiyasseh2023-tf}.

These state-of-the-art methods have been predominantly developed for robotic and laparoscopic procedures, with limited work addressing open surgery~\cite{Hamza2025}. Kinematics-based approaches are inherently inapplicable to open surgery, and external sensor-based methods (e.g., IMUs, electromagnetic trackers) require additional hardware and calibration that may disrupt surgical workflows. This underscores the need for vision-based models capable of operating unobtrusively in open surgical environments.

To address this gap, the Open Suturing Skills (OSS) Challenge was organized over two consecutive years as part of the EndoVis challenge at MICCAI, establishing a dedicated platform for the development and benchmarking of automated vision-based methods for assessing surgical skill in open surgery suturing tasks. Through a standardized dataset and evaluation framework, the challenge provided the community with a structured opportunity to identify effective strategies for objective skill evaluation, with the ultimate goal of contributing to improved surgical training. Both the target and challenge cohorts consist of surgical trainees performing open surgery suturing tasks in a dry-lab setting, with skill assessment based solely on video data.

The challenge was held at the \emph{International Conference on Medical Image Computing and Computer Assisted Intervention (MICCAI)} in the years 2024 and 2025 as part of the EndoVis challenge series. This paper summarizes the challenge design, dataset, evaluation metrics, and results from the participating teams for both years according to the transparent reporting of the biomedical image analysis (BIAS) challenge guidelines~\cite{MAIERHEIN2020BIAS}.

Contributions:
\begin{itemize}
    \item The design and organization of the MICCAI 2024 and 2025 Open Suturing 
    Skills (OSS) Challenge, the first dedicated challenge for vision-based skill assessment in open surgery.
    \item Establishing a benchmark dataset, tasks, and evaluation 
    framework for vision-based surgical skill assessment.
    \item A comprehensive summary and comparative analysis of the methods submitted 
    by participating teams, including key insights gained from the various approaches.
    \item A unique cross-year comparison between the 2024 and 2025 challenges, 
    examining the evolution of methods, data (baseline improvements for Task 2), 
    and performance.
\end{itemize}

\section{MICCAI 2024 Challenge Overview}\label{sec:challenge2024}
\subsection{Challenge Design}\label{sec:challengedesign24}
\subsubsection*{Organization}
\begin{figure*}[ht]
    \centering
    \includegraphics[width=\linewidth,keepaspectratio]{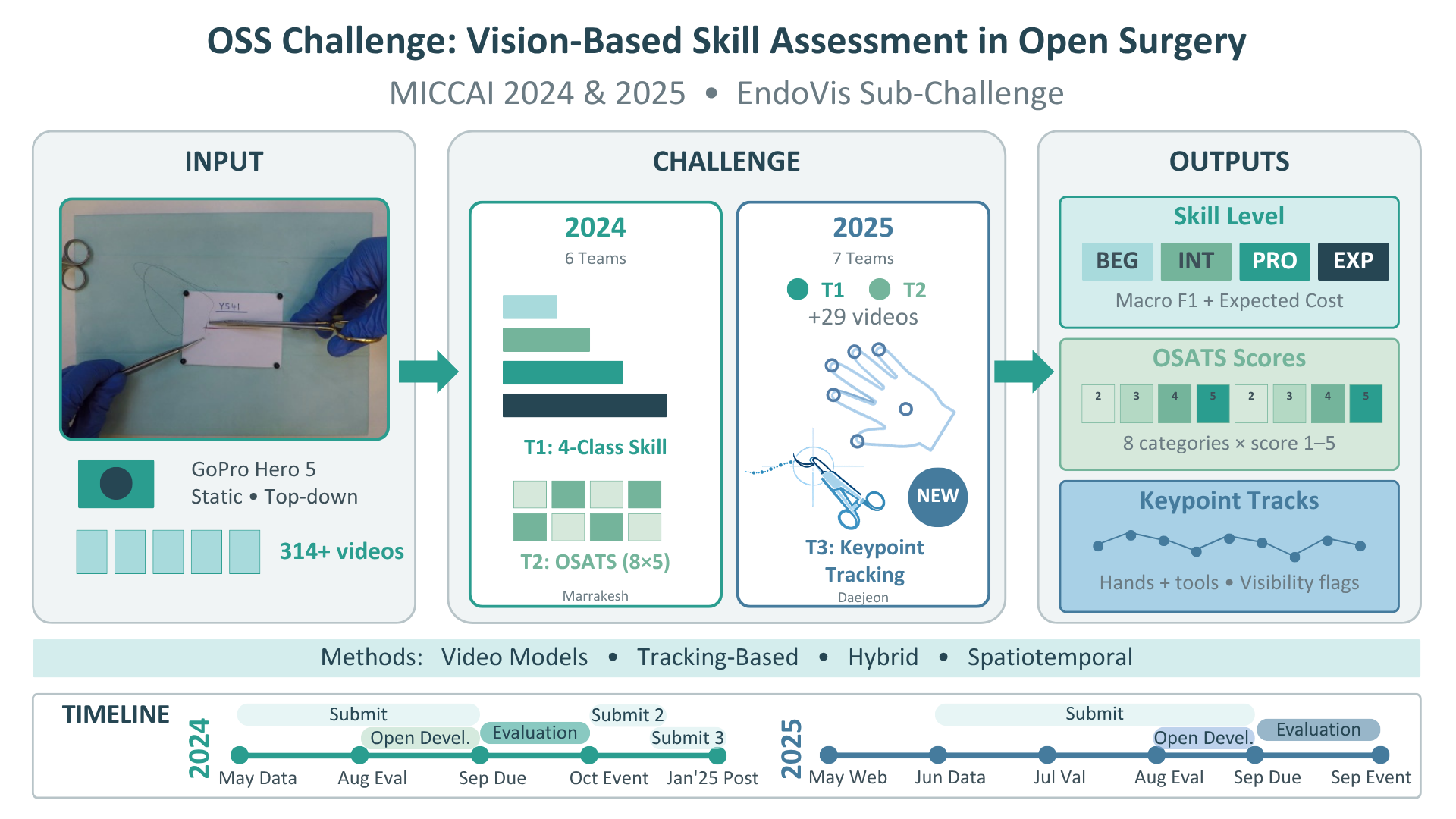}
    \caption{Overview of the 2024 and 2025 MICCAI EndoVis Open Suturing Skills Subchallenges.}
    \label{fig:1}
\end{figure*}

The challenge was hosted at MICCAI 2024 in Marrakesh, Morocco as a subchallenge under the EndoVis challenge\footnote{https://endovis.org/}. It was jointly organized by the Dresden University of Technology (TUD), the National Center for Tumor Diseases (NCT) Dresden, and the University Hospital RWTH Aachen. The NCT and TUD handled challenge evaluation, participant communication, and website management, while RWTH Aachen carried out data collection and annotation. The challenge was hosted on Synapse\footnote{https://www.synapse.org/Synapse:syn54123724/wiki/626561}, where all relevant information was made available. To ensure accountability, participants were required to sign the challenge rules before gaining access to the data.

The challenge followed a structured design\footnote{https://opencas.dkfz.de/endovis/wp-content/uploads/2024/05/39-Endoscopic-Vision-Challenge-2024.pdf} detailing the tasks, evaluation criteria, and the overall framework.  It was divided into two independent tasks, each evaluated separately. An award pool of 1,000 EUR, sponsored by Huawei, was evenly distributed between the two tasks.

The full challenge timeline, as visualized in Figure~\ref{fig:1}, was published on the challenge website. It consisted of three submission rounds. In the first round, teams used only the officially provided challenge data. In the second round, teams could update and resubmit their solutions using the same dataset. In the third round, teams were additionally allowed to use supplementary annotations. For the official challenge, participants could only use the provided dataset along with previously published datasets and pretrained models. Access to the test data was restricted to organizing team members on a need-to-know basis. Members of the organizing institutes could participate but were not eligible for awards. This paper includes all results, evaluations, and analyses from the challenge including the teams' method descriptions as a joint publication.  Participants were not permitted to publish individual challenge results prior to the release of this paper.

Submissions were required to be fully automatic and containerized using Docker. Evaluation was performed on a dedicated machine equipped with up to four NVIDIA RTX A5000 GPUs, two Xeon Silver 4216 CPUs (16 cores each), and 377 GB RAM, though submissions were limited to a single GPU.

\subsubsection*{Mission}
The mission of this challenge is to advance the field of automated surgical skill assessment by focusing on the prediction of surgical skill using the \emph{Objective Structured Assessment of Technical Skills (OSATS)}~\cite{martin1997}. It consists of eight categories, each scored on a 5-point Likert scale, providing a structured assessment of technical ability. The \emph{Global Rating Score (GRS)}, computed as the sum of these individual scores, serves as a holistic measure of overall surgical performance and is frequently used for evaluating training programs and interventions. Predicting the GRS from video data offers the potential to deliver immediate, objective feedback to trainees and educators. By challenging participants to develop vision-based methods for accurate skill prediction, this challenge sought to advance surgical skill assessment in open surgery and contribute to improved training and evaluation practices.

\textbf{Task 1: GRS-based skill classification.}
Task 1 challenges participants to accurately predict the skill level of a given surgical video. The skill level for each video is determined using the annotated GRS score (ranging from 8 to 40) and categorized into four levels: beginner (GRS~$<$~16), intermediate (16~$\leq$~GRS~$<$~24),  proficient (24~$\leq$~GRS~$<$~32), and expert (32~$\leq$~GRS). By framing this as a skill classification task, the challenge builds on existing research while emphasizing real-world relevance. Accurate skill classification from video data serves as an important first step toward providing actionable feedback to both trainees and educators.

\textbf{Task 2: OSATS prediction.}
Task~2 requires a more fine-grained assessment by predicting scores for each individual OSATS category. The OSATS assessment comprises eight categories, each rated on a 1-5 scale~\cite{denadai2014,martin1997}: respect for tissue, time and motion, instrument handling, flow of operation, suture technique, final quality, knowledge of procedure, and overall performance~\cite{Peters2023}. This task encourages the development of methods that provide category-specific feedback, enabling trainees to identify targeted areas for improvement. Accurately predicting these scores from video data represents an important further step toward advancing automated surgical skill evaluation.

\subsection{Dataset}\label{sec:dataset24}
\subsubsection*{Description}
The dataset extends the publicly available AIxSuture dataset\footnote{https://doi.org/10.5281/zenodo.7940583}~\cite{Peters2023-data,Hoffmann2024}. The AIxSuture consists of 314 videos, each approximately 5 minutes long, recorded at 30 fps and totaling around 100 GB. Videos were captured with a GoPro Hero 5 in a standardized, static bird's-eye view and depict students performing open surgery suturing in a dry-lab environment. Each video was rated by three independent raters using the OSATS scale, with annotation guidelines described in detail by~\citet{Peters2023}.

One goal of this challenge was to investigate whether deep-learning models trained on this dataset can reliably recognize expert performance. 
Since few AIxSuture videos show expert-level performance (GRS >\ 32), we collected additional recordings of the open suturing task from 30 experts and 35 students. The experts were surgical dental residents with 5+ years of experience, recorded under the same conditions as the original study but only once. These videos were also rated by three expert raters using OSATS, with one rater overlapping from the original annotation effort. Note that all recordings with GRS >\ 32 are classified as expert in the classification task even if they were performed by students.

Of the additional recordings, 15 expert videos were provided as training data and the remaining 50 were kept as unseen test data. All 314 AIxSuture videos plus the 15 additional videos constituted the training set, uploaded to Synapse with no prescribed train-validation split. However, due to a technical error, the ground truth OSATS scores for the 15 additional training videos were unavailable to participants during the challenge. To mitigate this, teams were allowed to submit an updated solution after the challenge date using the full training dataset including these scores. Results from both original and updated submissions are reported in the results section.

\begin{table*}
\centering
\caption{
Number of videos in the training and test sets of the dataset versions used in the OSS 2024 and 2025 challenges.}
\label{tab:datasets}
\begin{tabular}{llc|c}
\toprule
\textbf{Dataset} & \textbf{Train (Public)} & \textbf{Test (Private)} & \textbf{Total} \\
\midrule
AIxSuture~\cite{Peters2023-data} & 314 & 0 & 314 \\
2024 challenge & 329 (15 unlabeled experts) & 50 & 379 \\
2024 + experts & 329 (with full labels) & 50 & 379 \\
2025 challenge & 349 & 59 & 408 \\
\bottomrule
\end{tabular}
\end{table*}

\begin{figure*}[ht] 
    \centering
    \begin{subfigure}{0.48\textwidth}
        \centering
        \includegraphics[width=\linewidth]{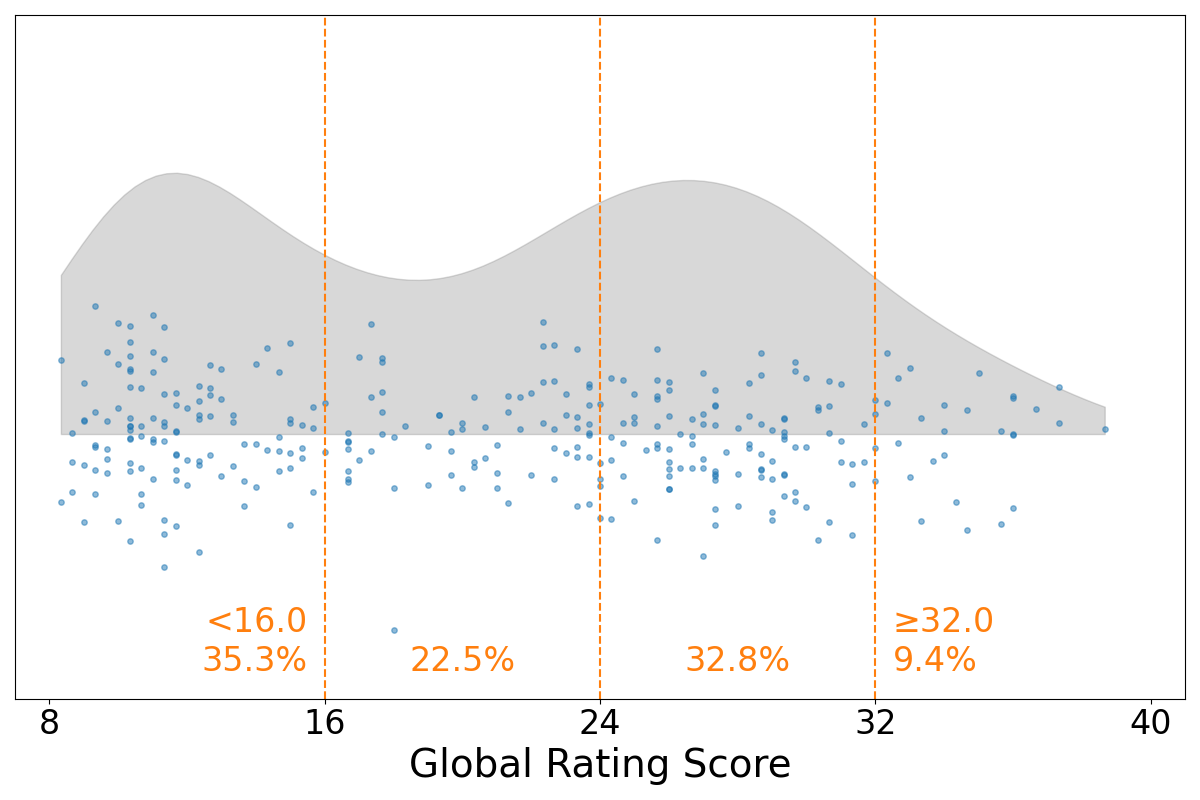}
        \caption{2024 Challenge Training Set}
        \label{fig:datadist_train_24}
    \end{subfigure}
    \hfill
    \begin{subfigure}{0.48\textwidth}
        \centering
        \includegraphics[width=\linewidth]{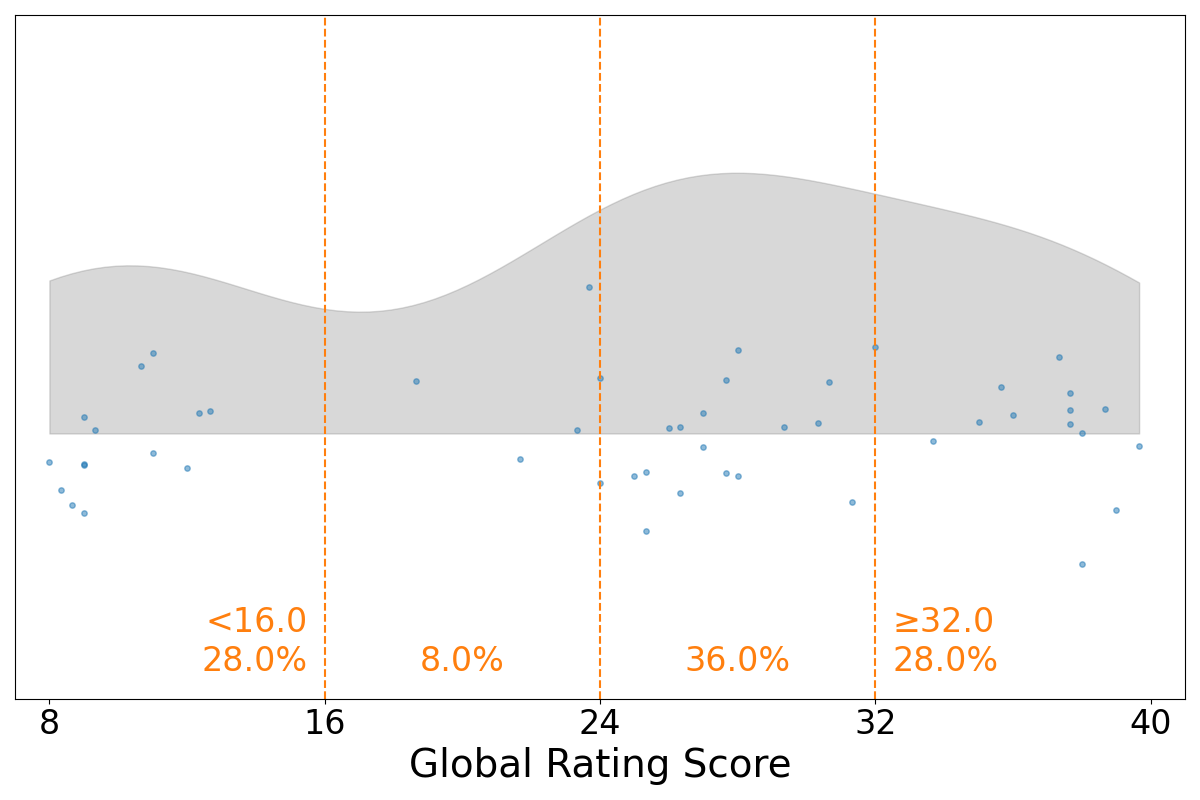}
        \caption{2024 Challenge Test Set}
        \label{fig:datadist_test_24}
    \end{subfigure}
    
    \vspace{0.5cm}
    
    \begin{subfigure}{0.48\textwidth}
        \centering\includegraphics[width=\linewidth]{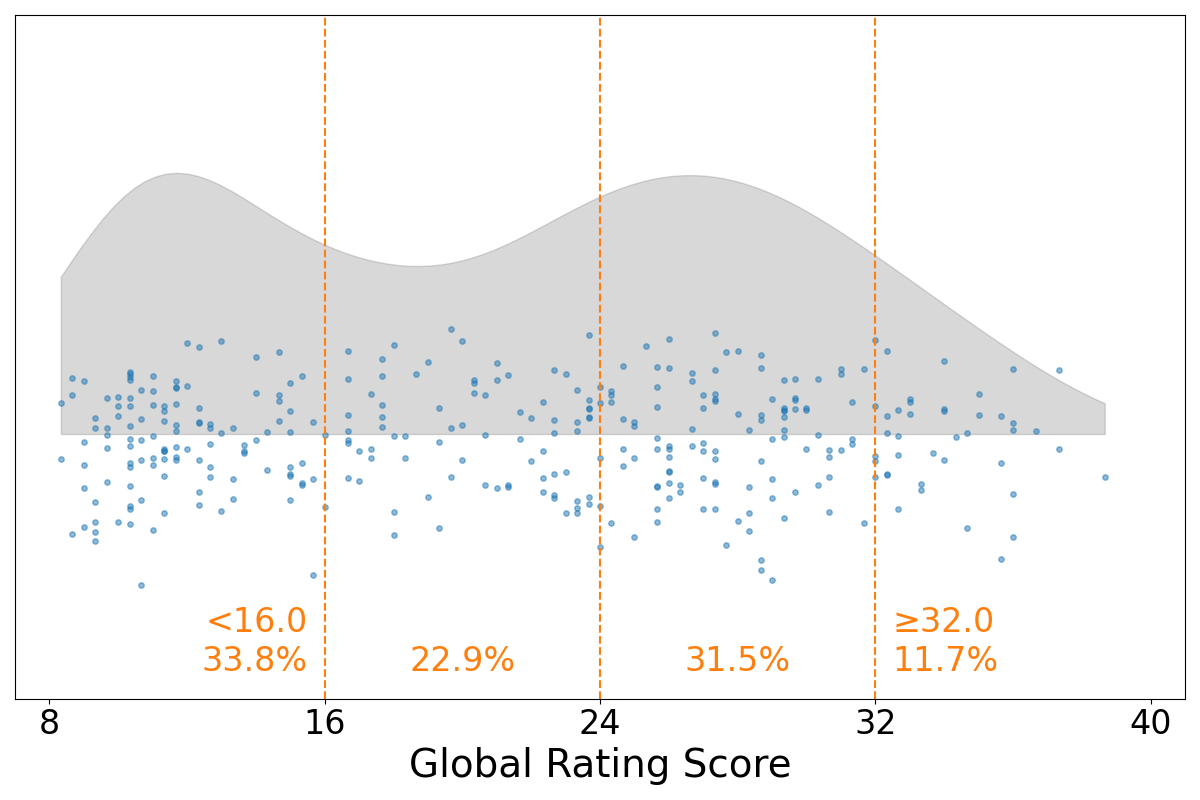}
        \caption{2025 Challenge Training Set}
        \label{fig:datadist_train_25}
    \end{subfigure}
    \hfill
    \begin{subfigure}{0.48\textwidth}
        \centering\includegraphics[width=\linewidth]{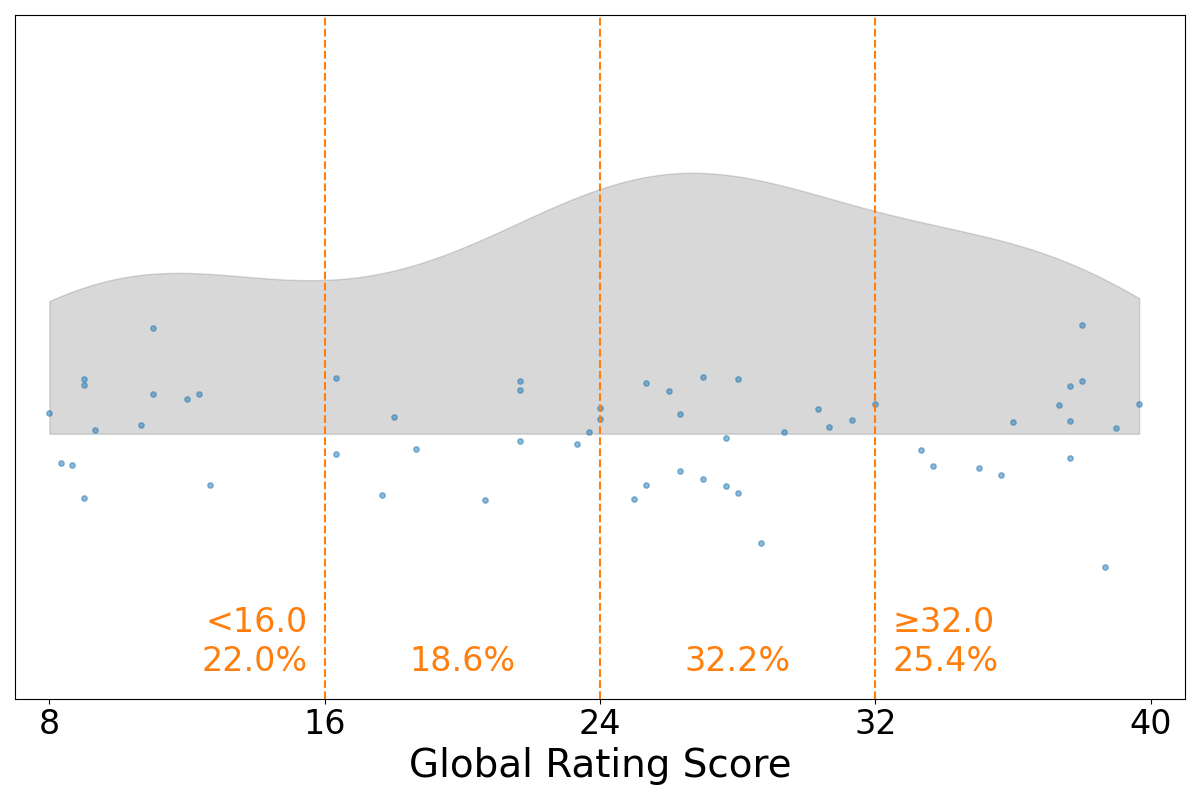}
        \caption{2025 Challenge Test Set}
        \label{fig:datadist_test_25}
    \end{subfigure}
    
    \caption{Distributions of GRS scores (Task 1) for the training and test sets for the 2024 and 2025 challenges. The 2024 data includes the additional expert samples.}
    \label{fig:datadist}
\end{figure*}

A summary of the dataset compositions is provided in Table~\ref{tab:datasets}. Both the training and test sets show elevated counts at the beginner and proficient levels, likely due to the underlying study design~\cite{Peters2023}, in which students were recorded before and after a dedicated teaching session. Expert-level performances are significantly underrepresented in the training set, while the test set contains a higher proportion of proficient and expert videos. This imbalance highlights the challenge of accurately predicting GRS scores across all skill levels, particularly for expert performances. Distributions for Task 2 are provided in the supplementary material.

\subsubsection*{Inter-Rater Agreement Analysis}
The data was annotated by five total raters, where each video was rated by three different raters.

Inter-rater reliability was assessed using the intraclass correlation coefficient (ICC)~\cite{Shrout1979}, defined as “the proportion of a variance that is attributable to the objects of measurement”~\cite{Mcgraw1996}, in this case the video recordings, rather than to measurement error or rater disagreement. ICC values range from 0 to 1, with higher values indicating greater reliability.

Taking into account that videos were assessed by different subsets of the raters, the schema according to Hove~\cite{Hove2024} was adopted, using the equation for absolute agreement of averaged ratings, for an incomplete, balanced two-way design: ICC$(A, k) = \frac{\sigma_\nu^2}{\sigma_\nu^2 + (\sigma_r^2 + \sigma_e^2)/k}$, where $\sigma_\nu^2$ is the variance between videos, $\sigma_r^2$ is the variance between raters, $\sigma_e^2$ is the residual error variance, and $k=3$ is the number of ratings per video. The values were calculated using the R/shiny application ICC4IRR~\cite{ICC4IRR2025}. 

For the GRS, the ICC is 0.92 (95\% CI: 0.89-0.95) for the training set and 0.94 (95\% CI: 0.87-0.99) for the test set, indicating good IRA in both subsets. For the OSATS categories, ICC values ranged from 0.79 to 0.91 for the training set, with the highest reliability observed for \emph{knowledge of specific procedure} (ICC = 0.91) and the lowest for \emph{respect for tissue} (ICC = 0.79). In the test set, the ICC values were generally higher, indicating that the videos in the test set were easier to rate consistently. These results suggest generally good agreement among raters, though certain aspects of surgical skill assessment may be more subjective and prone to variability. The full results of the IRA analysis can be found in the supplementary material.

\subsection{Evaluation Metrics}\label{sec:evaluation24}
On both tasks, the teams were evaluated based on two metrics: macro-averaged F1 score and expected cost (EC). The F1 was selected as it combines both precision and recall and is more meaningful than accuracy given uneven class distribution. Expected cost was chosen to consider the ordinality of classes.

The macro F1 score is calculated as
\begin{equation*}
    \text{F1}_\text{macro} = \frac{1}{C} \sum_{c=1}^C \frac{2\cdot \text{TP}_c}{2\cdot \text{TP}_c + \text{FP}_c + \text{FN}_c}
\end{equation*}
where the F1 score is first computed per class~$c$ and then averaged across all classes.

The F1 score measures absolute prediction correctness, treating all misclassifications equally regardless of their magnitude. However, the classes in this challenge have a natural order (e.g., beginner << intermediate << proficient << expert for Task 1), making the distance between prediction and ground truth relevant. EC was therefore included as a complementary metric to reflect class ordinality. EC generalizes the error rate for situations in which not all errors have equal severity~\cite{Maier-Hein2022-reloaded}.

With linear weights, it is defined as
\begin{equation*}
   \text{EC} = \frac{1}{N}\sum_{i=1}^C\sum_{j=1}^C w_{ij} n_{ij} \text{, with } w_{ij} = \frac{|i-j|}{C-1},
\end{equation*}
where $N$ is the number of samples, $C$ is the number of classes,  $n_{ij}$ is the number of predictions of class $j$ for ground truth class $i$ and $w_{ij}$ is the corresponding error weight. In this form, it behaves similarly to mean absolute error.

Each task was evaluated and ranked independently. Annotations from the three raters were averaged in all cases. For Task 2, metrics were first computed per OSATS category and then averaged across categories. The overall ranking for each task was established by separately ranking models by F1 and EC, then averaging both ranks. In the case of a tie, EC is prioritized. The evaluation code is available publicly online\footnote{{https://gitlab.com/nct\_tso\_public/challenges/miccai2024/snippet}}.

Ranking stability and statistical significance were assessed using bootstrapping~\cite{Efron1994} and the Wilcoxon Signed Rank Test. Bootstrapping is a recommended method for evaluating the ranking variability and performance stability~\cite{maierhein2018ranks}. For bootstrapping, 10,000 samples were drawn with replacement from the test set, and evaluation metrics were calculated for each sample to obtain per-team means and standard deviations. Ranks were computed per bootstrap iteration and metric. Rank stability was assessed by computing the percentage of times each team achieved each rank, as well as the top 5 rank permutations. The Wilcoxon Signed Rank Test was applied as a two-sided pairwise comparison on the bootstrapped metric values to quantify statistical significance of performance differences between teams.

\section{MICCAI 2025 Challenge Overview}\label{sec:challenge2025}
\subsection{Challenge Design}\label{sec:challengedesign25}
The 2025 challenge organization and mission mirror that of the 2024 challenge with the exception of an additional third task. Details specific to the 2025 challenge are described in the following sections.
\subsubsection*{Organization}
The challenge was held at MICCAI 2025 in Daejeon, South Korea. There was an award pool of 450 EUR provided by the 6G Life Project funded by the Federal Ministry of Research, Technology and Space of Germany. The award amount was divided equally among the three tasks. The challenge was hosted on Synapse\footnote{https://www.synapse.org/Synapse:syn66256386/wiki/631724}. Further details to the challenge organization can be found in the challenge design document\footnote{https://zenodo.org/records/15075458}.

The challenge timeline, visualized in Figure~\ref{fig:1}, started with the website launch and ended on with the submission deadline. Written reports could be submitted until September 17th. The challenge concluded and final results were reported at the EndoVis Challenge event at MICCAI 2025 on September 27th.

\subsubsection*{Mission}
The core mission of the challenge as well as that of Tasks 1 and 2 are described in Section~\ref{sec:challengedesign24}. Task 3 focuses on tracking keypoints in the surgical scene. The goal of Task 3 is to compare to objective alternatives for skill assessment, and to connect skill with typical trajectories and motion metrics. In the current vision-based setting, this data would stem from tracking results. However, before these motion descriptors can be connected with skill, existing tracking methods must be verified in their efficacy in tracking key objects in the surgical scene.

The participants are required to track several keypoints across different objects: left and right hand, scissors, needle holder, tweezers, and needle. The aim is to accurately and consistently track these points throughout the video sequence.

\subsection{Dataset}\label{sec:dataset25}
The dataset builds upon the one used in the 2024 edition but differs in two aspects: additional data were provided for Tasks 1 and 2, and further labels were introduced for Task 3.
\subsubsection*{Description}
For Tasks 1 and 2, 20 videos were added to the train set and nine videos were added to the test set, all annotated by three independent raters using OSATS. The distributions can be seen in Figure~\ref{fig:datadist}. In the 2025 edition, additional intermediate and expert videos were included to mitigate class imbalance for Task 1, resulting in a slightly flatter distribution compared to 2024.

For Task 3, a subset of the overall dataset was annotated with keypoints for the following objects: left and right hand, scissors, needle holder, tweezers, and needle. The training set comprises 52 videos annotated at 1 frame per minute (fpm), including segmentation masks for tools and hands. An additional 8 videos were annotated at 1 frame per second (fps) and provided as validation data. Evaluation was performed on 16 videos from the private test set, also annotated at 1 fps. Videos were selected to ensure that the dataset splits are stratified by skill level. In total, 1,816 frames were annotated. Each keypoint was assigned a visibility flag: visible, hidden, or out-of-frame. Annotations were produced by one annotator and reviewed by a second, using LabelBox\footnote{https://app.labelbox.com/}. Further details are provided in the annotation protocol in the supplementary material.

\begin{table*}
\centering
\caption{Challenge 2025 training set keypoint (KP) distributions. The count column indicates the total number of annotated instances of that object. The percentages are calculated as the number of hidden or out-of-frame KPs divided by the total count of KPs.}
\label{tab:datat3_25}
\begin{tabular}{lccc}
\toprule
\textbf{Object} & \textbf{No. of Instances} & \textbf{Hidden KP (\%)} & \textbf{Out-of-Frame KP (\%)} \\
\midrule
Left Hand & 296 & 33.73 & 15.93 \\
Right Hand & 294 & 39.40 & 18.88 \\
Scissors & 301 & 11.85 & 9.86 \\
Tweezers & 299 & 16.16 & 3.79 \\
Needle Holder & 298 & 8.84 & 2.91 \\
Needle & 302 & 29.25 & 5.85 \\
\midrule
Total & 1790 & 26.47 & 11.46 \\
\bottomrule
\end{tabular}
\end{table*}

The distribution of annotated tools is shown in Table~\ref{tab:datat3_25}. Objects are distributed relatively evenly across the dataset. The needle is the most frequently annotated and most frequently hidden object, primarily because it is small and often occluded by fingers or hands. The needle holder is the least frequently hidden and out-of-frame object. The hands have the highest out-of-frame and hidden rates, as they frequently move outside the camera's field of view when pulling thread or manipulating materials, and fingers often occlude each other. Overall, keypoints are hidden approximately 26.5\% and out-of-frame 11.5\% of the time. Validation and test set distributions show similar values with slightly higher out-of-frame rates: 26.0\% hidden and 18.3\% out-of-frame for validation, and 24.1\% hidden and 21.4\% out-of-frame for the test set. Further details are provided in the supplementary material.

\subsubsection*{Inter-Rater Agreement Analysis}
Inter-rater agreement analysis was performed only for the GRS and OSATS scores, as keypoint annotations were produced by a single annotator and reviewed by a second. The ICC was computed as described in Section~\ref{sec:dataset24}. Due to the large overlap in data between the 2024 and 2025 editions, ICC values are similar. The GRS ICC is 0.92 (95\% CI: 0.89–0.95) for the training set and 0.94 (95\% CI: 0.89–0.97) for the test set. For the OSATS categories, training set ICC values ranged from 0.79 to 0.90, again with the highest reliability for \emph{knowledge of specific procedure} and the lowest for \emph{respect for tissue}. The test set showed similarly high ICC values, consistent with the 2024 results. Full IRA results are provided in the supplementary material.

\subsection{Evaluation Metrics}\label{sec:evaluation25}
Tasks 1 and 2 were evaluated using the same metrics as in the 2024 challenge (see Section~\ref{sec:evaluation24}).

Task 3 was evaluated using the Higher Order Tracking Accuracy (HOTA) metric~\cite{Luiten2020}, which is a combination of multiple submetrics. It combines and balances the detection of objects, the localization of objects, and the correct association of objects to tracks. HOTA is computed as follows~\cite{Luiten2020}:
\begin{equation*}
    \text{HOTA} = \int_{0}^{1}{\text{HOTA}_\alpha d\alpha} \approx \frac{1}{19} \sum_{\alpha\in\{0,0.05,\ldots,0.95,1\}} \text{HOTA}_\alpha
\end{equation*}

Here, HOTA$_\alpha$ is calculated as the geometric mean of detection accuracy (DetA$_\alpha$) and association accuracy (AssA$_\alpha$) at a given localization threshold $\alpha$. The threshold $\alpha$ defines the minimum similarity between a prediction and a ground truth item for them to be considered a match. By integrating over multiple thresholds, the localization accuracy is included in the final score.

Originally, this metric was based on bounding box tracking, using IoU as the base similarity metric. Since Task 3 focuses on keypoint tracking, the metric was adjusted to use the euclidean distance as the similarity metric. The source code\footnote{https://github.com/amuck667/TrackEval/tree/devel-kp} is public and was provided to the participants during the developmental period of the challenge.

The same methods were used for statistical evaluation and ranking stability as the 2024 challenge. Please  refer to section~\ref{sec:evaluation24}.

\section{Participation and Results of the 2024 Challenge}\label{sec:participandresults24}
Four teams submitted to the 2024 MICCAI Challenge: Jmees, SK, Syangcw, and Algoritmi. Three of these (Jmees, SK, and Syangcw) submitted solutions for both tasks, while Algoritmi submitted only for Task~1. After the challenge concluded, two further teams (Perk and Scalpel) submitted solutions for both tasks, bringing the total to six participating teams. Teams that had participated in the competitive phase were also able to update and resubmit their solutions during this period. Results from all submissions, including updated ones, are reported in the results section. Full implementation details to teams' methods can be found in the supplementary material.

\subsection{Baseline}
The baseline model provided by the NCT was derived from an established method~\cite{Funke2019,Hoffmann2024}, which combines a spatiotemporal video model (3D CNN) to analyze short clips of a video with a simple strategy to aggregate information over the complete video duration. The video model is pretrained on the Kinetics-400 dataset~\cite{Carreira2017}.

Video frames are extracted from the videos at 5 fps, downsampled, and center cropped. For inference, the video is divided into slightly overlapping 16-frame clips and the video model, X3D-M~\cite{feichtenhofer2020}, is used to compute the feature representation of each clip. Then, the clip-level features are aggregated into a global video-level feature by means of average pooling. Finally, a MLP head computes a skill score based on the global feature.

Three individual models, forming an ensemble, are trained on a Huber loss to predict the GRS score between 8 and 40, using clip-wise data augmentation (color and geometric transformations) and a segment-based sampling strategy~\cite{Wang2016} to randomize the clips picked from each video at training time. The subsequent classification of the predicted GRS to the skill level is hard-coded based on the predefined score ranges. The baseline was trained analogously for the OSATS prediction in Task~2. Here, a separate ensemble was trained for each individual OSATS category.

\subsection{Team SK}
Team SK included Satoshi Konso, Satoshi Kasai, and Kousuke Hirasawa. They proposed using the suture count as a proxy metric for the participant's skill since it correlated strongly with the scores. Notably, the suture count was provided in the ground truth annotations of the challenge data. The team used a ConvNext-Base~\cite{Liu_2022} model which was pre-trained in OpenCLIP~\cite{Cherti2023} on LAION~\cite{Schuhmann2022} and fine-tuned on ImageNet-12k followed by ImageNet-1k~\cite{Russakovskyimagenet15}.

The proposed method predicts the GRS using 16 images from the end of the video, extracted at 1 fps and concatenated into a $3\times16$ channel image after center cropping and downsampling. The number of stitches is also predicted as an auxiliary task. Both predictions are treated as regression problems. During inference, only the GRS results are employed. After which, the GRS is classified into the four categories for task 1. For the second task, the calculated GRS from task 1 is distributed across the eight OSATS categories. This is obtained by weighting the GRS score with a normalized weight gathered from the average score within the respective OSATS category.

\subsection{Team Jmees}
Team Jmees included Shunsuke Kikuchi, Atsushi Kouno, and Horiki Matsuzaki. Their approach was to derive skill-related features by tracking the instrument tips using a segmentation model, Mask2Former~\cite{Cheng2022}. The segmentation model was transferred from the team's submission to the previous SurgToolLoc challenge hosted at MICCAI 2023~\cite{surgtoolloc2023}, which had been trained on the publicly available dataset of that challenge.

The full raw frames were used in training and inference, no cropping strategy was used. Since the segmentation model was trained by other datasets, the frames were also not normalized or augmented for these tasks. 

The overall method operates in three stages: instrument tool tip detection per frame, feature extraction of the tool tip time series data using 1D convolutions and residual blocks combined with a gated recurrent unit (GRU)~\cite{Cho2014} and multi-head attention, and final score regression using LightGBM~\cite{Ke2017}. The GRU is trained to predict OSATS, GRS, and all auxiliary attributes listed in the ground truth file. The final GRS predictions from regression were subsequently scaled to match the four required classes for the GRS task. For the OSATS task, the categories are directly classified by the LightGBM model.

\subsection{Team Algoritmi}
Team Algoritmi consisted of Tiago Jesus, Andr\'e Ferreira, and Victor Alves. The team only submitted a solution for Task 1. They leverage a YOLO model~\cite{Redmon2015} and a multi-layer perceptron (MLP) to build their solution. A YOLOv5 model was used to track hands and was pretrained with public hand tracking data for sign language\footnote{https://github.com/SegwayWarrior/Gesture\_Recognition\_opencv\_yolov5/tree/master/sign\_lang\_detection}.

Each video is subsampled to 3000 frames. Since most videos contain over 9000 frames, the team repeated this for each video at different starting frames, resulting in three sub-videos with 3000 frames of the original. The ratings from the raters were kept unmerged, resulting in a total of 2,250 cases (250 original videos $\times$ 3 sub-videos $\times$ 3 rater labels) for the training dataset. For the validation set, the mean of all raters was used, but each sub-video was considered separately.

Hand features are extracted from the last layer of the pre-trained YOLO model. These features are reduced by Principal Component Analysis (PCA) to a size of $(3000, 3000)$. An MLP consisting of one layer rates each video based on the PCA-reduced YOLO features on a continuous scale which is subsequently categorized into fixed class bins. Only the MLP model was trained; the YOLO model used the pre-trained weights.

\subsection{Team Scalpel}
Team Scalpel consisted of Roi Papo, Idan Smoller, Ori Meiraz, Evangelos Mazomenos, Shlomi Laufer, and Danail Stoyanov. The team employs a combination of temporal modeling and feature extraction techniques to analyze surgical videos.

For OSATS prediction, they use a pipeline that integrates frame-level features extracted from a YOLOv11 model~\cite{Redmon2015} with global video-level statistics, such as idle time and average speed, gathered from the same model. Furthermore, a specialized model trained in suture detection was employed to gain global stitch descriptors (number, size, spacing of sutures). 

The YOLO model is fine-tuned using the RoHan technique~\cite{rohan2025} to adapt to the surgical domain, and its penultimate layer embeddings are used to represent each video frame. For this, the team manually labeled bounding boxes around both tools and hands (right hand, left hand, forceps, needle driver, scissors and simulator) for a sample of 500 frames. These YOLO embeddings are processed through a multi-headed attention mechanism~\cite{Vaswani2017} with positional encoding to capture temporal context, which is then combined with global features for final predictions. The model employs multi-task learning with three prediction heads to enhance performance.
For GRS prediction, the authors use a LightGBM-based~\cite{Ke2017} pipeline that relies on the global descriptors, optimized through hyperparameter tuning and balanced resampling to address class imbalance. The pipeline includes 4-fold cross-validation and early stopping to ensure robust performance.

\subsection{Team Perk}
Team Perk included Rebecca Hisey, Nooshin Maghsoodi, Gabriella d'Albenzio, Marina Music, and Bining Long. They sought to integrate metrics derived from tool and hand motion with the procedural understanding provided by workflow recognition.

To train models for object detection and workflow recognition, the team annotated a substantial subset of the data with workflow phases and bounding box labels for instruments and hands. Videos were downsampled to 10 fps and images were resized. The core method consisted of instrument and hand tracking, workflow recognition, metric calculation based on the previous two steps, and score prediction using an ensemble of shallow machine-learning models (support vector machine, MLP, and logistic regression).

To track the motion of hands and instruments, they used a YOLOv8~\cite{Redmon2015}. For workflow recognition, the team utilized an approach consisting of a ResNet50~\cite{He2016} as a frame-wise feature extractor and a TeCNO~\cite{Czempiel2020} Temporal Convolutional Network (TCN) as video-level temporal model. One phase label was predicted for each frame in the video. The CNN and TCN were trained independently on the phase labels in two subsequent steps, where the TCN received frame-wise YOLO embeddings as inputs in addition to the CNN features.

Based on the instrument and hand tracking, and workflow recognition results, the team calculated a variety of descriptive features for each video, including the number of completed sutures, number of knots, and average time per stitch. From these features, the most relevant ones were identified using a Spearman correlation and further reduced by a PCA. With the selected features, the ensemble of shallow machine-learning models was trained to predict the score for each scoring criteria, which is obtained by averaging over all models in the ensemble. Specifically, one ensemble is trained per scoring criteria.

\subsection{Team Syangcw}
Team Syangcw consisted of Yang Shu, Yihui Wang, and Hao Chen. They employ a spatiotemporal video transformer model, Surgformer~\cite{Yang2024}, which extends the TimeSFormer~\cite{Bertasius2021} model with hierarchical temporal attention (HTA) and aggregated spatial attention (ASA). Initially, Surgformer was developed for online phase recognition and predicts the phase label of the final frame of a video clip. The SurgFormer model was pretrained on Kinetics-400~\cite{Carreira2017} by TimeSFormer~\cite{Bertasius2021} with the remaining layers randomly initialized.

The data is resampled to 1 fps and the shorter side is resized to 360 pixels. During training, random scaling and cropping are applied to the data to obtain $224\times224$ input images. Starting from a random start frame, a total number of 48 frames are sampled by selecting every fourth frame to create a video clip, corresponding to approximately three minutes of video, based on which the skill score of the overall video is predicted. The video clip is first converted into spatial-temporal tokens with an additional class token appended and passed to sequentially stacked transformer blocks. Then, the skill score is predicted based on the class token. During inference, six video clips are sampled from each video. After computing the score for each clip, the outcomes are ranked and the mode of the scores is selected as the final prediction.

\subsection{Scores and Ranks}\label{sec:results24}
This challenge paper includes results from the challenge hosted at MICCAI 2024 as well as its continuation beyond this date. The teams were ranked according to best model performance in each task. As the tasks were evaluated and ranked separately, each had its own winner.

\subsubsection{Task 1: GRS-based Classification}
The final ranking of the official MICCAI 2024 Challenge was from first to last: SK, Syangcw, Jmees, and Algoritmi. The full challenge evaluation results for this task can be found in Table~\ref{tab:eval24}. This table also includes the results from the teams submitting after the challenge deadline. Out of all participants, Team SK achieved the highest F1 score of 0.55 during the original challenge, while Team Perk attained an F1 score of 0.62 in the post-challenge submissions, marking the best performance over all participants. Both teams also achieved superior EC scores in their respective submission phases in comparison to the other teams. Both Perk and Scalpel had a better F1 score than Jmees, indicating that they were able to predict the correct class more often, but their EC scores were still higher than SK, suggesting that their predictions were less accurate in terms of ordering. However, no submission was able to beat the baseline. Further details on the performance of each team can be found in the supplementary material.

\begin{table*}[ht]
\centering
\caption{2024 Challenge Results. Baseline and best results per submission phase, task, and metric are marked in bold. Teams marked with an asterisk added own labels to the challenge dataset.}
\label{tab:eval24}
\begin{tabular}{l|cc|cc}
\toprule
 & \multicolumn{2}{c|}{\textit{Task 1: GRS}} & \multicolumn{2}{c}{\textit{Task 2: OSATS}} \\
\textbf{Team} & \textbf{F1 $\uparrow$} & \textbf{EC $\downarrow$} & \textbf{F1 $\uparrow$} & \textbf{EC $\downarrow$} \\
\midrule
\textbf{Baseline} & \textbf{0.65} & \textbf{0.11} & \textbf{0.39} & \textbf{0.16} \\
\multicolumn{5}{c}{\textit{Original Challenge (MICCAI 2024)}} \\
\midrule
SK & \textbf{0.55} & \textbf{0.13} & \textbf{0.38} & \textbf{0.16} \\
Syangcw & 0.36 & 0.35 & 0.31 & 0.16 \\
Jmees & 0.20 & 0.35 & 0.21 & 0.26 \\
Algoritmi & 0.16 & 0.46 & - & - \\
\midrule
\multicolumn{5}{c}{\textit{Post-Challenge Submissions}} \\
\midrule
Perk* & \textbf{0.62} & \textbf{0.15} & \textbf{0.37} & \textbf{0.17} \\
Scalpel* & 0.38 & 0.24 & 0.31 & 0.22 \\
\bottomrule
\end{tabular}
\end{table*}

\begin{figure*}[ht]
    \centering
    \begin{subfigure}{0.48\textwidth}
        \centering
        \includegraphics[width=\linewidth]{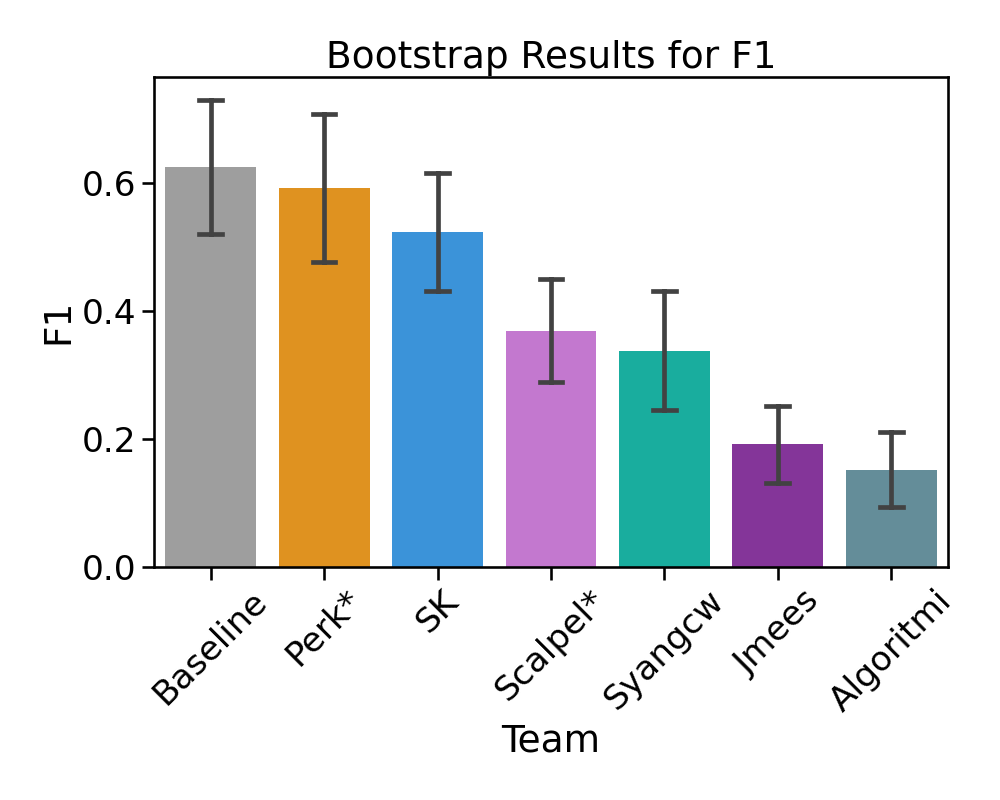}
        \label{fig:t1boot_f1_24}
    \end{subfigure}
    \hfill
    \begin{subfigure}{0.48\textwidth}
        \centering
        \includegraphics[width=\linewidth]{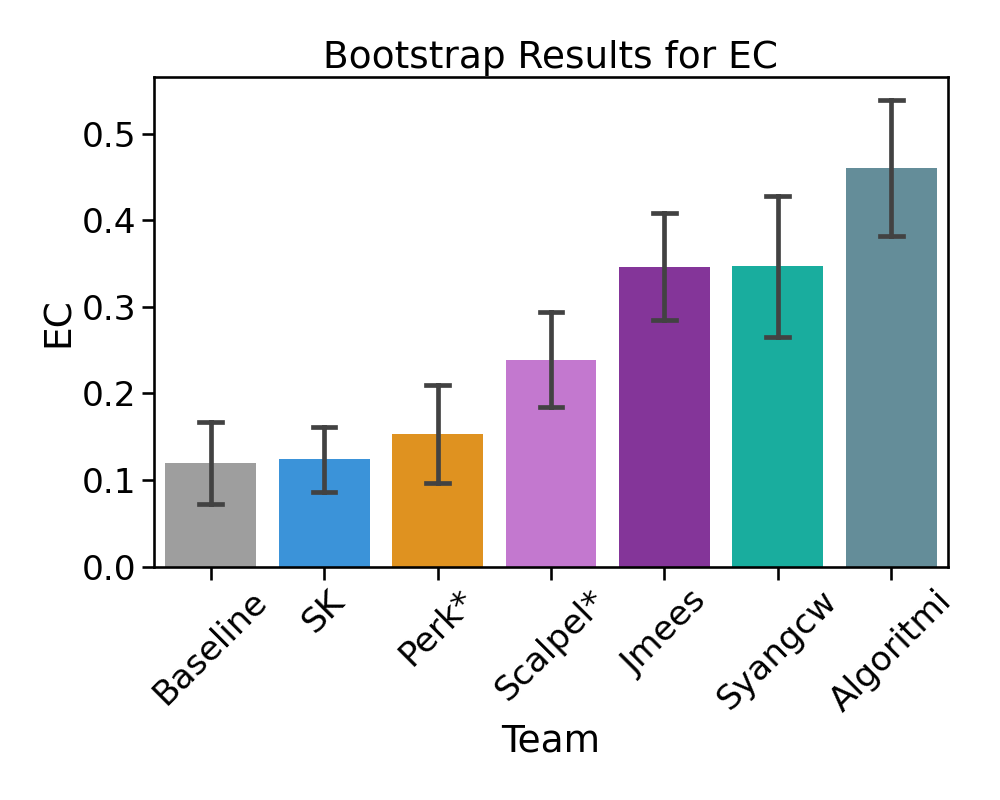}
        \label{fig:t1boot_ec_24}
    \end{subfigure}
    
    \vspace{0.5cm}
    
    \begin{subfigure}{0.48\textwidth}
        \centering\includegraphics[width=\linewidth]{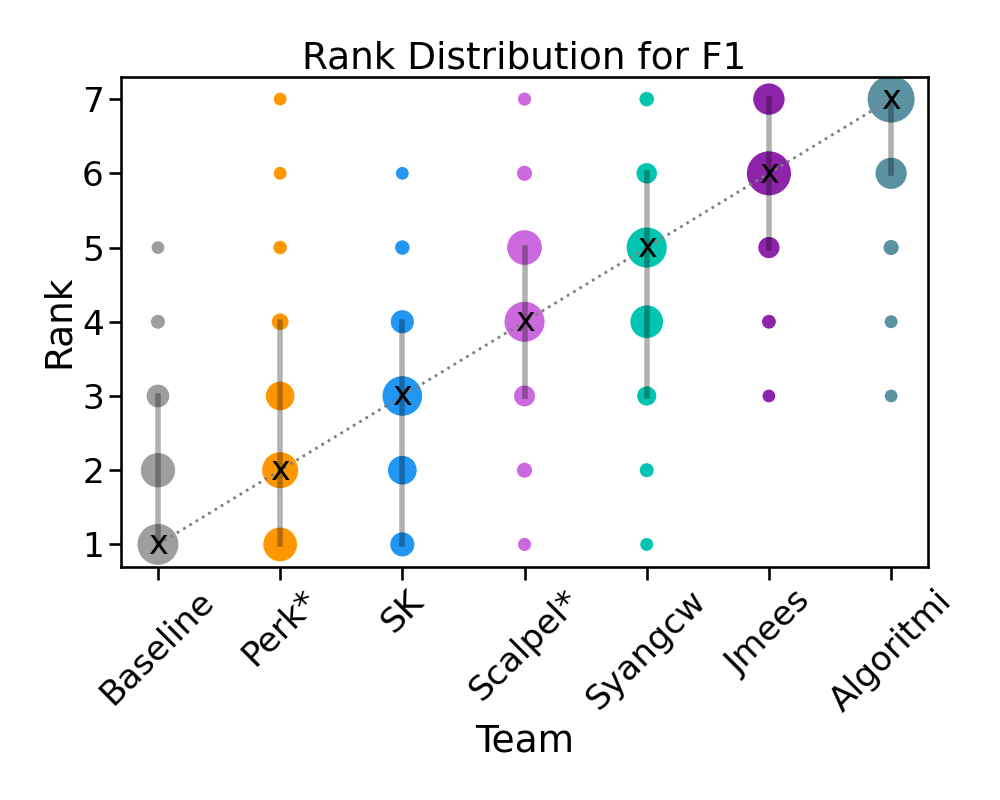}
        \label{fig:t1ranks_f1_24}
    \end{subfigure}
    \hfill
    \begin{subfigure}{0.48\textwidth}
        \centering\includegraphics[width=\linewidth]{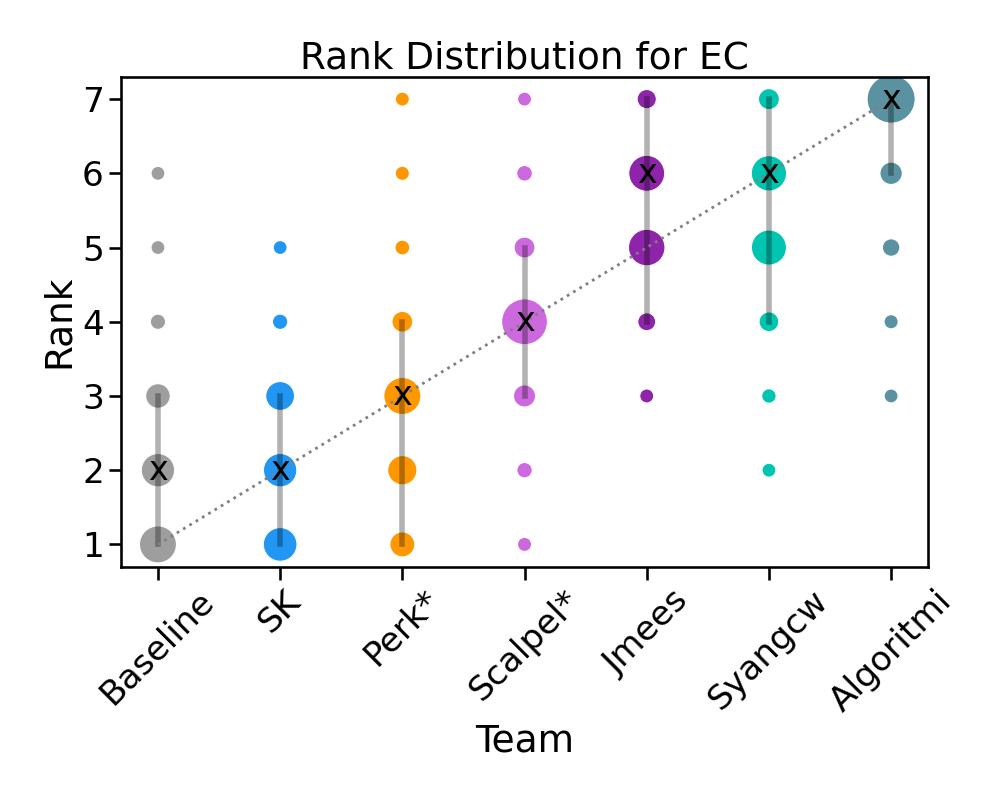}
        \label{fig:t1ranks_ec_24}
    \end{subfigure}

    \caption{Task 1: GRS Metric performance (top row) and rank analysis (bottom row) on the challenge dataset for F1 and Expected Cost (EC) after bootstrapping with 10,000 repetitions. Error bars denote the standard deviation for the performance graphs. Rank plot bubble size corresponds to rank frequency, solid lines are 95\% confidence intervals, and crosses denote the median rank of that team.  Team ranks are sorted by mode.}
    \label{fig:t1stats24}
\end{figure*}

Bootstrap and rank analysis shown in Figure~\ref{fig:t1stats24} support results. The bootstrap and  rank distribution analysis reveals that the baseline model consistently outperforms all other submissions, with a clear margin in both F1 score and EC.  Both analyses support the initial challenge ranking as well as the ranking including the post-challenge submissions by mirroring the ranking results. All pairwise comparisons for both metrics yielded  $p<0.001$, confirming statistically distinct performance tiers between teams, with the sole exception of Jmees and Syangcw's EC ($p=0.6$), whose close performance is clearly visible in the bootstrap and rank graphs.

When adding the additional expert data, performance slightly increases for most teams as shown in Table~\ref{tab:updateeval24}. Team SK again achieves the best F1 score of 0.75 and the best EC of 0.07 after retraining, outperforming all other teams as well as the baseline. Team Perk also shows improved performance with an F1 score of 0.69 and an EC of 0.14, maintaining its position as the second-best performer among all teams. Team Jmees and Scalpel also benefit from the additional data, indicating a moderate improvement. Only team Algoritmi's results remained unchanged.

\begin{table*}[ht]
\centering
\caption{2024 Results with additional expert data for training. Baseline and best results per task and metric are marked in bold. Teams marked with an asterisk added own labels to the challenge dataset.}
\label{tab:updateeval24}
\begin{tabular}{l|cc|cc}
\toprule
 & \multicolumn{2}{c|}{\textit{Task 1: GRS}} & \multicolumn{2}{c}{\textit{Task 2: OSATS}} \\
\textbf{Team} & \textbf{F1 $\uparrow$} & \textbf{EC $\downarrow$} & \textbf{F1 $\uparrow$} & \textbf{EC $\downarrow$} \\
\midrule
\textbf{Baseline} & \textbf{0.67} & \textbf{0.11} & \textbf{0.44} & \textbf{0.14} \\
\midrule
SK & \textbf{0.75} & \textbf{0.07} & 0.39 & \textbf{0.15} \\
Jmees & 0.36 & 0.27 & 0.24 & 0.27 \\
Algoritmi & 0.16 & 0.46 & - & - \\
Perk* & 0.69 & 0.14 & \textbf{0.49} & 0.16 \\
Scalpel* & 0.54 & 0.15 & 0.38 & 0.22 \\
\bottomrule
\end{tabular}
\end{table*}

\subsubsection{Task 2: OSATS Prediction}
Participating teams for this task included SK, Syangcw, and Jmees for the original challenge phase, and Perk and Scalpel joining after the challenge event. Results are shown in the last column pair of Table~\ref{tab:eval24}.

Similarly to Task 1, team SK achieved the highest F1 score of 0.38 during the original challenge, while Team Perk attained an F1 score of 0.37 in the post-challenge submissions. Both teams also achieved superior EC scores in their respective submission phases in comparison to the other teams. However, no submission was able to beat the baseline. In general, performance was more similar than for Task 1. F1 performance was very marginal for this task while EC did not drop as significantly. This suggests that while the methods were able to capture approximate relative ordering of test set samples, they were not able to do so with high precision or recall.

\begin{figure*}[ht]
    \centering
    \begin{subfigure}{0.48\textwidth}
        \centering
        \includegraphics[width=\linewidth]{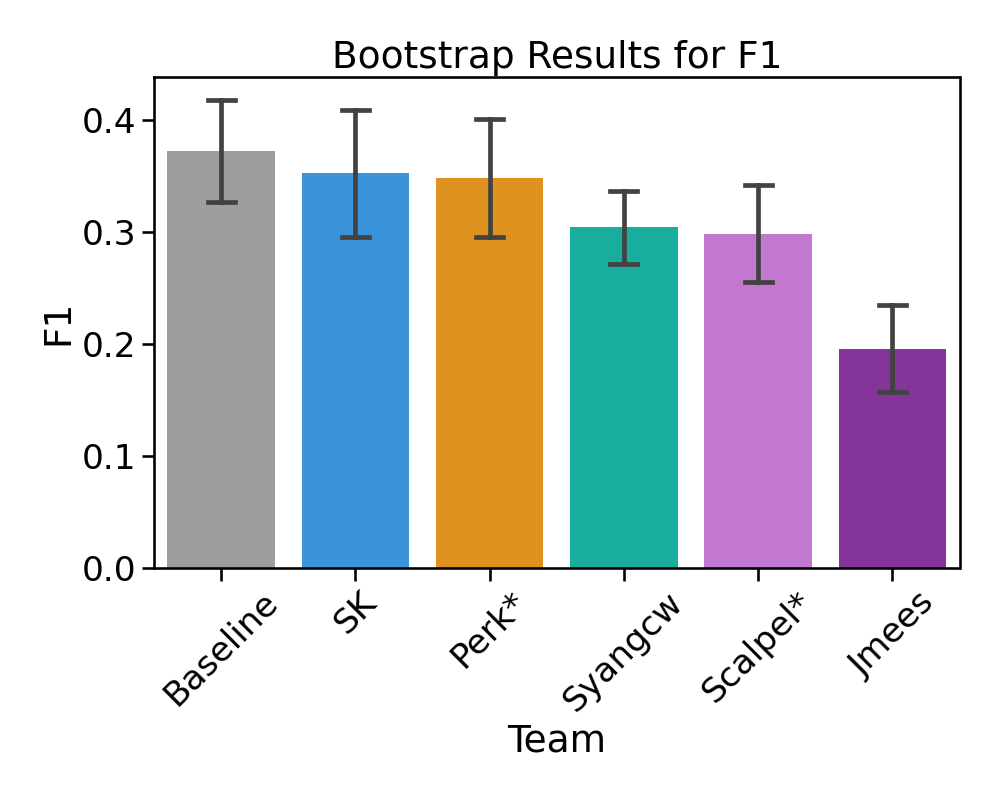}
        \label{fig:t2boot_f1_24}
    \end{subfigure}
    \hfill
    \begin{subfigure}{0.48\textwidth}
        \centering
        \includegraphics[width=\linewidth]{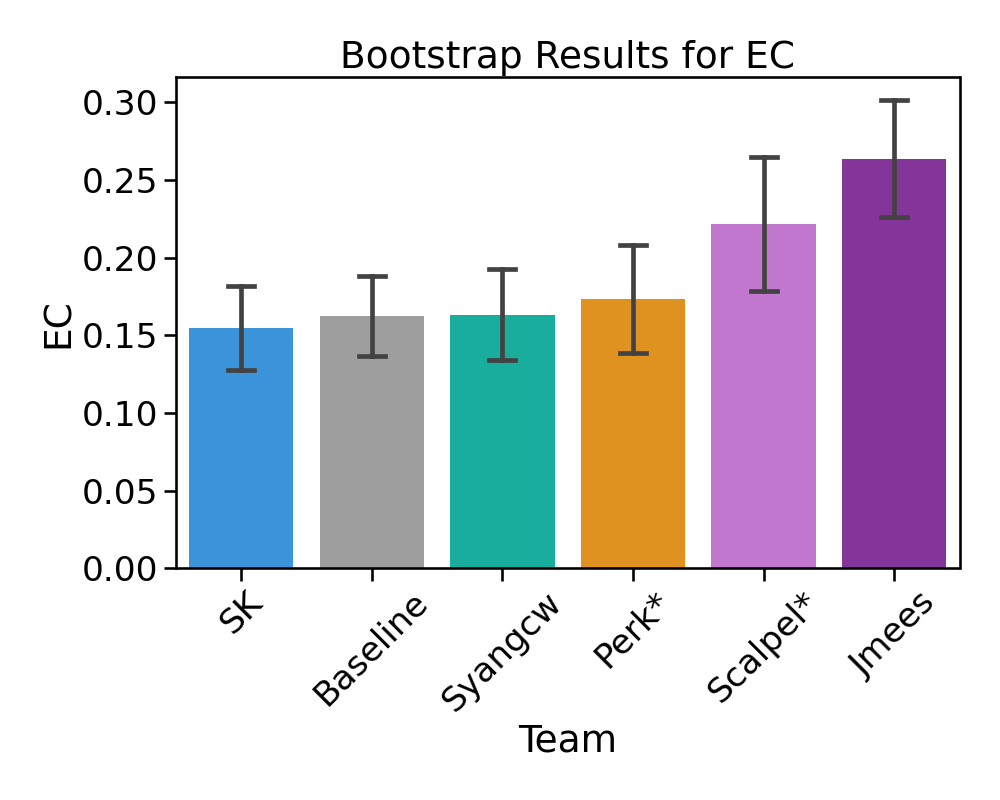}
        \label{fig:t2boot_ec_24}
    \end{subfigure}
    
    \vspace{0.5cm}
    
    \begin{subfigure}{0.48\textwidth}
        \centering\includegraphics[width=\linewidth]{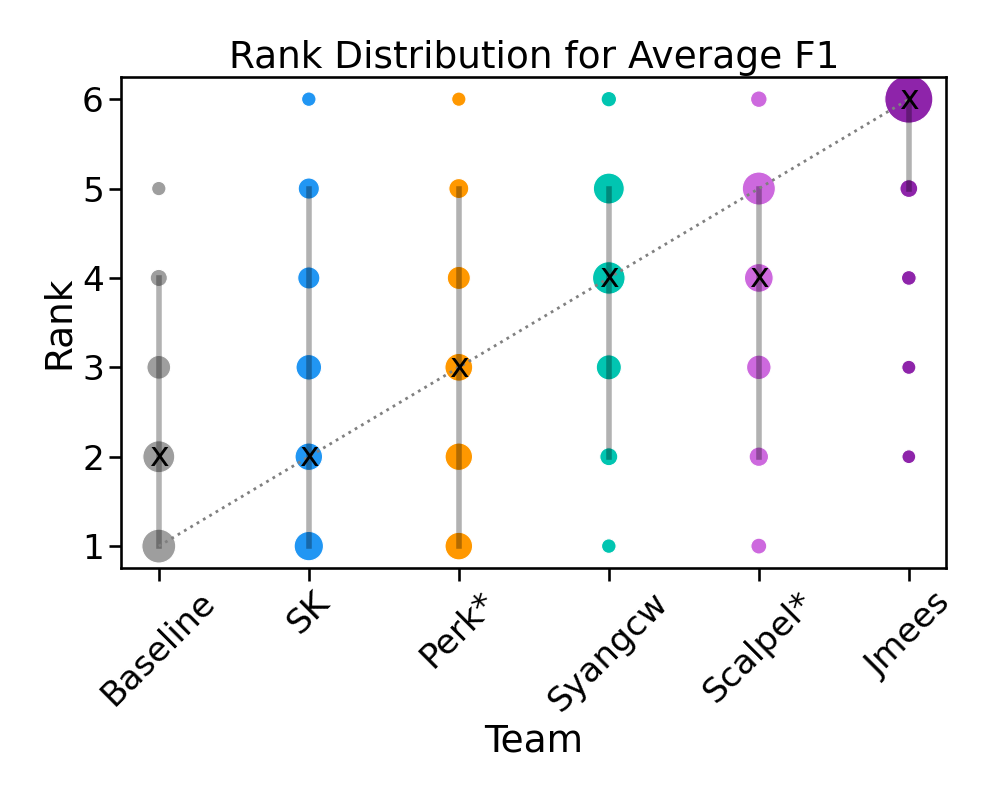}
        \label{fig:t2ranks_f1_24}
    \end{subfigure}
    \hfill
    \begin{subfigure}{0.48\textwidth}
        \centering\includegraphics[width=\linewidth]{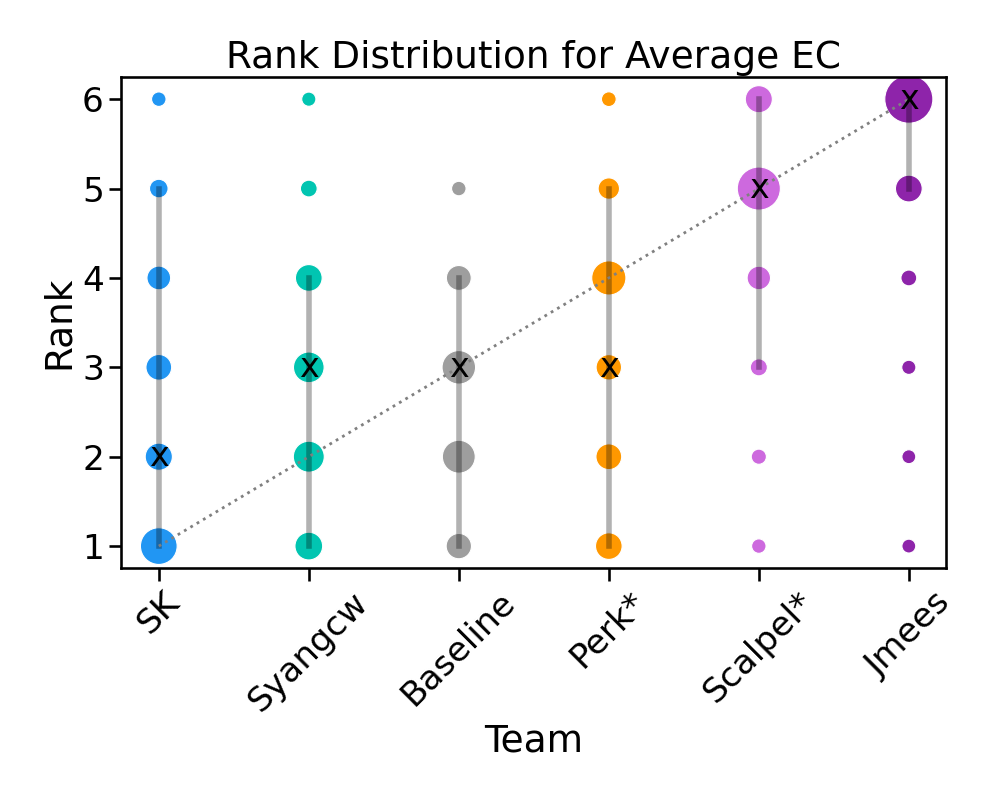}
        \label{fig:t2ranks_ec_24}
    \end{subfigure}
    
    \caption{Task 2: OSATS Metric performance (top row) and rank analysis (bottom row) for F1 and Expected Cost (EC) after bootstrapping with 10,000 repetitions. Error bars denote the standard deviation for the performance graphs. Rank plot bubble size corresponds to rank frequency, solid lines are 95\% confidence intervals, and crosses denote the median rank of that team. Team ranks are sorted by mode.}
    \label{fig:t2stats24}
\end{figure*}

The bootstrap and rank analyses shown in Figure~\ref{fig:t2stats24} support these results. The baseline model outperforms all other submissions in F1 but drops below team SK in EC. Team Perk, which submitted after the challenge deadline, closely approaches the baseline in F1 but falls short in EC. Team SK ranks between the baseline and Perk in F1 while outperforming both in EC.

The rank analysis reveals considerable variability across all teams, with 95\% confidence intervals spanning several ranks. Team SK, the top performer during the original challenge, maintains a strong position closely behind the baseline in F1 and ranks highest in EC, though it is closely followed by the baseline and team Syangcw. Team Perk shows a similar median rank but a slightly lower most frequent rank of four. All pairwise comparisons for both metrics yielded  $p<0.001$, confirming statistically distinct performance tiers between teams, with the sole exception of the baseline's and Syangcw's EC ($p=0.006$), whose close performance is visible in the bootstrap and rank graphs. The overall variability in ranks across bootstrap samples reflects model sensitivity to specific data subsets and may also indicate the intrinsic difficulty and heterogeneity of the task. When performance differences between teams are small, even minor changes in sampled data can alter leaderboard positions, suggesting that the task does not strongly separate methods or contains challenging cases handled inconsistently by different approaches. This underscores the importance of interpreting rankings together with uncertainty estimates rather than as fixed performance hierarchies.

When comparing to the results with the additional expert data, results improved as well. As shown in Table~\ref{tab:updateeval24}, Team Perk achieved the highest F1 score of 0.49 after retraining, outperforming all other teams as well as the baseline. Team SK also shows improved performance with an F1 score of 0.39 and achieving the best EC from all teams of 0.15. Team Jmees also benefits from the additional data, achieving an F1 score of 0.24 and an EC of 0.27, indicating a moderate improvement. Team Scalpel's performance also improves with an F1 score of 0.38 and an EC of 0.22, reflecting a positive impact from the additional expert data. 

\section{Participation and Results of the 2025 Challenge}\label{sec:participandresults25}
Seven teams submitted to the 2025 MICCAI Challenge: Saeid, Jmees, SK, MediSC\_OyeSS, Algoritmi, MoriLabNU, and Omar. Teams Saeid and SK submitted to Tasks 1 and 2; teams Omar and MoriLabNU submitted to task 3; teams Algorithmi, Jmees, and MediSC\_OyeSS submitted to all three tasks. Team Omar retracted their solution after the challenge took place on September 27th, 2025. After this date the challenge was closed to further submissions. Full implementation details to teams' methods can be found in the supplementary material.

\subsection{Task 1 and 2: Skill Assessment}
\subsubsection{Baseline}
The baseline model for tasks 1 and 2 is the same as the one used for the 2024 challenge, see Section~\ref{sec:participandresults24} for details.

\subsubsection{Team SK}
Team SK, comprising Satoshi Konso and Satoshi Kasai, participated in tasks 1 and 2. The team fundamentally changed their approach from the previous year, instead, utilizing clips from the beginning to evaluate skill.

The network architecture is identical for GRS and OSATS prediction, with models trained separately. An input video is processed by VideoMAE~\cite{Tong2012}, pretrained on Kinetics-400~\cite{Carreira2017}, which serves as the short-term video feature extractor and produces a temporal sequence of 768-dimensional features. These features are temporally pooled and passed through an MLP with one hidden layer, trained jointly as a classifier and regressor. The final outputs include classification results (4 classes for GRS, $5\times8$ classes for OSATS) and a regression result, where regression outputs are normalized to [0, 1] via softmax and scaled to the target range (GRS: 8-40, OSATS: 1-5). During inference, only the classification results are used.

\subsubsection{Team Saeid}
Team Saeid consisting of Saeid Rezaei submitted valid solutions for tasks 1 and 2. Unfortunately, his solution to task 3 did not produce a sucessful result. He implemented a Generative Reward Machine (GRM)~\cite{Rezaei2025} to automate the model optimization process for all tasks of the OSS challenge. By trying to learn promising model configurations from early training dynamics, the approach aims to significantly reduce the computational cost and time required to develop high-performing models for surgical video analysis.

From each video, 16 frames are uniformly sampled to form a representative clip, and each sampled frame is resized to $128\times128$ and normalized. For the backbone networks, the team used R(2+1)D-18 network~\cite{Tran2018} pretrained on Kinetics-400 for tasks 1 and 2. For both tasks, the model head is replaced with linear heads: 4 classes for GRS classification, 8 continuous values for OSATS regression.

The GRM framework operates in two stages: an offline estimation stage and an online search stage. In the first stage, an estimator is trained for each batch size class to predict final model performance from early training dynamics. In the second stage, a Q-learning agent performs an efficient online search over batch size-hyperparameter pairs using an $\epsilon$-greedy exploration strategy. After training a selected configuration for a limited number of epochs, a dynamics feature vector is extracted and a reward is computed based on the learned estimator from stage one. This reward guides the agent toward promising configurations, significantly reducing computational costs.

\subsubsection{Team MediSC\_OyeSS}
Team MediSC\_OyeSS, further referred to as MediSC, included Jieun Park, Soyeon Shin, Daehong Kang, Seungjae Hong, Youngbin Kong, Seoi Jeong, Kyu Eun Lee, and Hyoun-Joong Kong. They submitted solutions for all three tasks. Their solutions integrate temporal domain knowledge, focus on the latter phase of the procedure, and remove irrelevant background regions through preprocessing.

For tasks 1 and 2, 16 frames were extracted from a time interval near the end of the video. The time interval selected was based on expert observation. The frames were center-cropped to $448\times448$, resized to $224\times224$, and normalized. A ConvNeXt-base model~\cite{Todi2023} was used with a modified head for a four-class classification required for task 1. Task 2 employed a MVitv2-S~\cite{Li2021MViTv2} for spatiotemporal modeling and produces 8 category outputs corresponding to the OSATS subscales. During inference, frame-level predictions were aggregated through hard voting, with ties resolved randomly.

\subsubsection{Team Algoritmi}
Team Algorithmi included Tiago Jesus, Andr\'e Ferreira, Guillherme Barbosa, Jo\~ao Carvalho, Leonardo Barroso, Mariana Ribeiro, Nuno Gomes, Rafael Piexoto, Rodrigo Ralha, and Victor Alves. The team developed three distinct solutions for the three challenge tasks.

For Task 1, they designed a three-stage pipeline. Their approach begins with a non-linear frame sampling strategy to focus on the final, most informative stages of the surgical procedure. Each frame is then processed to create a 4-channel input tensor, augmenting the standard RGB data with the result of a Sobel edge-detecting operation to emphasize instrument and suture details. This sequence of feature-enhanced frames is fed into a fine-tuned InceptionV3~\cite{Szegedy2015} architecture, which leverages pre-trained ImageNet~\cite{Russakovskyimagenet15} weights to extract visual features. The model uses the information from the frame sequence to predict the final skill category, providing an end-to-end solution for automated GRS classification.

For Task 2, they designed a two-stage pipeline. In the first stage, a pre-trained YOLOv5~\cite{Redmon2015} object detection model is applied to dynamically identify and crop the most relevant region of interest (ROI) in each video frame. In the second stage, the cropped ROIs are processed using a ResNet50~\cite{He2016} backbone to extract visual features. To model the procedural flow and temporal dependencies of surgical tasks, the sequence of extracted features is fed into a Long Short-Term Memory (LSTM) network. The LSTM’s output is then used to predict scores across eight categories with separate classification heads.

\subsubsection{Team Jmees}
Team Jmees, consisting of Shunsuke Kikuchi, submitted solutions for all three tasks of the challenge. Task 1 was addressed as 4-bin classification via numeric regression and binning, Task 2 as 8D regression, and Task 3 as per-frame 27-keypoint tracking for 7 tracks (see Section~\ref{sec:tracking_jmees}).

Tasks 1 and 2 are handled similarly using a video transformer backbone, Swin3D~\cite{Yang2023}, augmented with a light 3D auxiliary segmentation head that operates on a coarse $7\times7\times T'$ spatio-temporal grid. This segmentation head, trained with pseudo-masks, introduces spatially-aware supervision to stabilize the temporal features and improve skill estimation robustness. The Swin3D model is pretrained with Kinetics-400~\cite{Carreira2017}, and the segmentation head is pretrained with Imagenet-22k~\cite{Russakovskyimagenet15}. The videos are downsampled to 10 fps and 96 frames are sampled evenly across the entire sequence. Frames are resized to $224\times224$. The main output heads of the Swin3D model predict either a single numeric GRS value (Task 1) or an 8-dimensional OSATS regression vector (Task 2), with auxiliary multitask outputs such as segmentation logits and class predictions for time, group, and sutures. The GRS score is subsequently categorized into the predefined classes.

\subsection{Task 3: Keypoint Tracking}
\subsubsection{Team MediSC\_OyeSS}
For task 3, the team used RTMDet~\cite{Lyu2022} (MMDetection) and HRNet-w48/CSPNeXt~\cite{Li2023HRNeXt,Chen2024} (MMPose) for surgical tool and hand detection, and pose estimation, respectively, pretrained with ImageNet and COCO. The Detection module detected tools and hands, cropped the found instances, and passed these to the pose estimation module. The pose model returned keypoints of the found instance as well as confidence scores.

\subsubsection{Team Algoritmi}
For Task 3, they implemented a three-stage pipeline combining object detection and classification with keypoint detection. In the first stage, a SAM2~\cite{Ravi2025} model segments the objects. All masks are then passed to the second stage, XGBoost~\cite{Chen2016}, to classify. For each object, an individual UNet model is used to predict the keypoints. To train the SAM2 network, the binary masks provided by the challenge were merged to a single multi-class mask.

\subsubsection{Team Jmees}
\label{sec:tracking_jmees}
Task 3 uses a heatmap-based keypoint detector. The architecture processes six input channels (RGB, optical flow, and foreground masks) through a ConvNeXt-based~\cite{Todi2023} encoder. The decoder employs a top-down fusion mechanism to generate heatmaps for keypoint position, visibility scores, and segmentation outputs via three separate heads, while temporal consistency is reinforced through flow warping. Heatmaps are further refined during inference using the maximum value with a centroid offset to predict precise keypoint coordinates and visibility states. The preprocessing stages involve generating optical flow using the method by \citet{Farnebaeck2003} and constructing the input from resized and normalized video frames. The output includes keypoint coordinates and other task-specific metrics, with post-processing ensuring alignment to the original video resolution.

\subsubsection{Team MoriLabNU}
Team MoriLabNU, for simplicity team Mori, included Xinkai Zhao, Yuichiro Hayashi, Masahiro Oda, Takayuki Kitasaka, and Kensaku Mori. They only submitted to task 3. The team implemented a four-stage pipeline: segmentation, hand keypoint estimation, instrument landmark extraction, and temporal tracking.

In the first stage, a Mask R-CNN~\cite{He2020} with a ResNet-50 backbone and feature pyramid network, initialized from COCO, detects hand bounding boxes and instrument masks per frame. In the second stage, each detected hand box is cropped at fixed resolution and processed by RTMPose-M~\cite{Jiang2023} using discretized coordinate encoding (SimCC) to predict six keypoints per hand. In the third stage, an auxiliary model computes a thin skeleton, principal axis, and local thickness map for each detected instrument mask. From these shape cues, three semantic landmarks per tool type are derived to describe hand-tool relations.

For tracking, association is based on agreement of anatomically meaningful hand keypoints, complemented by spatial overlap between boxes (OKS-IoU)~\cite{Ronchi2017}. A short exponential moving average reduces flicker without smoothing out fast gestures. Tracks are created conservatively to avoid identity fragmentation and removed after a short miss window. When instrument landmarks are available, three ordered points with visibility flags are appended to the track output.

\subsection{Scores and Ranks}\label{sec:results25}
This section presents the results of the 2025 challenge phase. A total of six teams participated in this phase (seven submissions, one retraction), including three teams from the previous MICCAI 2024 challenge and three new teams.

\subsubsection{Task 1: GRS-based Classification}
Task 1 participation included Teams Saeid, SK, Algoritmi, Jmees, and MediSc. The results of the evaluation are shown in the first two columns of Table~\ref{tab:eval25}. The baseline model outperforms all submissions with an F1 score of 0.65 and an EC of 0.12. Team Saeid achieved the best performance among the submitted models in both metrics, with an F1 score of 0.62 and an EC of 0.17. Team MediSC followed with an F1 score of 0.47 and an EC of 0.23. However, Jmees performed on par with MediSC, achieving a lower F1 score of 0.41 but a better EC of 0.21. Further details on the performance of each team can be found in the supplementary material.

\begin{table*}[ht]
\centering
\caption{Challenge 2025 Results. Baseline and best metrics per task and per metric are bolded.}
\label{tab:eval25}
\begin{adjustbox}{width=\columnwidth}
\begin{tabular}{l|cc|cc|c}
\toprule
 & \multicolumn{2}{c|}{\textit{Task 1: GRS}} & \multicolumn{2}{c|}{\textit{Task 2: OSATS}} & \textit{Task 3: Tracking} \\
\textbf{Team} & \textbf{F1 $\uparrow$} & \textbf{EC $\downarrow$} & \textbf{F1 $\uparrow$} & \textbf{EC $\downarrow$} & \textbf{HOTA $\uparrow$} \\
\midrule
\textbf{Baseline} & \textbf{0.65} & \textbf{0.12} & \textbf{0.47} & \textbf{0.16} & - \\
\midrule
Saeid & \textbf{0.62} & \textbf{0.17} & 0.19 & 0.33 & - \\
SK & 0.31 & 0.32 & 0.26 & 0.22 & - \\
Algoritmi & 0.32 & 0.27 & \textbf{0.30} & 0.21 & 0.15 \\
Jmees & 0.41 & 0.21 & 0.29 & \textbf{0.20} & \textbf{0.20} \\
MediSC & 0.47 & 0.23 & 0.30 & 0.21 & 0.16 \\
Mori & - & - & - & - & 0.01 \\
\bottomrule
\end{tabular}
\end{adjustbox}
\end{table*}

\begin{figure*}[ht]
    \centering
    \begin{subfigure}{0.48\textwidth}
        \centering
        \includegraphics[width=\linewidth]{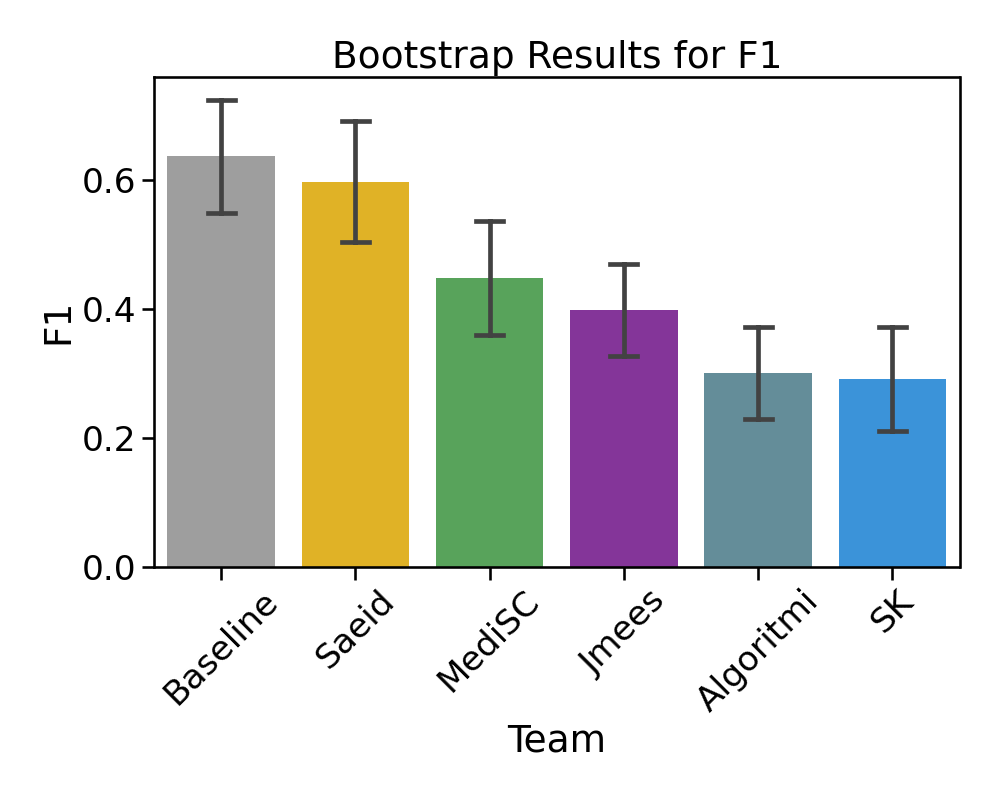}
        \label{fig:boot_f1_25}
    \end{subfigure}
    \hfill
    \begin{subfigure}{0.48\textwidth}
        \centering
        \includegraphics[width=\linewidth]{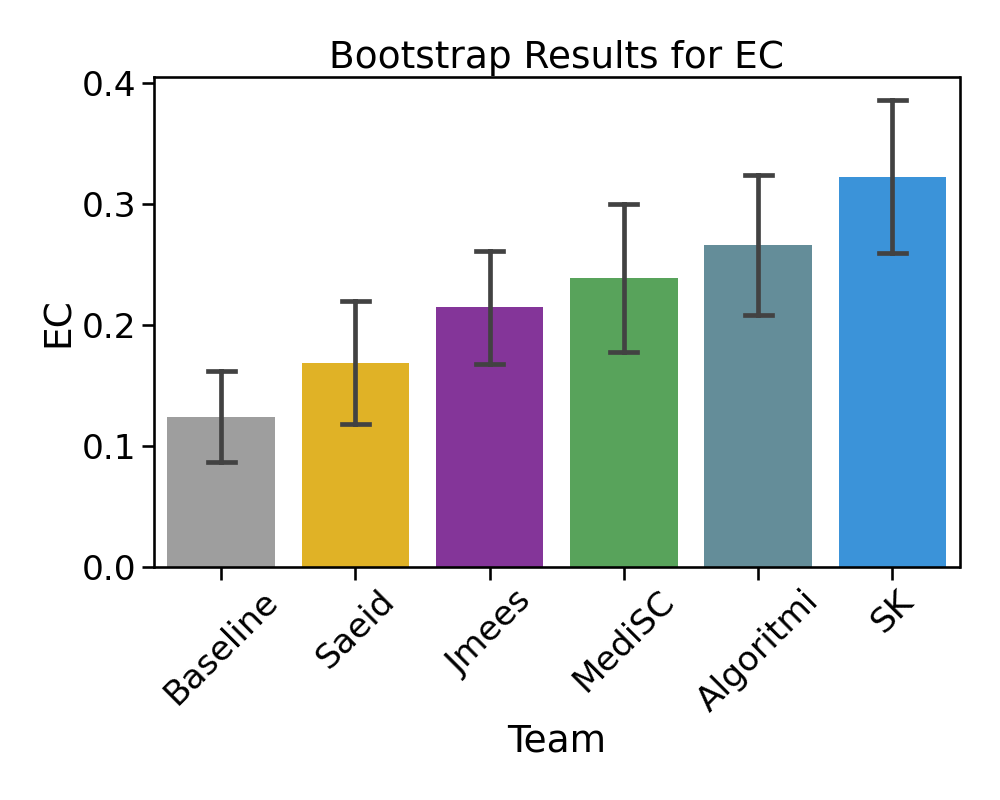}
        \label{fig:boot_ec_25}
    \end{subfigure}
    
    \vspace{0.5cm}
    
    \begin{subfigure}{0.48\textwidth}
        \centering\includegraphics[width=\linewidth]{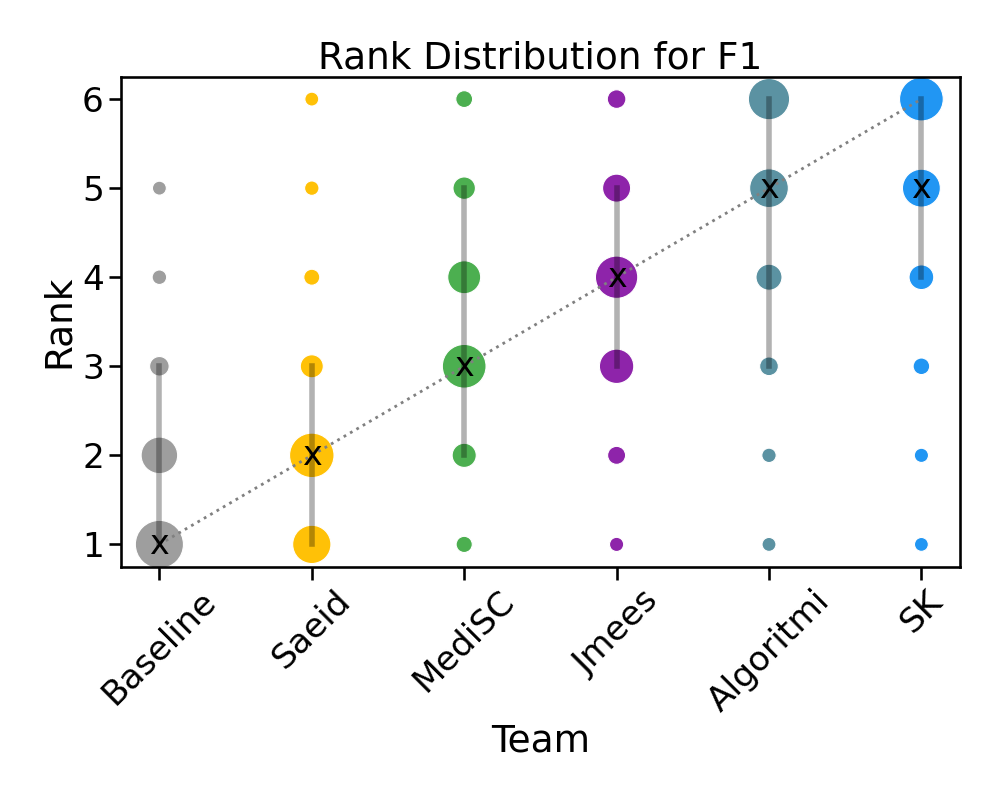}
        \label{fig:ranks_f1_25}
    \end{subfigure}
    \hfill
    \begin{subfigure}{0.48\textwidth}
        \centering\includegraphics[width=\linewidth]{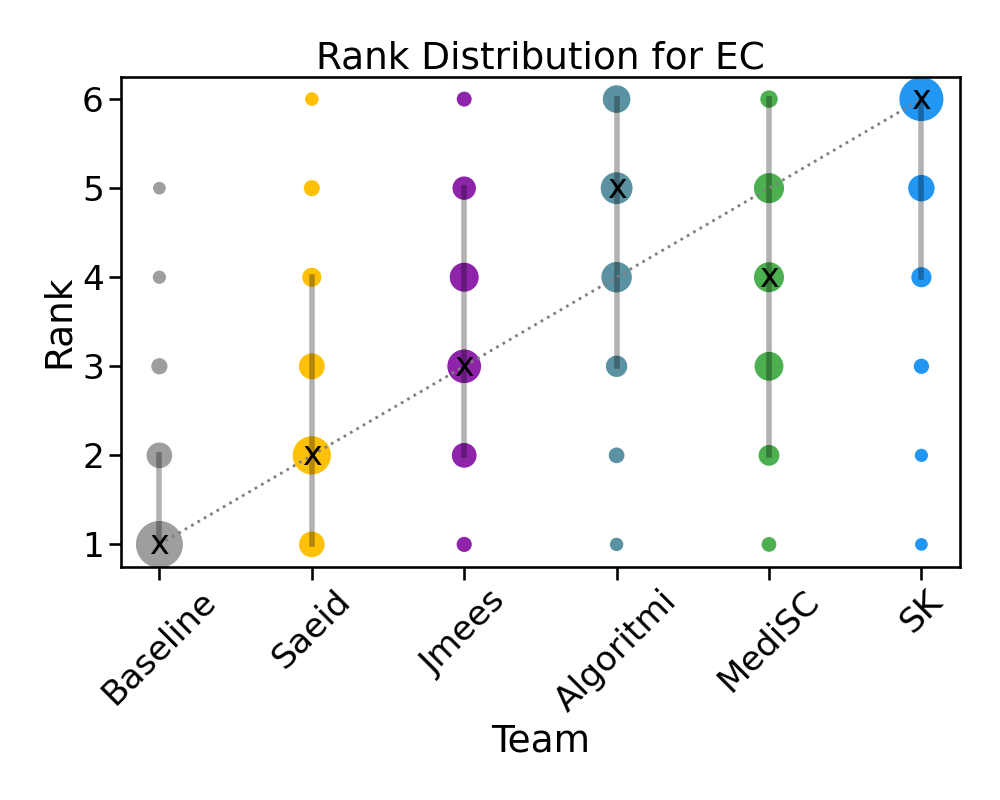}
        \label{fig:ranks_ec_25}
    \end{subfigure}
    
    \caption{Task 1: GRS Bootstrapping (top row) and rank analysis (bottom row) for F1 and Expected Cost (EC) after bootstrapping with 10,000 repetitions. Error bars denote the standard deviation for the performance graphs. Rank plot bubble size corresponds to rank frequency, solid lines are 95\% confidence intervals, and crosses denote the median rank of that team. Team ranks are sorted by mode.}
    \label{fig:t1stats25}
\end{figure*}

Saeid's superior performance is further supported by the bootstrapping and rank analysis shown in Figure~\ref{fig:t1stats25}. The bootstrap analysis reveals that the baseline model consistently outperforms all other submissions, with a clear margin in both F1 score and EC. Team Saeid shows a notable performance, closely approaching the baseline in F1 score, but still falling short in EC. For the remaining teams, the bootstrapping and rank analysis uncover some slight performance instablities although across the board bootstrapping results support the original performance. Team MediSC outperforms team Jmees in F1 but not EC which mirrors results without bootstrapping. In the rank analysis, performance between Algoritmi and MediSC for EC also seems to be very close. The variability in ranks among other teams indicates a less consistent performance across different bootstrap samples. However, all pairwise comparisons for both metrics yielded  $p<0.001$, confirming statistically distinct performance tiers between teams.

\subsubsection{Task 2: OSATS Prediction}
Performance in Task 2 was not as clear as for Task 1. Here, participating teams performed very close. Results can be viewed in the second column pair in Table~\ref{tab:eval25}. The same teams participated in Task 2 as for Task 1.

\begin{figure*}[ht]
    \centering
    \begin{subfigure}{0.48\textwidth}
        \centering
        \includegraphics[width=\linewidth]{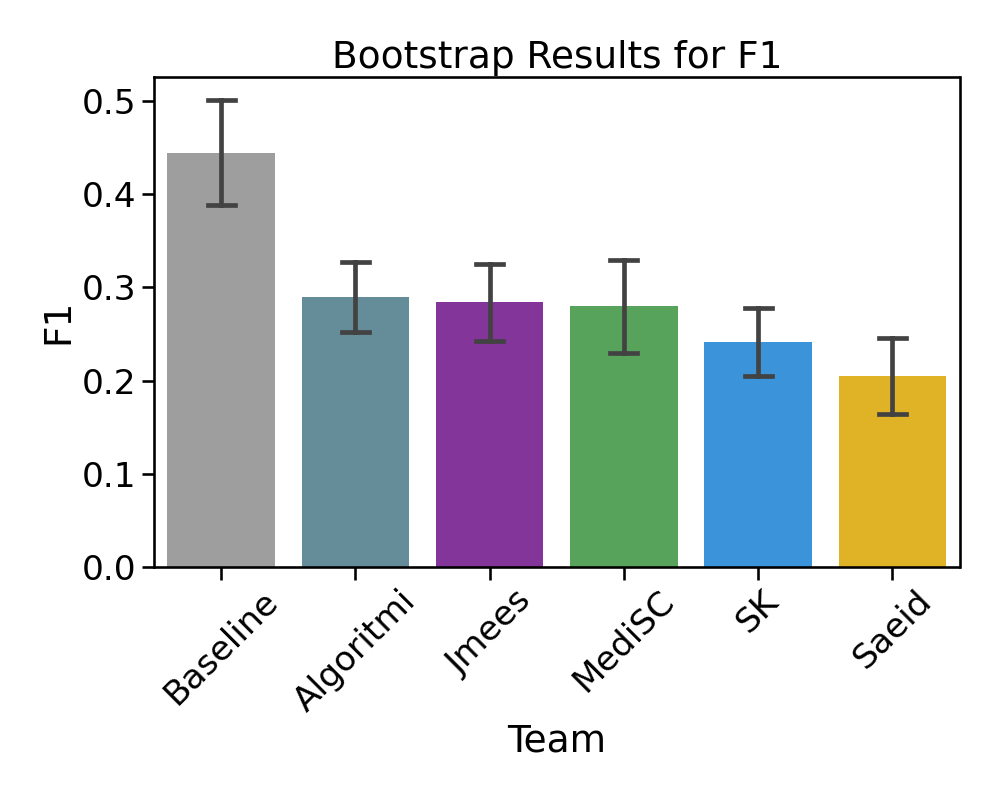}
        \label{fig:t2boot_f1_25}
    \end{subfigure}
    \hfill
    \begin{subfigure}{0.48\textwidth}
        \centering
        \includegraphics[width=\linewidth]{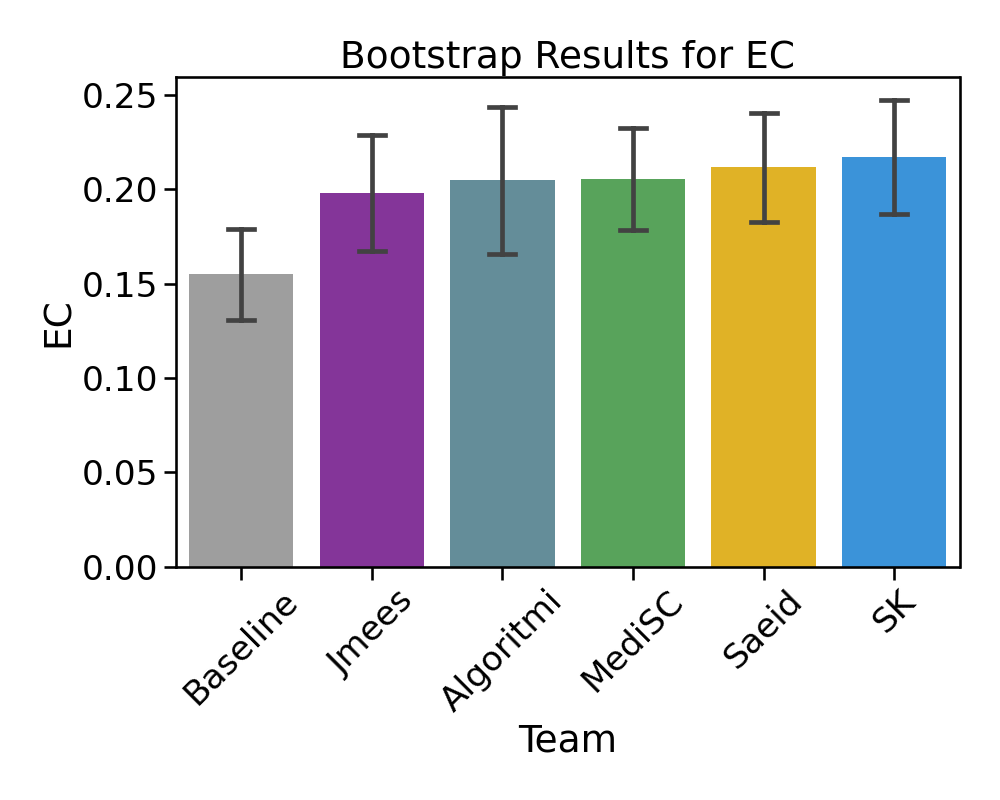}
        \label{fig:t2boot_ec_25}
    \end{subfigure}
    
    \vspace{0.5cm}
    
    \begin{subfigure}{0.48\textwidth}
        \centering\includegraphics[width=\linewidth]{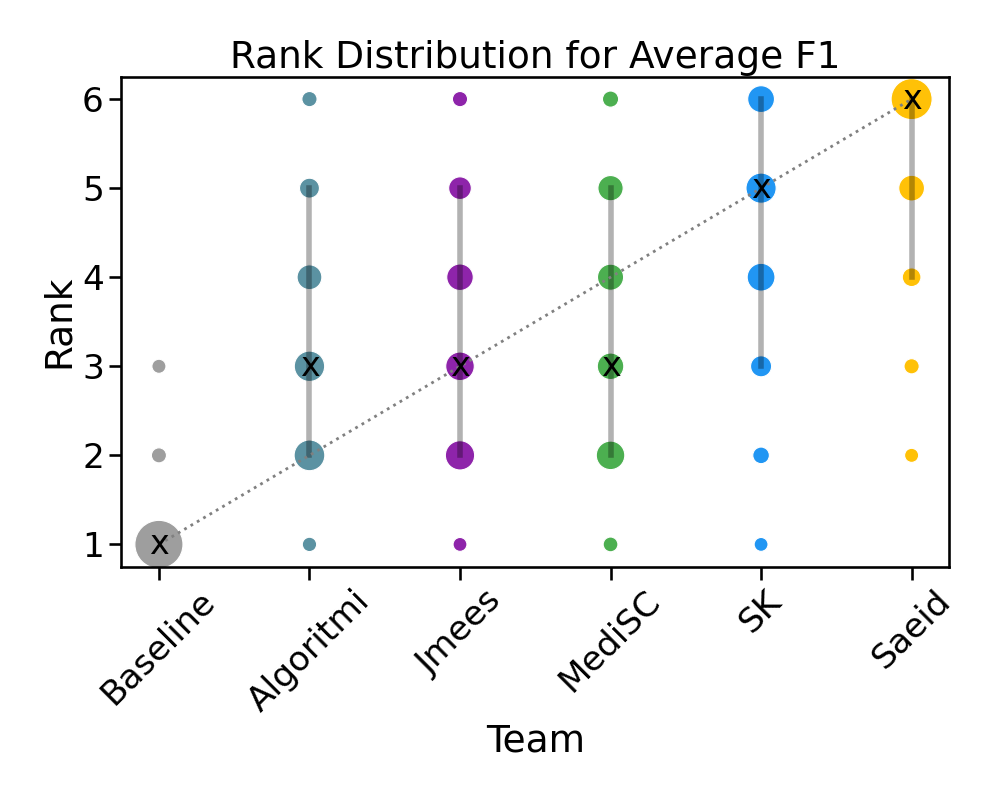}
        \label{fig:t2ranks_f1_25}
    \end{subfigure}
    \hfill
    \begin{subfigure}{0.48\textwidth}
        \centering\includegraphics[width=\linewidth]{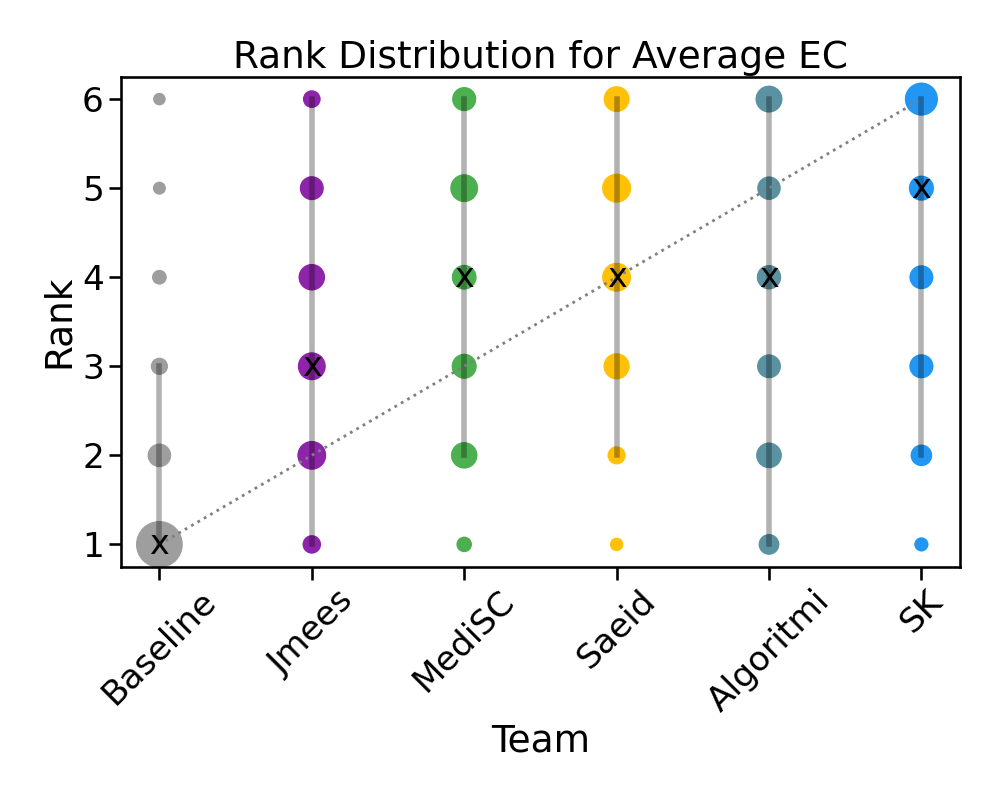}
        \label{fig:t2ranks_ec_25}
    \end{subfigure}
    
    \caption{Task 2: OSATS Bootstrapping (top row) and rank analysis (bottom row) for F1 and Expected Cost (EC) after bootstrapping with 10,000 repetitions. Error bars denote the standard deviation for the performance graphs. Rank plot bubble size corresponds to rank frequency, solid lines are 95\% confidence intervals, and crosses denote the median rank of that team. Team ranks are sorted by mode.}
    \label{fig:t2stats25}
\end{figure*}

As visible in the table, Team Algoritmi and Team MediSC performed very close with Team Algoritmi outperforming MediSC by a margin of 0.0032 for F1 score and 0.0011 for EC. Team Jmees performed similarly well, attaining an F1 score of 0.29. In terms of EC, Team Jmees achieved the best performance with an EC of 0.20, followed by Team Algoritmi and Team MediSC. Team Saeid and Team SK had lower performances in both metrics, with F1 scores of 0.19 and 0.26, and EC values of 0.33 and 0.22, respectively. This is similarly reflected in the bootstrapping results (see Figure~\ref{fig:t2stats25}) in which Algoritmi, Jmees, and MediSC all perform nearly the same for F1 with a slight edge for Jmees in EC. Despite the extremely close performance between teams, all pairwise comparisons for both metrics yielded  $p<0.001$, confirming statistically distinct performance tiers between teams, with the sole exception of Algoritmi and MediSC's EC ($p=0.0015$). In both results with and without bootstrapping, the baseline achieves clear superior performance, also achieving improved performance compared to the 2024 challenge edition.

Following the rank aggregation scheme, Jmees, Algoritmi, and MediSC tied for 1st place. Due to EC as tiebreaker metric, the first place went to Jmees. Taking the bootstrapping, the Wilcoxon test results, and the closeness of scores into consideration, it was decided to award both Algoritmi and MediSC second place.

\subsubsection{Task 3: Tracking}
\begin{figure*}
    \centering
    \begin{subfigure}{0.48\textwidth}
        \centering
        \includegraphics[width=\linewidth]{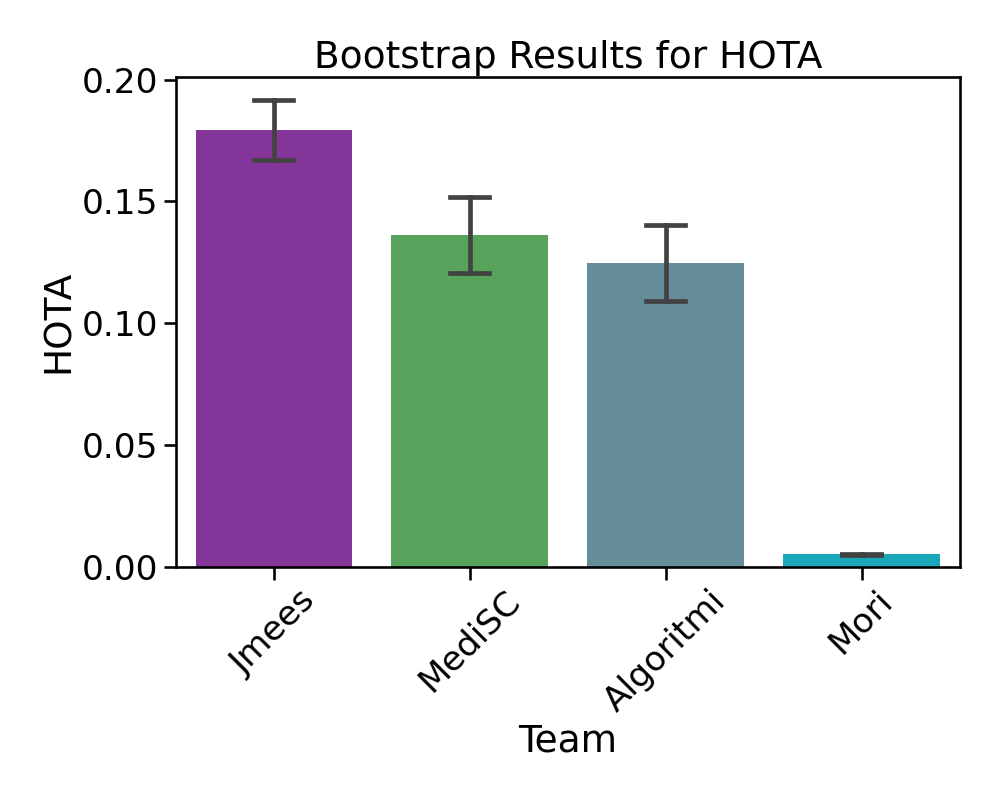}
        \label{fig:t3boot}
    \end{subfigure}
    \hfill
    \begin{subfigure}{0.48\textwidth}
        \centering
        \includegraphics[width=\linewidth]{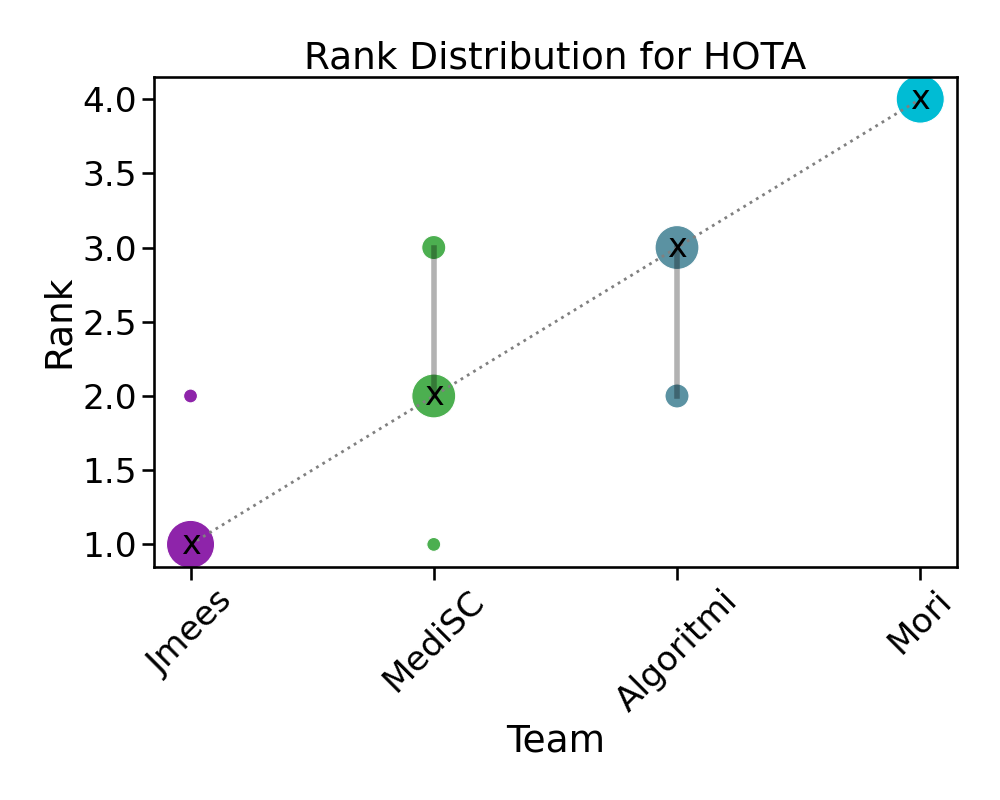}
        \label{fig:t3ranks}
    \end{subfigure}
    
    \caption{Task 3: Tracking Bootstrapping (left) and rank analysis (right) for HOTA after bootstrapping with 10,000 repetitions. Error bars denote the standard deviation for the performance graphs. Rank plot bubble size corresponds to rank frequency, solid lines are 95\% confidence intervals, and crosses denote the median rank of that team. Team ranks are sorted by mode.}
    \label{fig:t3stats25}
\end{figure*}

Task 3 participation included Teams Algoritmi, Jmees, MediSC, and Mori. The results are shown in the last column of Table~\ref{tab:eval25}. Team Jmees achieved the best performance with a HOTA score of 0.20, followed by MediSC (0.16) and Algoritmi (0.15), while Team Mori attained the lowest score (0.01). The bootstrapping and rank analysis shown in Figure~\ref{fig:t3stats25} further supports these findings, with Team Jmees consistently outperforming the other teams across bootstrap samples. However, pairwise significance testing reveals a more nuanced picture: while Jmees vs. MediSC yields $p=0.0013$, the comparison between Algoritmi and MediSC produces $p=0.3$, indicating no statistically significant separation between these two teams despite their apparently distinct performance in the bootstrap and rank graphs. This discrepancy may be attributable to the small test set size, which limits the statistical power of pairwise comparisons and can mask differences that appear visually meaningful in the bootstrapped distributions.

\section{Discussion}\label{sec:discussion}
The discussion covers the results of both the 2024 and 2025 challenge phases, with a focus on the performance of participating teams across the three tasks, as well as insights gained from the bootstrapping and rank analyses. This allows for a unique opportunity to analyze how teams adapted and improved their approaches based on the feedback and results from the initial challenge along with gaining insights on the effect of the data on the methods.

The section is structured into three subsections, each dedicated to one of the tasks: Task 1 (GRS classification), Task 2 (OSATS multiclass classification), and Task 3 (Tracking). For each task, it will discuss the performance of the teams, the impact of the additional expert data on their results for the 2024 challenge, and any notable trends or observations that emerged from the evaluation of the 2024 versus the 2025 versions. The section will also compare the results across tasks to identify any common themes or differences in team performance and methodological approaches.

\subsection{Task 1: GRS-based Classification}
\begin{figure*}[htbp]
    \centering

    \begin{subfigure}[t]{0.33\textwidth}
        \centering
        \includegraphics[width=\linewidth]{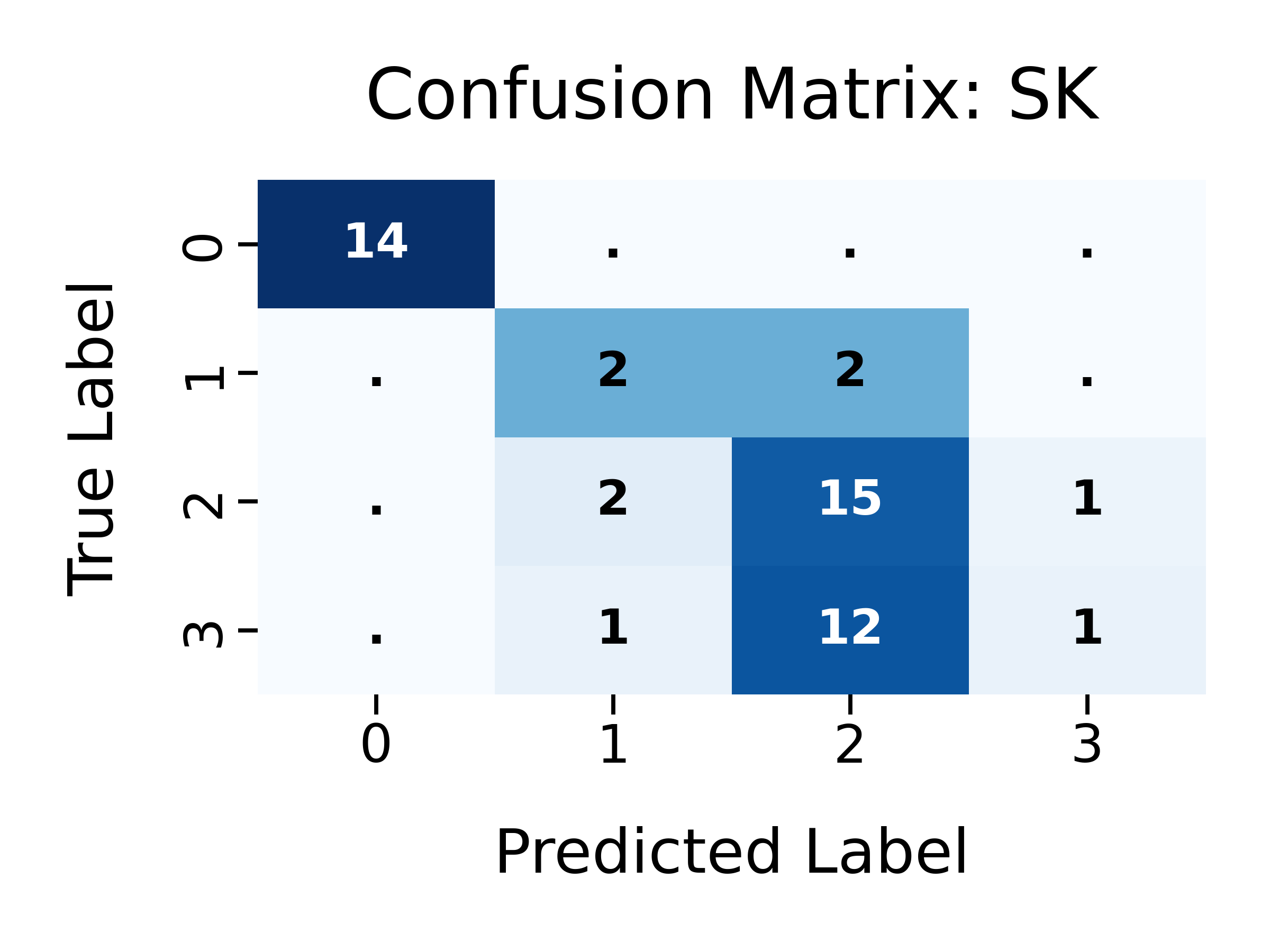}
    \end{subfigure}\hfill
    \begin{subfigure}[t]{0.33\textwidth}
        \centering
        \includegraphics[width=\linewidth]{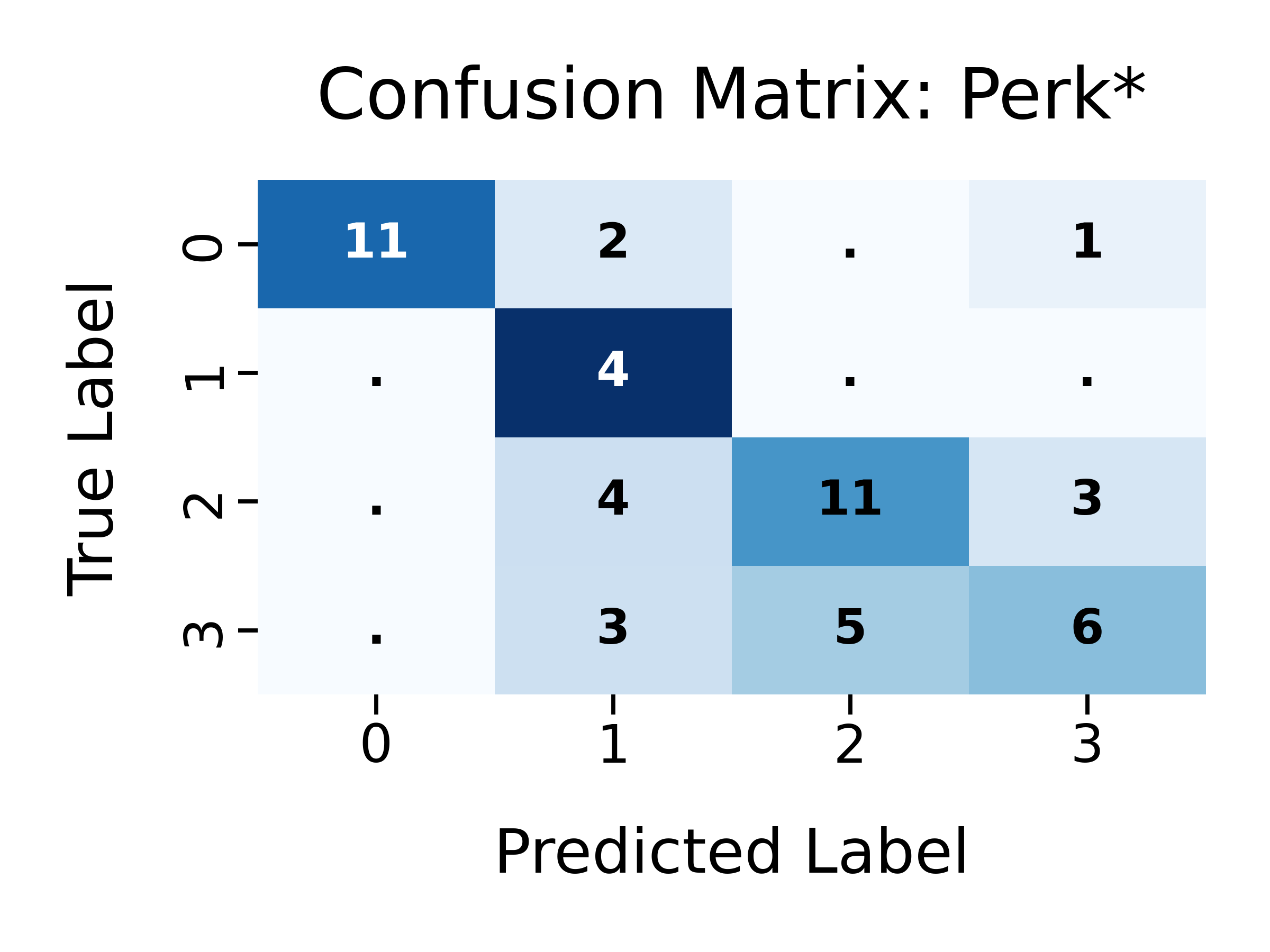}
    \end{subfigure}\hfill
    \begin{subfigure}[t]{0.33\textwidth}
        \centering
        \includegraphics[width=\linewidth]{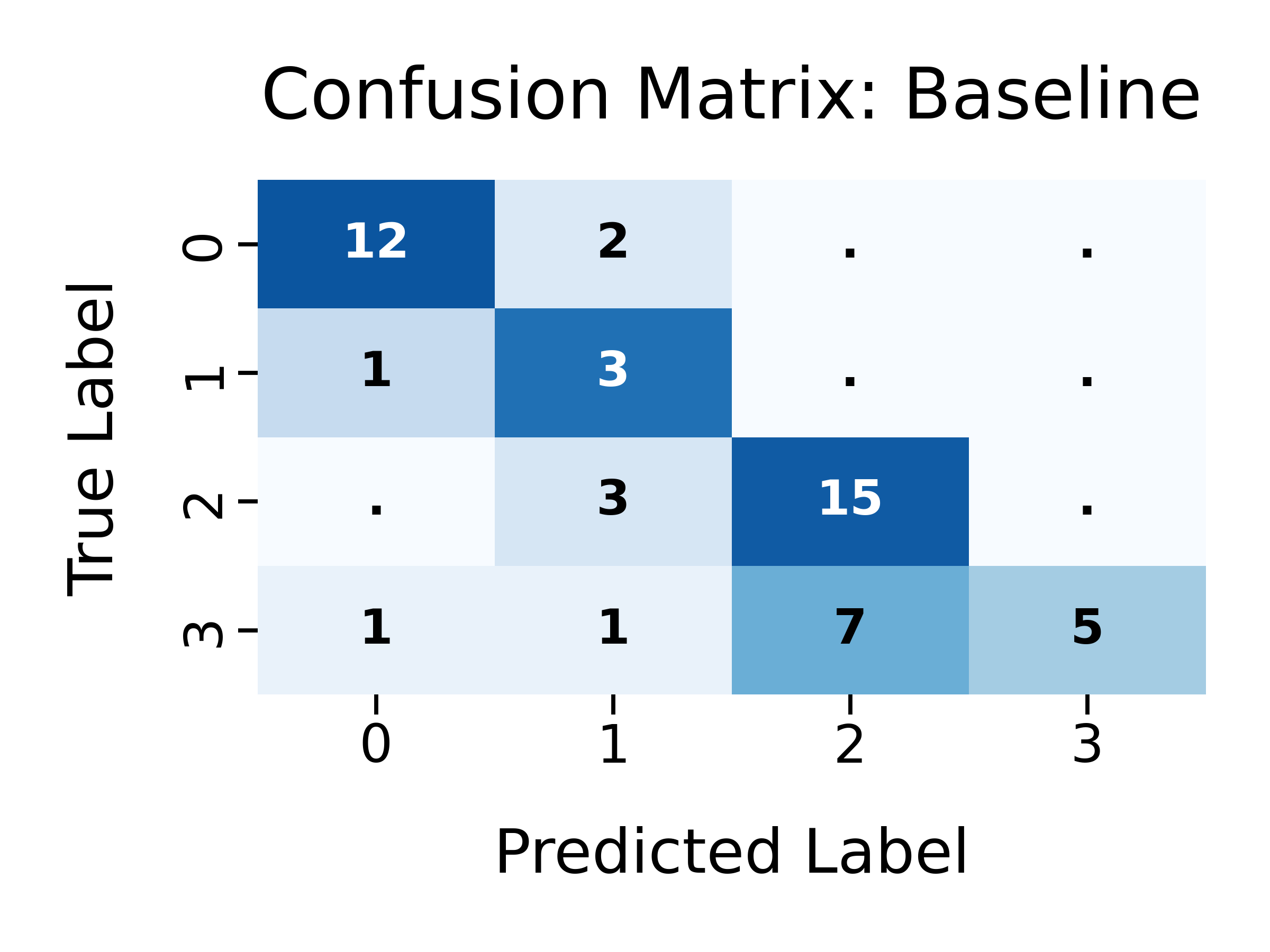}
    \end{subfigure}

    \vspace{0.6em}

    \begin{subfigure}[t]{0.33\textwidth}
        \centering
        \includegraphics[width=\linewidth]{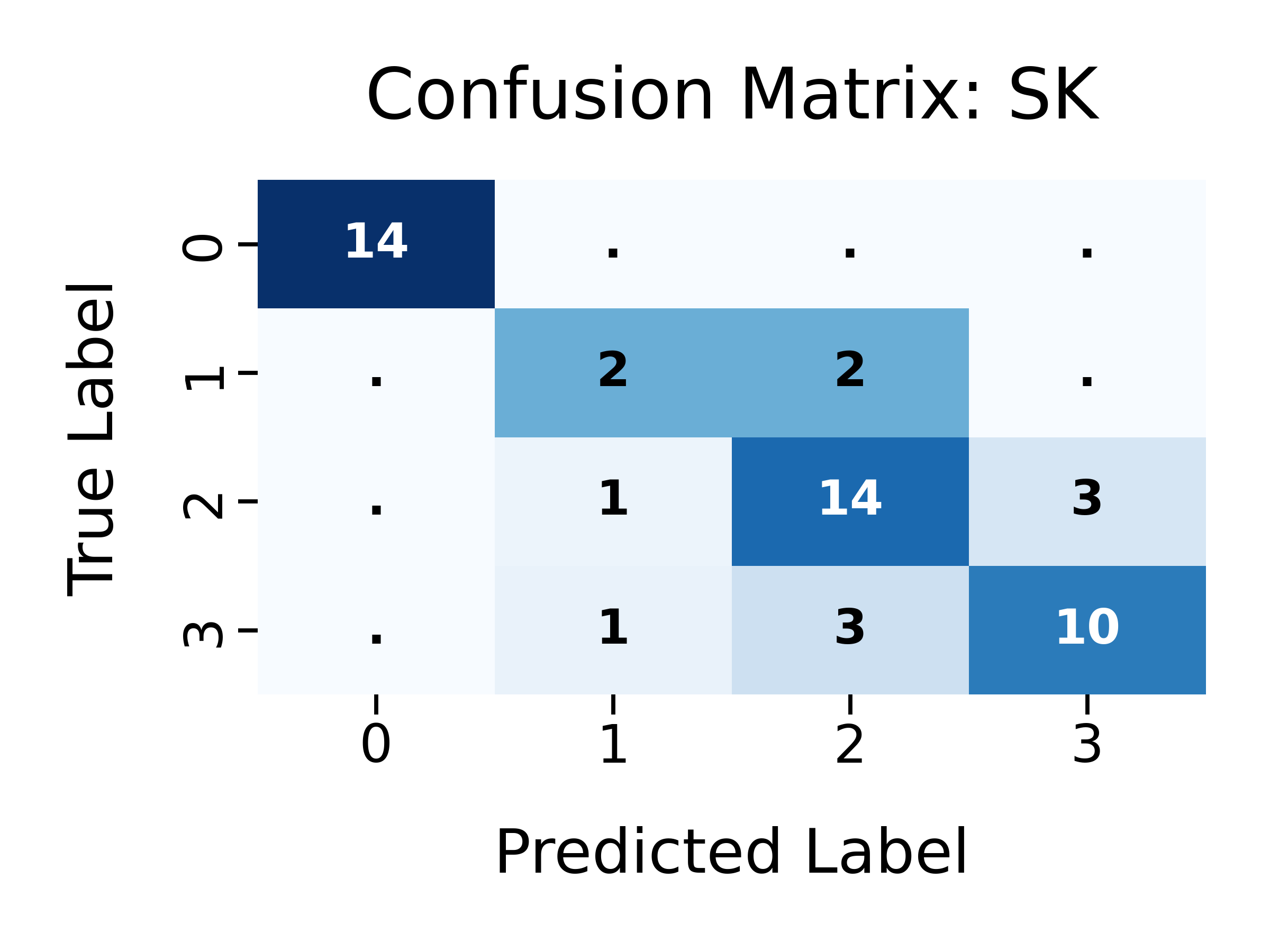}
    \end{subfigure}\hfill
    \begin{subfigure}[t]{0.33\textwidth}
        \centering
        \includegraphics[width=\linewidth]{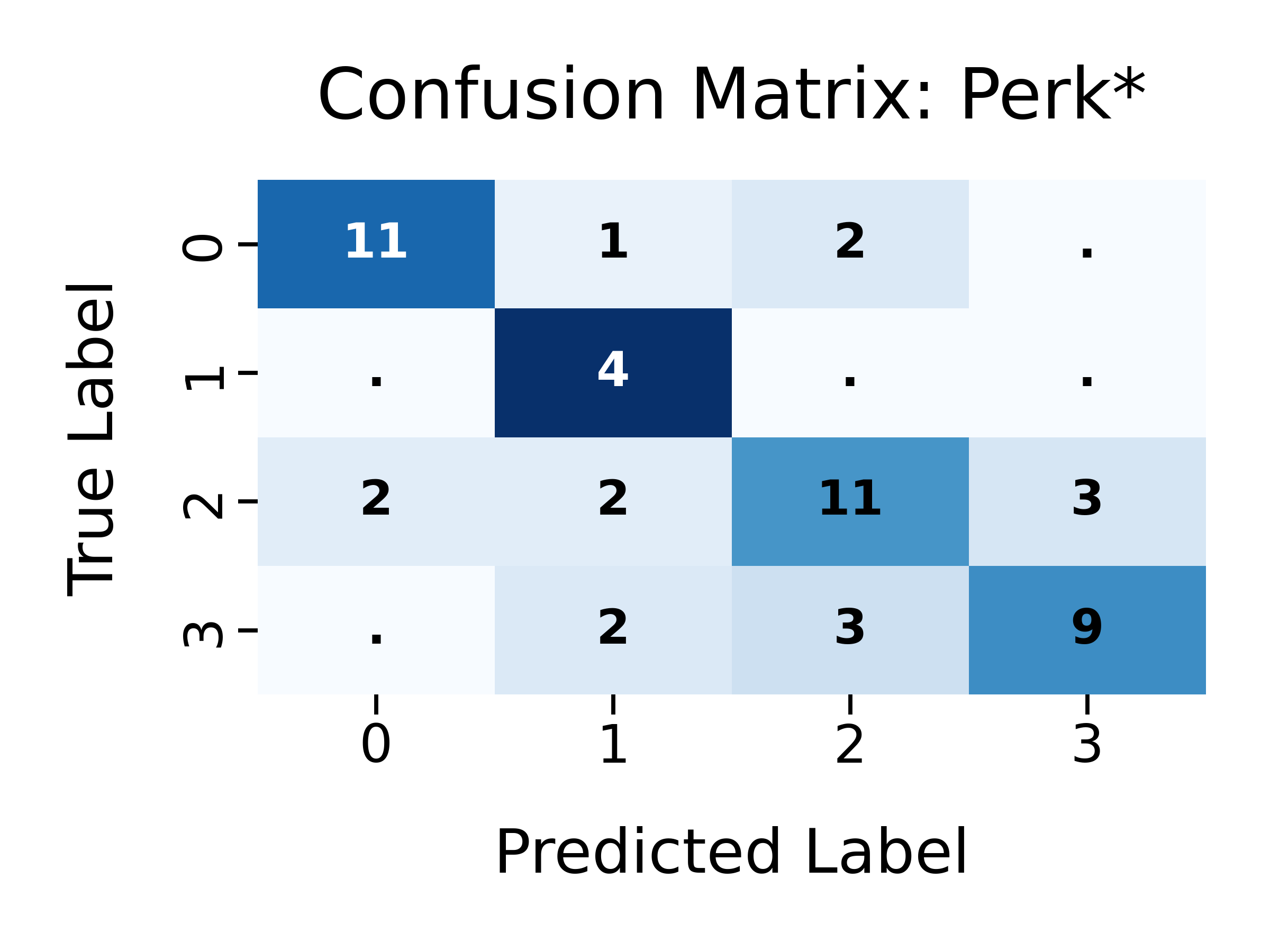}
    \end{subfigure}\hfill
    \begin{subfigure}[t]{0.33\textwidth}
        \centering
        \includegraphics[width=\linewidth]{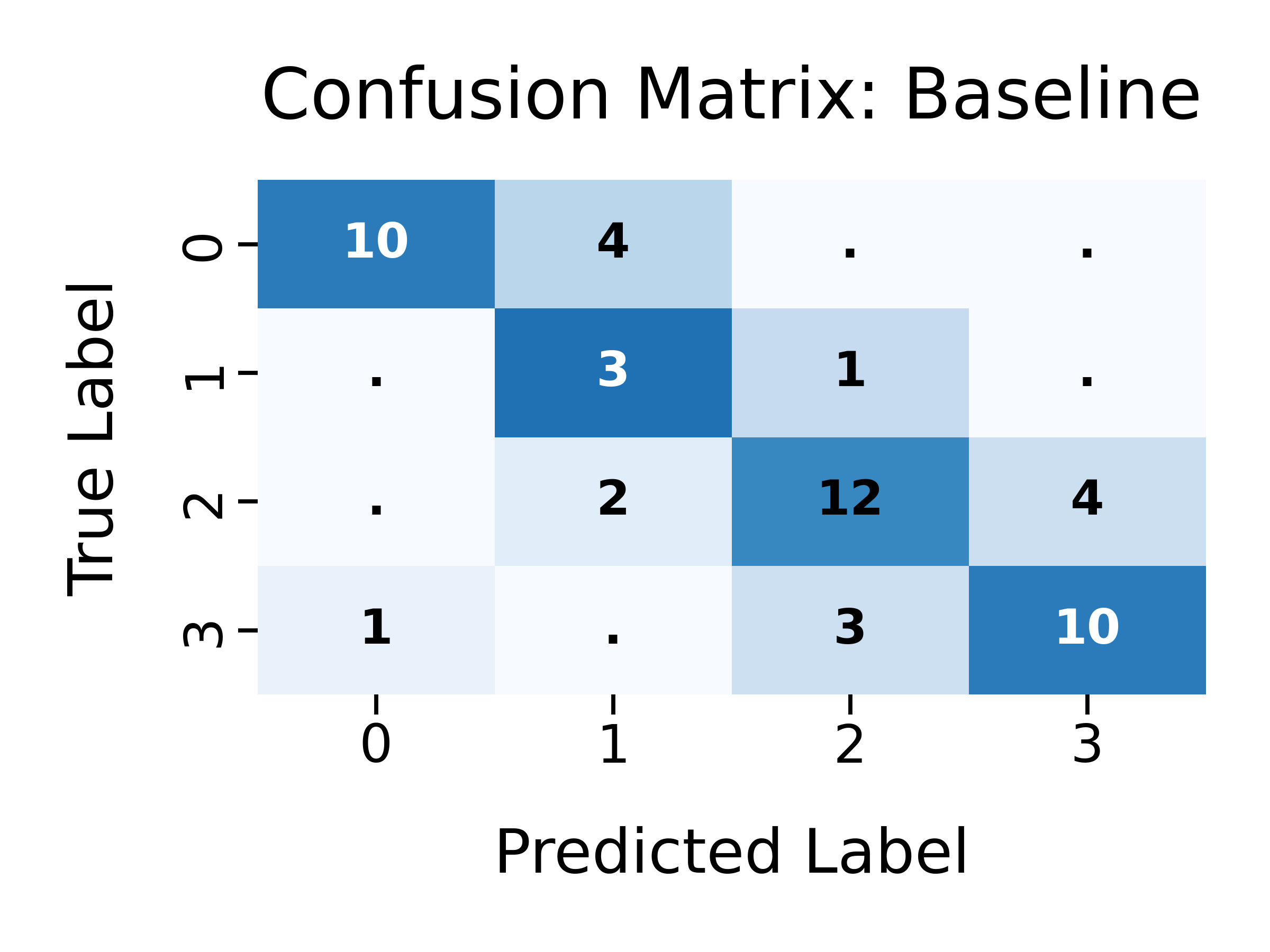}
    \end{subfigure}

    \caption{Task 1 confusion matrices of teams SK, Perk, and Baseline from the 2024 Challenge (top) compared with the additional expert training data confusion matrices (bottom).}
    \label{fig:cfm24_task1}
\end{figure*}

The 2024 challenge produced three particularly successful approaches: team SK (suture counting from final video frames), team Perk (phase-specific motion-based metrics combined with workflow features), and the baseline (a generic end-to-end video classification model). These fundamentally different strategies all achieved strong results, demonstrating that methodologically contrary approaches can succeed at comparable levels when executed well. This underscores the importance of training strategy, data augmentation, model selection, and data curation alongside core methodology. Notably, the baseline is a generic, relatively low-effort end-to-end model that performed competitively without extensive hand-engineering, while team Perk introduced considerable additional annotation effort to derive specialized features. Confusion matrices illustrating how each approach improved with additional expert data are shown in Figure~\ref{fig:cfm24_task1}.

Team SK's approach centers on counting completed sutures visible at the end of the video and mapping this count to a predicted skill level. Despite its simplicity, SK performed best among all participating teams, though compared to the baseline it only matches on expected cost. The F1 score shows SK achieves a close absolute score, while the EC reflects accurate skill-level ordering. This demonstrates that substantial task-relevant information is contained in the final suture count, which, as the team correctly identified, strongly correlates with GRS-based skill level.

In contrast, teams Jmees and Algoritmi, both incorporating tool or hand tracking as a first processing step, faced considerable difficulties. Jmees struggled to determine correct absolute skill levels while maintaining reasonable hierarchical ordering: as shown in Fig.~\ref{fig:cfm24_task1}, they produced no predictions for the expert category and no true positives for the intermediate category. Unlike Algoritmi, Jmees also predicted auxiliary information from training labels (similarly to SK), but built this on top of their tracking pipeline rather than direct visual features, compounding tracking errors with multi-task proxy prediction complexity. For both methods, the quality of their underlying tracking remained unclear, and they were likely disadvantaged by limited tracking-specific annotations. Additionally, noisy hand movement features further compounded these challenges, foreshadowing the tracking difficulties encountered in Task 3 of the 2025 challenge.

Team Perk performed exceedingly well while also using motion signals. Crucially, Perk combined motion features with workflow metrics (including suture counts, similarly to SK) and temporal duration measures, such as time per stitch, which typically correlate with surgical skill~\cite{Funke2026}. Furthermore, Perk trained dedicated tracking models using their own annotated data, yielding more reliable motion signals than teams relying on off-the-shelf tracking. This combination of workflow-aware features with robust, purpose-trained motion extraction likely accounts for Perk's strong performance relative to other tracking-based approaches.

Team Scalpel's mid-table performance further illuminates what distinguished Perk. Scalpel pursued a conceptually similar strategy, extracting stitch and movement descriptors from self-annotated tracking data and incorporating aggregated motion metrics (average speed, acceleration) rather than structured workflow representations. This placed Scalpel above Jmees and Algoritmi but clearly below Perk, suggesting two factors at play. First, dedicated training annotations for tracking models likely improved tracking quality relative to Algoritmi's more generic approach. Second, unlike Perk, Scalpel lacked explicit workflow context such as phase segmentation, suture counts, or per-stitch efficiency measures, leaving their temporal features too coarse to capture skill-relevant procedural structure. Scalpel's advantage over Jmees may additionally stem from leveraging intermediate network features from their detection model rather than relying on raw tracking coordinates, potentially retaining richer representations, though this interpretation remains speculative without further ablation.

One possible explanation for SK's strong performance is that the GRS-based skill level may be partly product-biased, meaning raters scored predominantly based on the final outcome. Both SK and the baseline appear to succeed in part by effectively capturing product-related information, with the 3D CNN potentially gaining an advantage by aggregating cues across frames. However, this product-bias hypothesis remains speculative and has not been formally studied. Automatic methods naturally discover factors correlating with annotated ground truth and will exploit shortcuts when training data is limited, but this does not mean the annotated score is reducible to these correlating factors alone. In the suturing task, a good result is typically linked with a skilled process, much as task completion time often correlates with GRS without fully capturing skill~\cite{Funke2026}. Without further analysis, it cannot be concluded that the baseline attends exclusively to product-related features. Indeed, research has shown that appearance bias in 3D convolutional networks stems from training dataset properties rather than being inherent to the architecture~\cite{Byvshev2022}, and 3D CNNs have proven effective for temporal tasks such as gesture recognition in surgical video~\cite{Funke2019-gesture}. The baseline's success therefore more likely reflects its capacity to learn both spatial and temporal cues relevant to the task. Team Perk's strong performance near the baseline level indicates that explicit encoding of product- and process-relevant descriptors can, when well-executed, approach the performance of implicit feature learning from video for GRS-based skill level prediction. Among lower-performing teams, Jmees and Algoritmi's reliance on tracking without sufficient annotated data likely contributed to weaker results, while Syangcw's transformer-based approach performed somewhat better, though the limited data regime posed challenges for its architecture~\cite{Peruzzo2024}. Two further caveats apply: predictions target a coarse GRS-based skill classification rather than the continuous GRS score, so conclusions may not generalize to the full scale; and as discussed below, end-focused methods did not necessarily perform well in the 2025 iteration, cautioning against over-interpreting the product-bias hypothesis from a single year's results.

The additional expert-level data improved performance across all teams, as expected given the increased training set size. However, given the small test set, these improvements should be interpreted cautiously, as reclassifying even a single video could substantially change the macro F1 score. Confusion matrices for the three leading methods (baseline, SK, and Perk) reveal what specifically changed in their predictions (see Fig.~\ref{fig:cfm24_task1}).

Particularly interesting is the shift in relative performance: while Perk outperforms SK with the original challenge dataset, SK overtakes Perk with the additional data. This suggests SK's suture-counting approach benefits disproportionately from additional expert data, likely because these examples strengthen the learned mapping between suture count and skill level. In contrast, Perk's process-oriented features do not benefit to the same degree, possibly because the additional data does not sufficiently enrich the motion and workflow signals Perk relies on. Notably, Perk performs very well on Task 2 and, with additional expert data, clearly outperforms both SK and the baseline on OSATS prediction, indicating that its features retain considerable value for more fine-grained, multi-dimensional assessment. This suggests that when sufficient data establishes a reliable mapping, explicitly engineered product descriptors can match or exceed process-oriented approaches for coarse skill classification, though the picture may differ for more detailed assessment dimensions.

In the 2025 iteration, the baseline again achieved the strongest performance, further cementing its robustness for GRS-based skill prediction. Team Saeid placed second, consistent with first-iteration observations: the R(2+1)D architecture retains effective spatial inductive biases while decomposing the temporal dimension efficiently, and systematic hyperparameter tuning likely maximized model capacity on a limited dataset. Team Jmees' third-place finish using a Video Transformer backbone represents a marked improvement over team Syangcw's transformer-based 2024 submission, suggesting the transformer paradigm is not inherently unsuitable but that architectural choices, training strategy, and implementation details matter considerably. Team MediSC's ConvNext approach placed fourth with competitive performance, bearing similarity to SK's successful 2024 strategy in its focus on discriminative spatial features. At the lower end, team Algoritmi placed fifth despite explicitly focusing on end-of-video frames; their combination of RGB and edge features with an older InceptionV3 architecture may have introduced unnecessary complexity without proportional gain. Team SK placed last despite their strong 2024 showing, having abandoned their successful final-frame strategy in favor of a VideoMAE backbone applied to early-video clips with a Softmax-based regression formulation. The combination of self-supervised representations applied to less discriminative early content with a suboptimal loss likely accounts for their decline. Together, the cases of Algoritmi and SK suggest that simply attending to the video's end is insufficient without appropriate representation and mapping, while SK's decline after shifting to early-video content may lend further support to the product-bias hypothesis, as early portions contain little information about the final surgical outcome. However, confounding architectural choices in both cases make it difficult to isolate the effect of temporal segment selection.

A significant shift in model architectures occurred between the two iterations: while the 2024 challenge featured many tracking and proxy-based approaches, the 2025 challenge saw more spatiotemporal video models, most likely influenced by the 2024 results in which 3D CNN-centered approaches clearly outperformed all other submissions. Across both iterations, the overarching pattern is that generality combined with careful model tuning tends to outperform more specialized but less thoroughly optimized approaches. However, for formative assessment, where the goal is to provide trainees with actionable feedback on how they perform rather than merely what they produce, process-oriented approaches remain essential even if they correlate less strongly with current GRS-based classifications. Realizing this potential requires fine-grained, skill-related annotation of each video segment, a substantial practical challenge that remains largely unaddressed.

\subsection{Task 2: OSATS Prediction}
From Task~1 to Task~2 there is a significant drop in F1 performance across all teams, unsurprising for several reasons. The OSATS comprises multiple operative competencies, transforming the problem from single-output to multi-output classification where each sub-category may depend on different aspects of performance. The move from four classes to five inherently increases prediction difficulty. Additionally, inter-rater reliability for individual OSATS categories tends to be lower than for the aggregate GRS, introducing additional noise into the ground truth. This pattern holds across both the 2024 and 2025 challenges.

In the 2024 challenge, the baseline achieved the strongest results, suggesting its spatiotemporal features capture the broadest range of quality indicators across OSATS dimensions. Notably, strong performance on the more process-oriented OSATS sub-categories indicates that the baseline captures more than product-related appearance features. Team SK followed in second place with Perk close behind, a flipped placement and narrower gap compared to the GRS task. This convergence may indicate that predicting the full OSATS table requires a more balanced combination of product and process information, as some sub-categories are inherently more process-oriented than the overall GRS-based skill level. The comparison between Perk and Scalpel is instructive: despite pursuing similar stitch and movement descriptor strategies, Perk's inclusion of workflow features provided an advantage over Scalpel (fifth place), which lacked procedural context such as phase segmentation or per-stitch efficiency measures. This suggests that understanding procedural structure contributes meaningfully when predicting across multiple assessment dimensions, though it also reflects the broader theme that quality of execution matters as much as the conceptual approach. Team Syangcw placed slightly above Scalpel despite its previously noted architectural challenges, while Jmees placed last, consistent with their Task~1 difficulties. Overall, the uniformly lower performance across all teams confirms that predicting fine-grained, multi-dimensional surgical skill from video remains an open challenge, and that features sufficient for coarse GRS-based classification do not straightforwardly generalize to individual OSATS competencies.

The additional expert data improved performance for all 2024 teams, though given the small test set, these improvements should be interpreted cautiously. The additional data likely provided more examples of nuanced OSATS scoring patterns, underscoring the importance of sufficiently large and diverse training sets for complex assessment tasks. Notably, with the additional data, team Perk clearly outperforms both SK and the baseline on OSATS prediction, suggesting that process-oriented features benefit substantially from increased and better balanced training data in this multi-dimensional context. This offers hope that more thorough evaluation may become achievable as datasets grow. However, the baseline still performs competitively, especially regarding EC, and required no additional annotation effort, providing a viable alternative to more complex hand-engineered approaches in the current data regime.

In the 2025 challenge, individual team performance was generally similar or slightly improved compared to their 2024 results, but global performance was worse (F1 scores at or below 0.3, EC at or above 0.2), with the notable exception of the baseline, which again achieved the best performance. Among participating teams, Algoritmi, MediSC, and Jmees performed at a broadly similar level despite different architectural choices: Algoritmi adapted by switching to a ResNet-LSTM combination and MediSC employed a Video Transformer, but neither yielded a decisive advantage, and the performance gap to the baseline remained large. Jmees' Video Transformer held steady relative to its GRS ranking, suggesting reasonable cross-task generalization. Most revealing are the poor results of teams SK and Saeid. SK's struggles carried over from the GRS task due to its suboptimal VideoMAE design reliant on early-video content. Saeid's R(2+1)D network, which achieved first place among participating teams on the single-output GRS task, proved inadequate for multi-output OSATS prediction, likely due to optimization difficulties when tuning for multiple scores simultaneously.

The overarching lesson from both iterations is that OSATS prediction demands richer, more comprehensive representations than GRS-based skill classification. The continued dominance of the baseline suggests that its combination of strong spatial features with implicit temporal modeling remains the most balanced approach for capturing the breadth of information encoded in the full OSATS assessment.

\subsection{Task 3: Keypoint tracking}
As this task only exists for the 2025 challenge, there is no comparison to be made with the 2024 challenge. The performance of the teams in Task~3 was generally lower compared to Tasks~1 and 2. This is likely due to the increased complexity of the tracking task, which requires not only accurate detection of surgical tools but also consistent tracking of their movements throughout the video. The best performing team, Jmees, achieved a HOTA score of 0.20, indicating that there is still significant room for improvement in this area.

The lower performance in Task~3 highlights the challenges associated with tracking surgical tools in a dynamic and complex environment. Factors such as occlusions, rapid movements, and varying lighting conditions can all contribute to the difficulty of accurately tracking tools throughout a surgical procedure. Additionally, the need for precise localization of keypoints adds another layer of complexity. This is especially difficult in the annotation process, where occluded or out-of-frame keypoints need to be estimated by the annotators, introducing noise into the ground truth data. In this challenge, the frequent occurrence of occlusions and out-of-frame keypoints likely contributed to the lower performance of the tracking models. Additionally, the HOTA metric may not have been ideally calibrated for the specific challenges of tracking in open surgical videos, which could have further impacted the reported performance scores.

These results also carry important implications for the broader goal of automatic surgical skill assessment. The limited precision of current tracking approaches constrains their applicability for downstream motion-based skill analysis, a finding consistent with the lower performance of tracking-dependent methods in the 2024 Task~1 and Task~2 results. Notably, no team in the 2025 challenge attempted to leverage their learned tracking representations to directly solve Tasks~1 or 2, which may reflect an awareness of these limitations. Improving keypoint tracking accuracy in surgical video therefore remains a critical prerequisite for enabling reliable, process-oriented skill assessment from motion features.

\section{Conclusion}\label{sec:conclusion}
Across two iterations of this challenge, nine teams contributed 17 distinct models, providing a comprehensive benchmark for automated surgical skill assessment from video. Several consistent patterns emerged across both tasks and years. For GRS-based skill classification, the strongest approaches combined generality with careful model tuning: the end-to-end 3D CNN baseline performed robustly in both iterations, while conceptually diverse methods (suture counting, phase-specific motion metrics, spatiotemporal architectures) all achieved competitive results when well-executed. This underscores that training strategy, data curation, and model selection matter at least as much as core methodology, and that methodological sophistication does not guarantee improved performance in low-data regimes. While there is suggestive evidence that product-related information represents a strong predictive signal for GRS-based classification, this does not imply that GRS is reducible to outcome quality alone, as skilled processes and skilled outcomes are typically linked.

For full OSATS prediction, temporal and process-oriented features provide more meaningful benefit, consistent with the multi-dimensional, behavior-anchored nature of OSATS sub-categories. This is evidenced by the improved relative performance of workflow-aware approaches (such as team Perk) in the OSATS context, particularly with sufficient training data. Nevertheless, uniformly poor absolute performance across most teams confirms that predicting the full OSATS table from video remains largely unsolved, though clear improvements with additional expert data suggest that progress is achievable as annotated datasets grow. Keypoint tracking proved similarly challenging due to frequent occlusions and out-of-frame instances, and the limited precision of current approaches directly constrains their applicability for downstream motion-based skill analysis, consistent with the lower performance of tracking-dependent methods across both iterations.

These findings expose a fundamental tension in automated surgical skill assessment: the metrics easiest to predict (GRS-based skill level) may be the least informative for training, while those most valuable for formative feedback (OSATS) remain the most challenging to model. Clinically, this distinction is critical. GRS offers a convenient summary score, but OSATS-level feedback identifying specific behavioral deficits across categories such as tissue handling, instrument use, and flow of operation is what trainees and educators need to guide deliberate practice. Bridging this gap requires not only more powerful models but also substantially larger annotated datasets with fine-grained skill-related labels for individual video segments, targeted feature engineering for individual OSATS sub-categories, and reliable keypoint tracking as a foundation for process-oriented analysis. Realizing the clinical promise of automated skill assessment hinges on advancing precisely the capabilities this challenge has shown to be most difficult.

\section*{Acknowledgments}

\section*{Declarations}
\textbf{Funding:} This project has been funded by the Stiftung Innovation in der Hochschullehre. The authors would like to thank the Federal Ministry of Research, Technology, and Space (BMFTR) for its support as part of the research program Communication Systems “Souverän. Digital. Vernetzt.”. Joint project 6G-life, project identification number: 16KIS2413K. Partially funded by the German Research Foundation (DFG, Deutsche Forschungsgemeinschaft) as part of Germany’s Excellence Strategy – EXC 2050/2 – Project ID 390696704 – Cluster of Excellence “Centre for Tactile Internet with Human-in-the-Loop” (CeTI) of TUD Dresden University of Technology as well as BMBF within the DAAD Konrad Zuse AI school SECAI (project 57616814).

Behrus Hinrichs-Puladi was supported by the Clinician Scientist Program of the Faculty of Medicine RWTH Aachen University.

Funda\c c\~ao para a Ci\^encia e Tecnologia (FCT) Portugal for the grant 2021.05068.BD (Tiago Jesus) and grant 2022.11928.BD (André Ferreira). This work was also supported by FCT within the R\&D Units Project Scope: UIDB/\allowbreak 00319/\allowbreak 2020.

E. Mazomenos and D. Stoyanov receive support by the EPSRC under the “Human-centred Machine Intelligence To Optimise Robotic Surgical Training (HuMIRoS)” (ref: EP/\allowbreak Z534754/\allowbreak 1). D. Stoyanov is also supported by the UK Department of Science, Innovation and Technology (DSIT), the Royal Academy of Engineering under the Chair in Emerging Technologies programme and the EPSRC under the “Optical and Acoustic imaging for Surgical and Interventional Sciences (OASIS)” (ref: UKRI145) grant. 

This study was supported by the NAVER Digital Bio Innovation Research Fund, funded by NAVER Corporation (Grant No. [3720250031]), the Institute of Information \& Communications Technology Planning \& Evaluation (IITP)-Global Data-X Leader HRD program grant funded by the Korea government (MSIT) (IITP-2024-RS-2024-00441407), and a grant of the Korea Health Technology R\&D Project through the Korea Health Industry Development Institute (KHIDI), funded by the Ministry of Health \& Welfare, Republic of Korea (grant number : RS-2025-02307233).

\noindent\textbf{Ethics:} The data used was authorized by the local ethics committee of the University Hospital RWTH Aachen (approval code EK 352/21 and EK 22-329) and was registered, including the study protocol, in the German Clinical Trials Register (DRKS00029307).

\noindent\textbf{AI Usage:} During the preparation of this work the authors used GitHub CoPilot and you.com in order to complete texts, spell and formulation check, and shorten or summarize the manuscript. After using this tool/service, the authors reviewed and edited the content as needed and take full responsibility for the content of the published article.

\noindent\textbf{Conflicts of Interest:} Stefanie Speidel and Danail Stoyanov serve as Associate Editors for the Medical Image Analysis journal.

\noindent\textbf{Data Availability:} \url{https://www.synapse.org/Synapse:syn58905622}

\noindent\textbf{Code Availability:} \href{https://www.synapse.org/Synapse:syn58982223}{Evaluation scripts}, \href{https://github.com/amuck667/TrackEval/tree/devel-kp}{HOTA Metric}, \href{https://github.com/sk-jp/oss2025}{Code Team SK 2025}, \href{URL}{Code Team syangcw}, \href{https://github.com/JmeesInc/OpenSuturingSkillsChallenge.git}{Code Team Jmees 2024}, \href{https://github.com/ShunsukeKikuchi/oss25_sol.git}{Code Team Jmees 2025} \href{https://github.com/trajesus/Suturing_Challenge_Approach}{Code Team Algoritmi 2024}, \href{https://github.com/trajesus/Suturing_Challenge_Approach}{Code Team Algoritmi 2025},
\href{https://github.com/RoiPapo/AIxSuture_Quality_Assessment}{Code Team Scalpel}, \href{https://github.com/RebeccaHisey/OSS_Challenge}{Code Team Perk}, 
\href{https://github.com/Dan-Dan-99/MediSC_OyeSS}{Code Team MediSC\_OyeSS 2025}

\appendix

\section{Dataset Details}
\subsection{2024 Challenge - Task 1 and 2}
\begin{figure*}[h!]
    \centering
    \includegraphics[width=\linewidth,keepaspectratio]{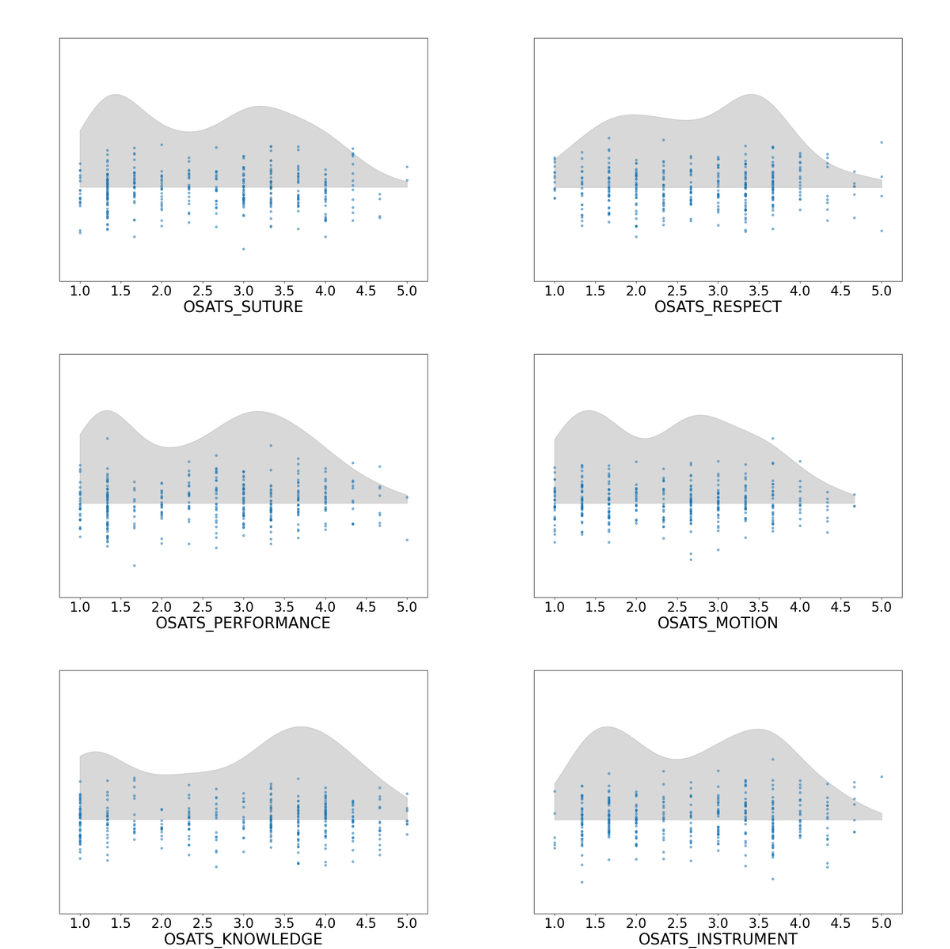}
    \caption{Distributions of the individual OSATS categories of the train set.}
    \label{fig:sup_dist24train}
\end{figure*}

\begin{figure*}[h!]
    \centering
    \includegraphics[width=\linewidth,keepaspectratio]{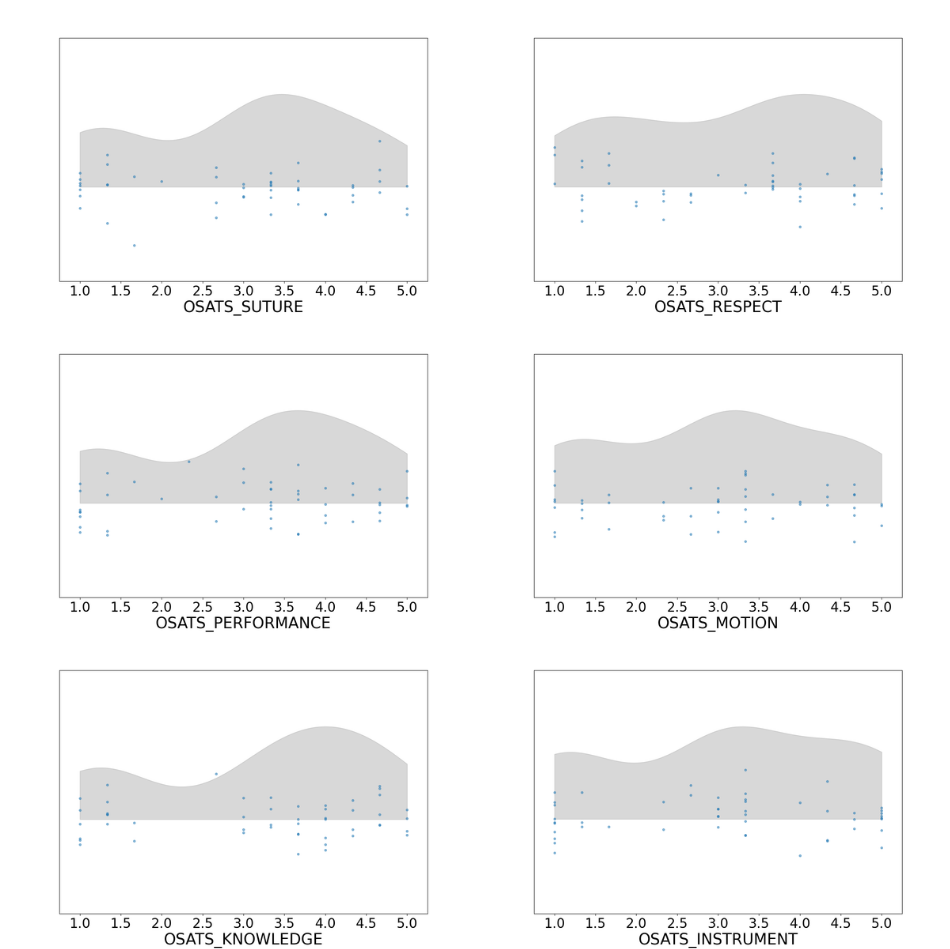}
    \caption{Distributions of the individual OSATS categories of the test set.}
    \label{fig:sup_dist24test}
\end{figure*}

Dataset distributions of the OSATS ratings for the train and test sets are seen in Figures~\ref{fig:sup_dist24train} and~\ref{fig:sup_dist24test}.

\subsection{2025 Challenge - Task 1 and 2}
\begin{figure*}[h!]
    \centering
    \includegraphics[width=\linewidth,keepaspectratio]{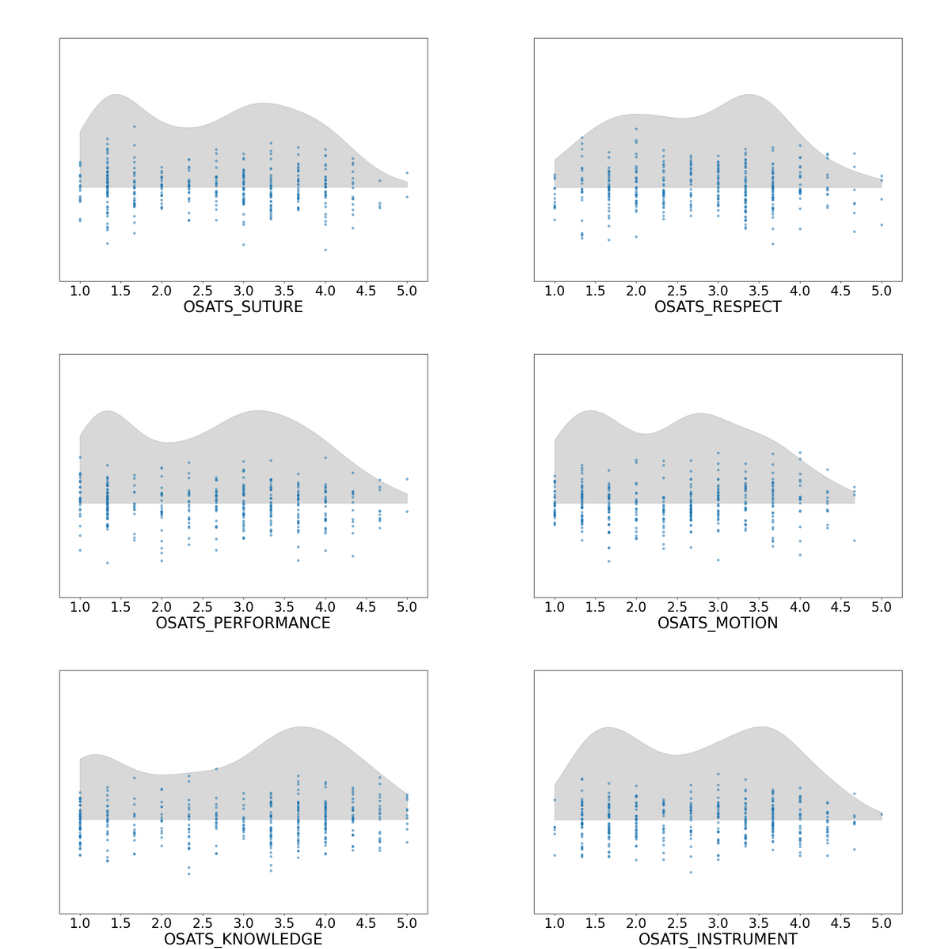}
    \caption{Distributions of the individual OSATS categories of the train set.}
    \label{fig:sup_dist25train}
\end{figure*}

\begin{figure*}[h!]
    \centering
    \includegraphics[width=\linewidth,keepaspectratio]{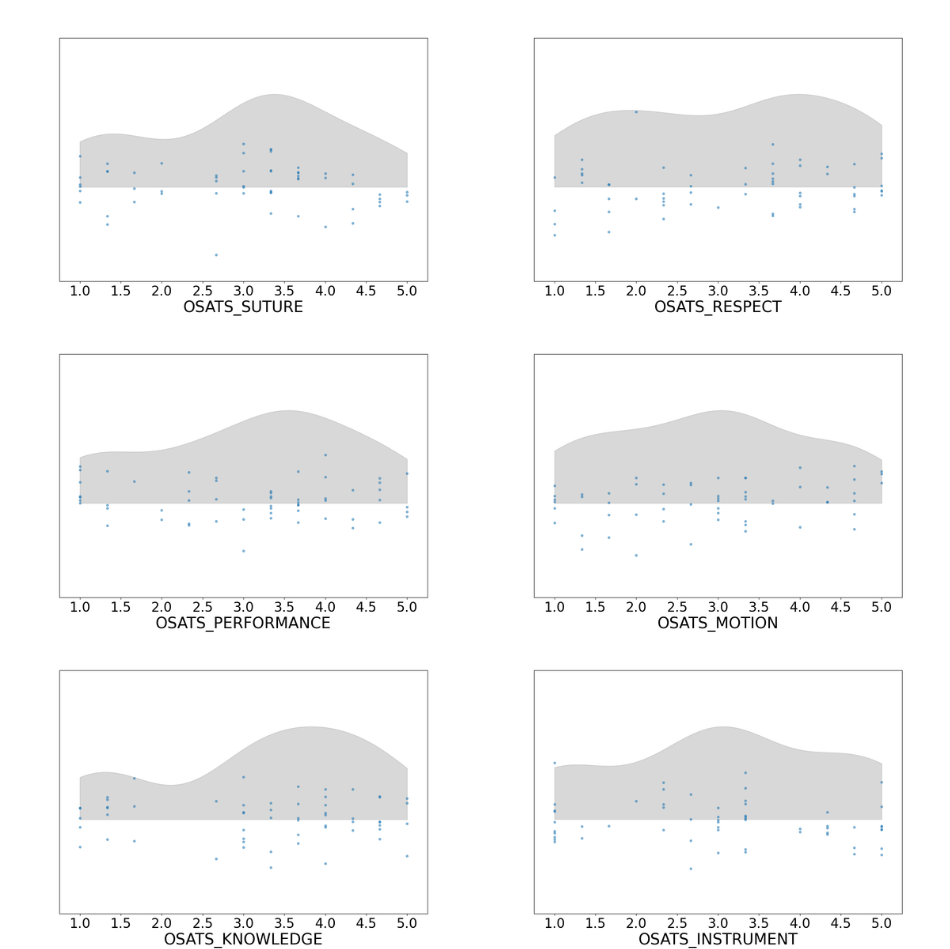}
    \caption{Distributions of the individual OSATS categories of the test set.}
    \label{fig:sup_dist25test}
\end{figure*}

Dataset distributions of the OSATS ratings for the train and test sets are seen in Figures~\ref{fig:sup_dist25train} and~\ref{fig:sup_dist25test}.

\subsection{2025 Challenge - Task 3}
Keypoint annotation distributions for the validation and test set can be found in Tables~\ref{tab:supt3dist_val} and~\ref{tab:supt3dist_test}
\begin{table*}[h!]
\centering
\caption{Challenge 2025 validation set keypoint (KP) distributions. The count column indicates the total number of annotated instances of that object. The percentages are calculated as the number of hidden or out-of-frame KPs divided by the total count of KPs.}
\label{tab:supt3dist_val}
\begin{tabular}{lccc}
\toprule
\textbf{Object} & \textbf{No. of Instances} & \textbf{Hidden KP (\%)} & \textbf{Out-of-Frame KP (\%)} \\
\midrule
Left Hand & 505 & 35.78 & 24.39 \\
Right Hand & 505 & 39.04& 24.72 \\
Scissors & 505 & 3.30 & 16.57 \\
Tweezers & 505 & 21.91 & 0.07 \\
Needle Holder & 505 & 3.89 & 10.03 \\
Needle & 505 & 29.31 & 21.25 \\
\midrule
Total & 3030 & 26.01 & 18.27 \\
\bottomrule
\end{tabular}
\end{table*}

\begin{table*}[h!]
\centering
\caption{Challenge 2025 test set keypoint (KP) distributions. The count column indicates the total number of annotated instances of that object. The percentages are calculated as the number of hidden or out-of-frame KPs divided by the total count of KPs.}
\label{tab:supt3dist_test}
\begin{tabular}{lccc}
\toprule
\textbf{Object} & \textbf{No. of Instances} & \textbf{Hidden KP (\%)} & \textbf{Out-of-Frame KP (\%)} \\
\midrule
Left Hand & 1003 & 30.31 & 27.19 \\
Right Hand & 1003 & 35.33& 33.05 \\
Scissors & 1003 & 9.64 & 15.25 \\
Tweezers & 1003 & 17.05 & 12.50 \\
Needle Holder & 1003 & 2.59 & 8.47 \\
Needle & 1066 & 31.86 & 14.72 \\
\midrule
Total & 6081 & 24.11 & 21.38 \\
\bottomrule
\end{tabular}
\end{table*}

\section{Inter Rater Agreement}
Details of the inter rater agreement analysis for the 2024-2025 train and test sets are seen in Table~\ref{tab:ira}
\begin{table*}[h!]
    \centering
\caption{Inter-rater agreement calculated with ICC(A, k=3). Values in brackets are the 95th percentile confidence intervals.}
\label{tab:ira}
\begin{tabular}{l|ll|ll}
    \toprule
Category & 2024 Train            & 2024 Test             & 2025 Train            & 2025 Test             \\
\midrule
GLOBAL\_RATING\_SCORE & 0.92 {[}0.89, 0.95{]} & 0.94 {[}0.87, 0.99{]} & 0.92 {[}0.89, 0.95{]} & 0.94 {[}0.89, 0.97{]} \\
OSATS\_RESPECT        & 0.79 {[}0.74, 0.83{]} & 0.88 {[}0.76, 0.96{]} & 0.79 {[}0.74, 0.83{]} & 0.88 {[}0.79, 0.93{]} \\
OSATS\_MOTION         & 0.85 {[}0.81, 0.89{]} & 0.91 {[}0.83, 0.97{]} & 0.84 {[}0.8, 0.88{]}  & 0.91 {[}0.84, 0.95{]} \\
OSATS\_INSTRUMENT     & 0.80 {[}0.71, 0.9{]}  & 0.92 {[}0.86, 0.97{]} & 0.80 {[}0.73, 0.88{]} & 0.91 {[}0.85, 0.95{]} \\
OSATS\_SUTURE         & 0.87 {[}0.83, 0.9{]}  & 0.92 {[}0.84, 0.97{]} & 0.85 {[}0.79, 0.9{]}  & 0.91 {[}0.85, 0.95{]} \\
OSATS\_FLOW           & 0.84 {[}0.8, 0.88{]}  & 0.90 {[}0.81, 0.95{]} & 0.83 {[}0.79, 0.88{]} & 0.88 {[}0.79, 0.93{]} \\
OSATS\_KNOWLEDGE      & 0.91 {[}0.86, 0.95{]} & 0.90 {[}0.81, 0.96{]} & 0.90 {[}0.87, 0.94{]} & 0.89 {[}0.82, 0.94{]} \\
OSATS\_PERFORMANCE    & 0.90 {[}0.86, 0.93{]} & 0.93 {[}0.87, 0.97{]} & 0.89 {[}0.85, 0.93{]} & 0.92 {[}0.87, 0.96{]} \\
OSATS\_FINAL\_QUALITY & 0.89 {[}0.86, 0.93{]} & 0.92 {[}0.84, 0.98{]} & 0.88 {[}0.84, 0.93{]} & 0.93 {[}0.87, 0.96{]} \\
\bottomrule
\end{tabular}
\end{table*}

\section{2025 Challenge Task 3 Annotation Protocol}
The annotation protocol to Task 3 can be found in the supplementary file S1.

\section{Participant Method Details}
Details to participant methods are included in Table~\ref{tab:sup_methods}

\begin{table*}[h!]
\centering
\caption{Overview of all submitted methods across both challenge iterations. Dashes indicate no information was provided.}
\label{tab:sup_methods}
\resizebox{\textwidth}{!}{%
\begin{tabular}{llcp{2cm}p{3cm}p{2.5cm}p{2cm}p{1.5cm}p{1cm}p{1.5cm}}
\toprule
\textbf{Team} & \textbf{Year} & \textbf{Task(s)} & \textbf{Backbone} & \textbf{Method} & \textbf{Frame Sampling} & \textbf{Loss} & \textbf{Optimizer} & \textbf{LR} & \textbf{Batch Size} \\
\midrule
Baseline & 2024 & 1, 2 & 3D CNN\newline (Kinetics-400) & End-to-end video\newline classification & 5 fps,\newline patch sampling & Huber & Adam & 5e-5 & 1 video \\
Baseline & 2025 & 1, 2 & 3D CNN\newline (Kinetics-400) & End-to-end video\newline classification & 5 fps,\newline patch sampling & Huber & Adam & 5e-5 & 1 video \\
\midrule
SK & 2024 & 1, 2 & VideoMAE\newline (Kinetics-400) & Final-frame suture counting $\rightarrow$ temporal pooling $\rightarrow$ MLP & Final frames & CE + Regression & -- & 1e-5 with\newline cosine annealing & 18 frames \\
SK & 2025 & 1, 2 & VideoMAE\newline (Kinetics-400) & Early-video clips $\rightarrow$ temporal pooling $\rightarrow$ MLP & Beginning\newline of video & Softmax-based\newline regression & -- & 1e-4 with\newline cosine annealing & 16 frames (MAE),\newline 30 frames (MLP) \\
\midrule
Perk & 2024 & 1, 2 & ResNet50 + YOLOv8\newline (self-annotated data) & Phase-specific motion metrics + workflow features + dedicated tracking & 10 fps & Weighted categorical\newline cross-entropy & Adam & 1e-7 to 1e-5 with\newline cosine annealing & 32 frames \\
\midrule
Scalpel & 2024 & 1, 2 & YOLO\newline (self-annotated data) & Stitch/movement descriptors + aggregated motion metrics & 10 most recent frames + whole video & Cross entropy\newline + MSE & Grid search & -- & -- \\
\midrule
Jmees & 2024 & 1, 2 & Mask2Former\newline + 1DCNN & Tracking + auxiliary\newline proxy prediction & Patch sampling & MAE + BCE\newline with Logits & AdamW & 1e-7 to 5e-5 with\newline cosine annealing & 12 \\
Jmees & 2025 & 1, 2 & Video\newline Transformer & Video classification with\newline auxiliary segmentation & 10 fps,\newline even sampling & L1 + class-weighted Dice\newline + cross entropy & Schedule-free\newline /AdamW & 1e-4 & 1x96 frames \\
Jmees & 2025 & 3 & ConvNeXt & RGB + Flow + masks,\newline heatmaps, Temporal\newline flow warping & All & Heatmap: BCE with Logits,\newline visibility: cross entropy,\newline keypoints: L1 & -- & 1e-4 & 8 \\
\midrule
Syangcw & 2024 & 1, 2 & Transformer & End-to-end video\newline classification & 7.5 fps,\newline random patches & Binary cross entropy & AdamW & 5e-4 with\newline cosine annealing & 4x48 frames \\
\midrule
Algoritmi & 2024 & 1 & YOLOv5\newline (Sign Language) & Tool/hand tracking & Full video & MSE & AdamW & 5e-5 with\newline ReduceLROnPlateau & 10 frames \\
Algoritmi & 2025 & 1 & InceptionV3\newline (ImageNet) & RGB + edge features,\newline end-of-video focus & End of video,\newline 128 frames & Weighted\newline cross entropy & AdamW & 1e-4 with\newline cosine annealing & 2 \\
Algoritmi & 2025 & 2 & YOLOv5 + ResNet\newline (ImageNet) + LSTM & ROI cropping $\rightarrow$\newline RGB $\rightarrow$ LSTM & End of video,\newline 128 frames & Cross entropy & Adam & 1e-4 & 2 \\
Algoritmi & 2025 & 3 & SAM2 + XGBoost\newline + UNet & Segmentation $\rightarrow$\newline classification $\rightarrow$\newline keypoints & All & MSE & -- & Base: 5e-6,\newline vision: 3e-6 & 2 \\
\bottomrule
\end{tabular}%
}
\end{table*}

\begin{table*}[h!]
\centering
\ContinuedFloat
\caption[]{Overview of all submitted methods (continued).}
\resizebox{\textwidth}{!}{%
\begin{tabular}{llcp{2cm}p{3cm}p{2.5cm}p{2cm}p{1.5cm}p{1cm}p{1.5cm}}
\toprule
\textbf{Team} & \textbf{Year} & \textbf{Task(s)} & \textbf{Backbone} & \textbf{Method} & \textbf{Frame Sampling} & \textbf{Loss} & \textbf{Optimizer} & \textbf{LR} & \textbf{Batch Size} \\
\midrule
Saeid & 2025 & 1 & R(2+1)D-18\newline (Kinetics-400) & GRM-based hyperparameter\newline optimization & 16 frames\newline evenly sampled & Cross entropy & AdamW & 1e-5 with\newline ReduceLROnPlateau & 4 \\
Saeid & 2025 & 2 & R(2+1)D-18\newline (Kinetics-400) & GRM-based hyperparameter\newline optimization & 16 frames\newline evenly sampled & MSE & AdamW & 1e-5 with\newline ReduceLROnPlateau & 4 \\
\midrule
MediSC & 2025 & 1 & ConvNext & Spatial feature\newline extraction & 16 frames at\newline end of video & Cross entropy & Adam & 1e-5 & 16 \\
MediSC & 2025 & 2 & Video\newline Transformer & Spatial feature\newline extraction & 16 frames at\newline end of video & Cross entropy & Adam & 1e-6, 1e-5,\newline 1e-4 & 16 \\
MediSC & 2025 & 3 & RTMDet + CSPNeXt\newline (ImageNet + COCO) & Detection +\newline pose estimation & All & -- & Adam/ AdamW & detection: 5e-4 with\newline MultiStepLR, pose:\newline 4e-3 with cosine annealing & 16 \\
\midrule
Mori & 2025 & 3 & Mask R-CNN (ResNet-50\newline + feature pyramid)\newline + RTMPose-M & Segmentation $\rightarrow$ hand keypoints $\rightarrow$ instrument landmarks $\rightarrow$ OKS-IoU tracking & All & -- & AdamW & 5e-3 with\newline cosine annealing & -- \\
\bottomrule
\end{tabular}%
}
\end{table*}

\section{Additional Results}
\subsection{2024 Challenge - Task 1}
\begin{figure*}[h!]
    \centering

    \begin{subfigure}[t]{0.33\textwidth}
        \centering
        \includegraphics[width=\linewidth]{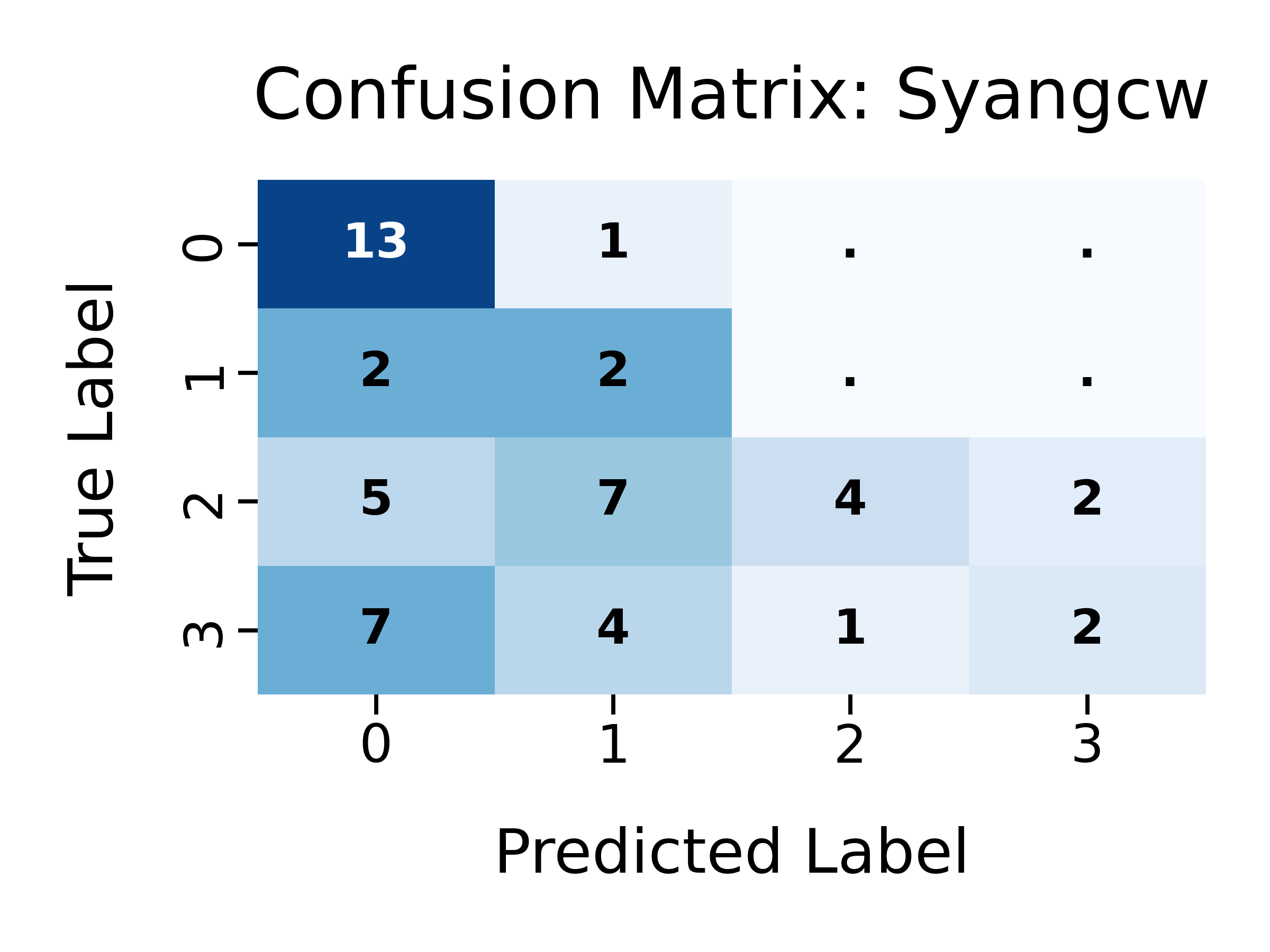}
    \end{subfigure}\hfill
    \begin{subfigure}[t]{0.33\textwidth}
        \centering
        \includegraphics[width=\linewidth]{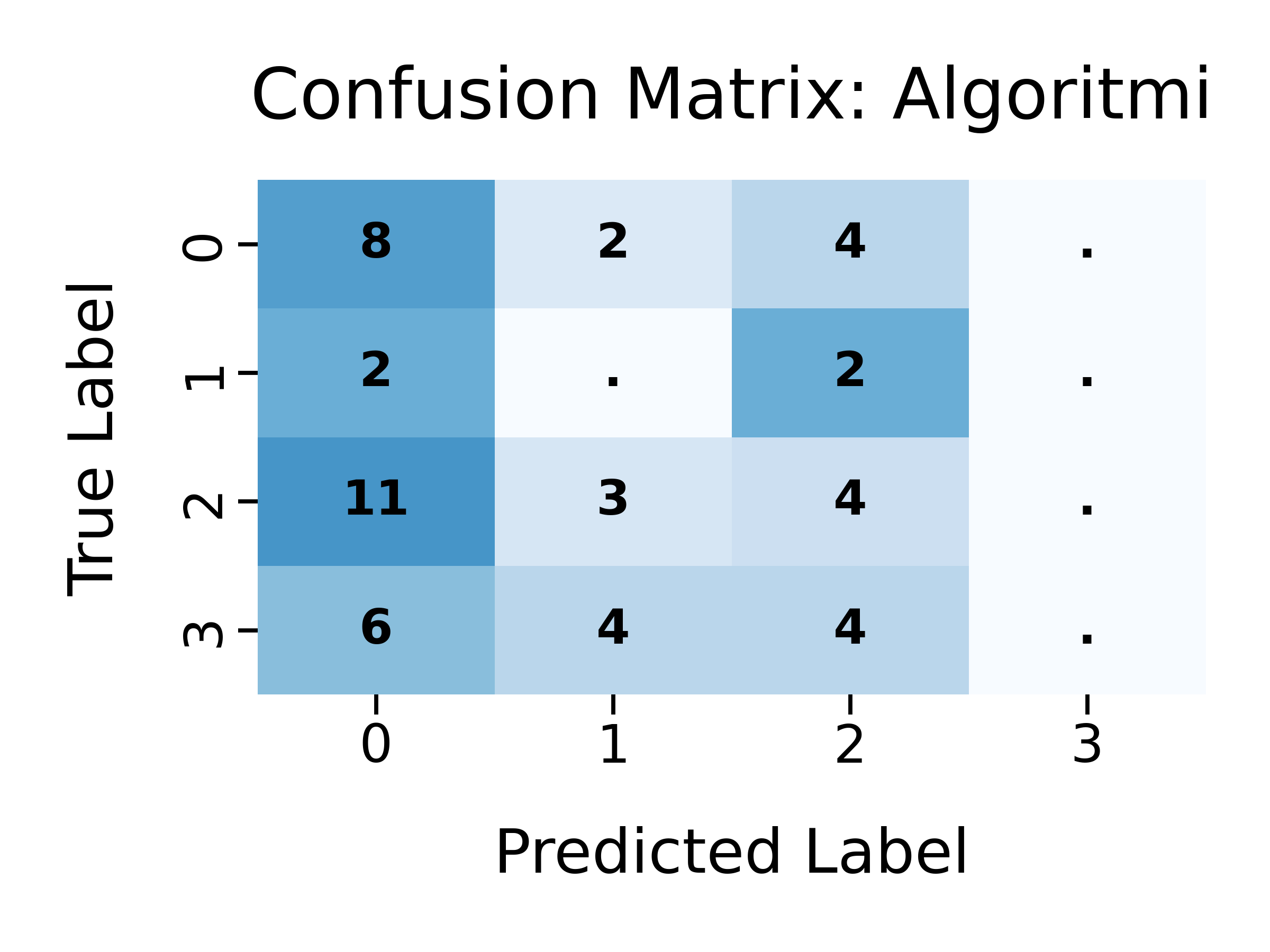}
    \end{subfigure}\hfill
    \begin{subfigure}[t]{0.33\textwidth}
        \centering
        \includegraphics[width=\linewidth]{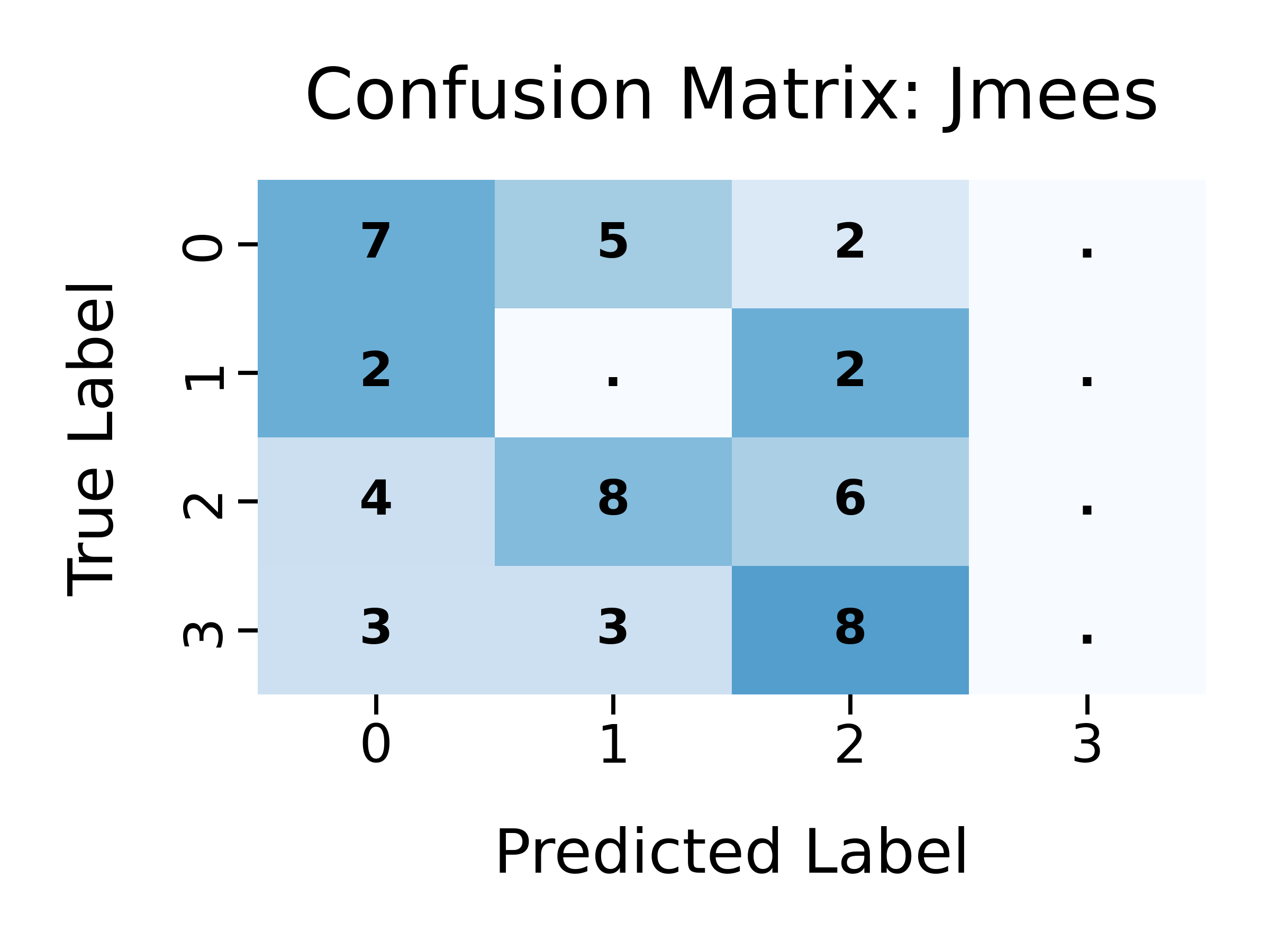}
    \end{subfigure}\hfill
    \begin{subfigure}[t]{0.33\textwidth}
        \centering
        \includegraphics[width=\linewidth]{24Baseline_cfm.png}
    \end{subfigure}

    \vspace{0.6em}

    \begin{subfigure}[t]{0.33\textwidth}
        \centering
        \includegraphics[width=\linewidth]{24SK_cfm.png}
    \end{subfigure}\hfill
    \begin{subfigure}[t]{0.33\textwidth}
        \centering
        \includegraphics[width=\linewidth]{24Perk_cfm.png}
    \end{subfigure}\hfill
    \begin{subfigure}[t]{0.33\textwidth}
        \centering
        \includegraphics[width=\linewidth]{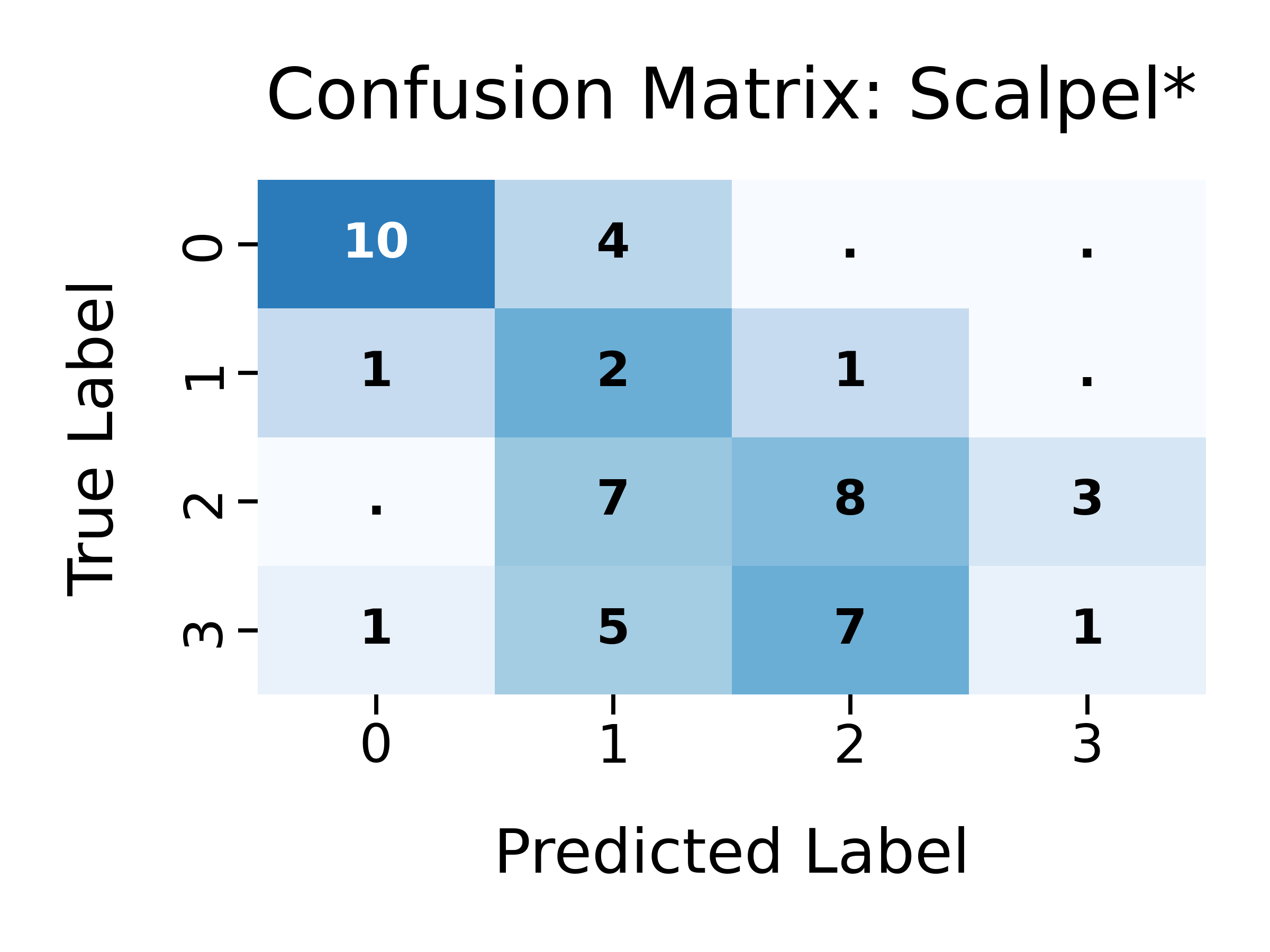}
    \end{subfigure}

    \caption{Task 1 confusion matrices of all teams for the 2024 Challenge.}
    \label{fig:cfm24sup_task1}
\end{figure*}

Confusion matrices for the 2024 Challenge Task 1 are seen in Figure~\ref{fig:cfm24sup_task1}.

\subsection{2025 Challenge - Task 1}
\begin{figure*}[h!]
    \centering

    \begin{subfigure}[t]{0.33\textwidth}
        \centering
        \includegraphics[width=\linewidth]{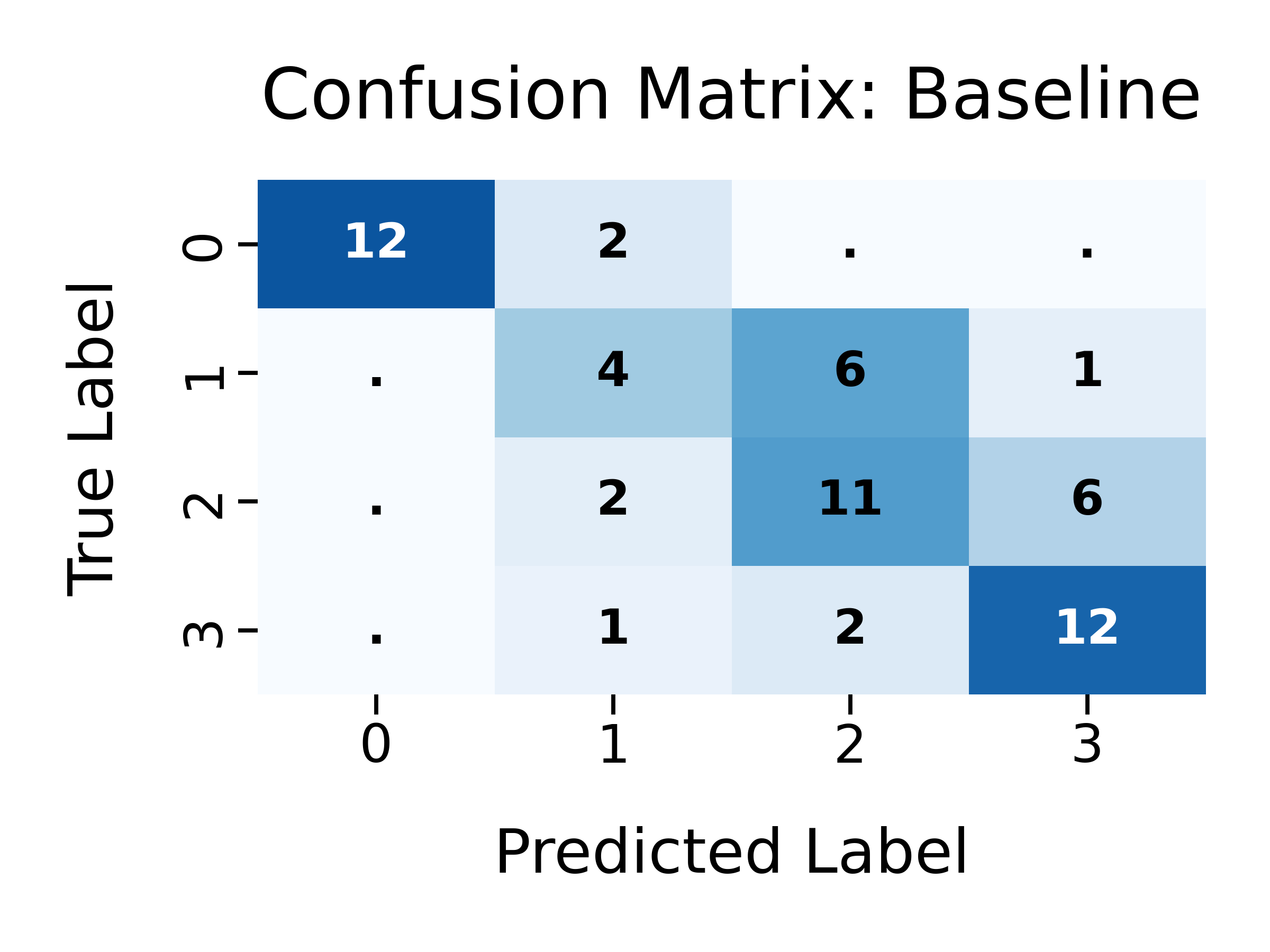}
    \end{subfigure}\hfill
    \begin{subfigure}[t]{0.33\textwidth}
        \centering
        \includegraphics[width=\linewidth]{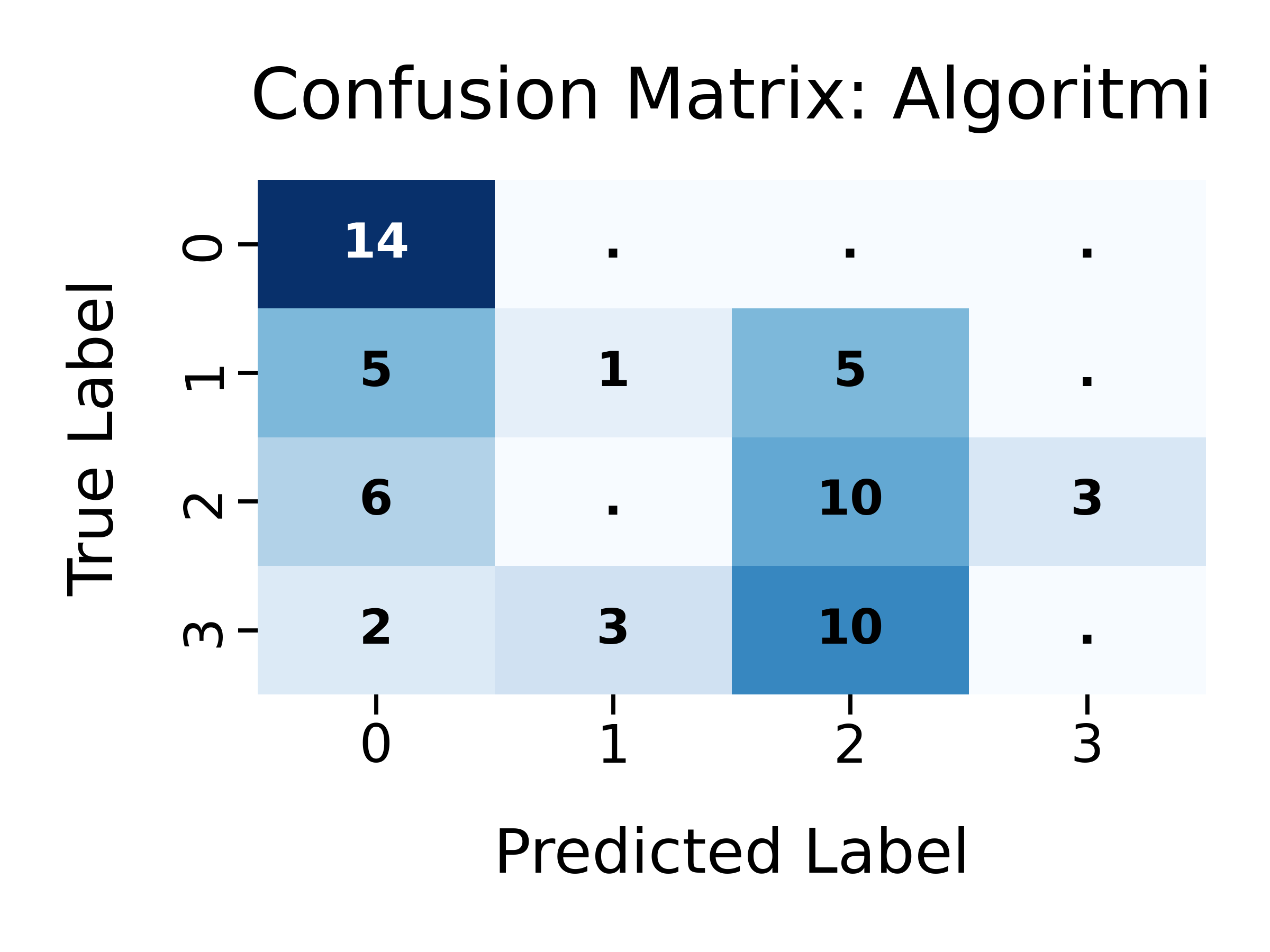}
    \end{subfigure}\hfill
    \begin{subfigure}[t]{0.33\textwidth}
        \centering
        \includegraphics[width=\linewidth]{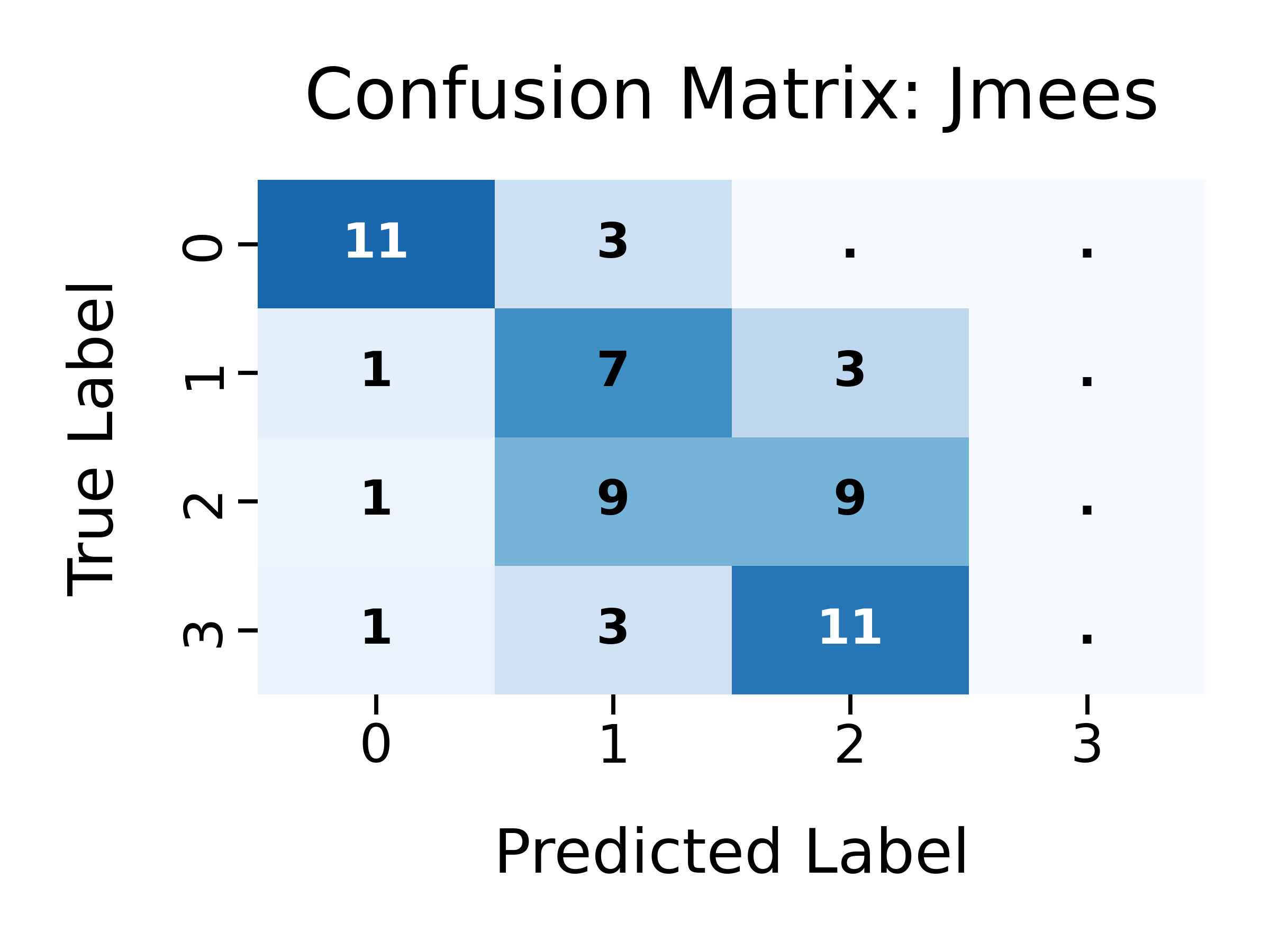}
    \end{subfigure}
    \vspace{0.6em}

    \begin{subfigure}[t]{0.33\textwidth}
        \centering
        \includegraphics[width=\linewidth]{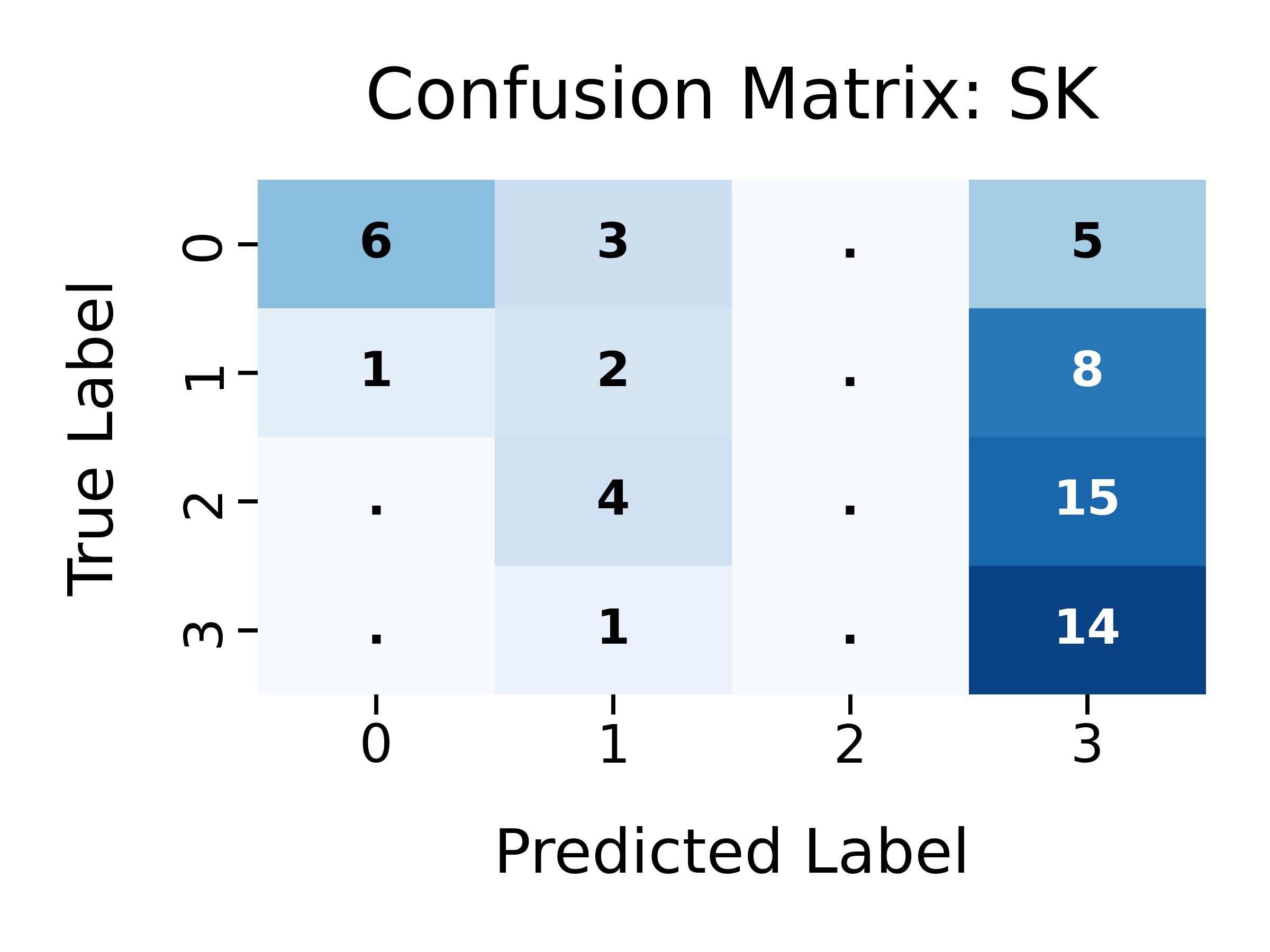}
    \end{subfigure}\hfill
    \begin{subfigure}[t]{0.33\textwidth}
        \centering
        \includegraphics[width=\linewidth]{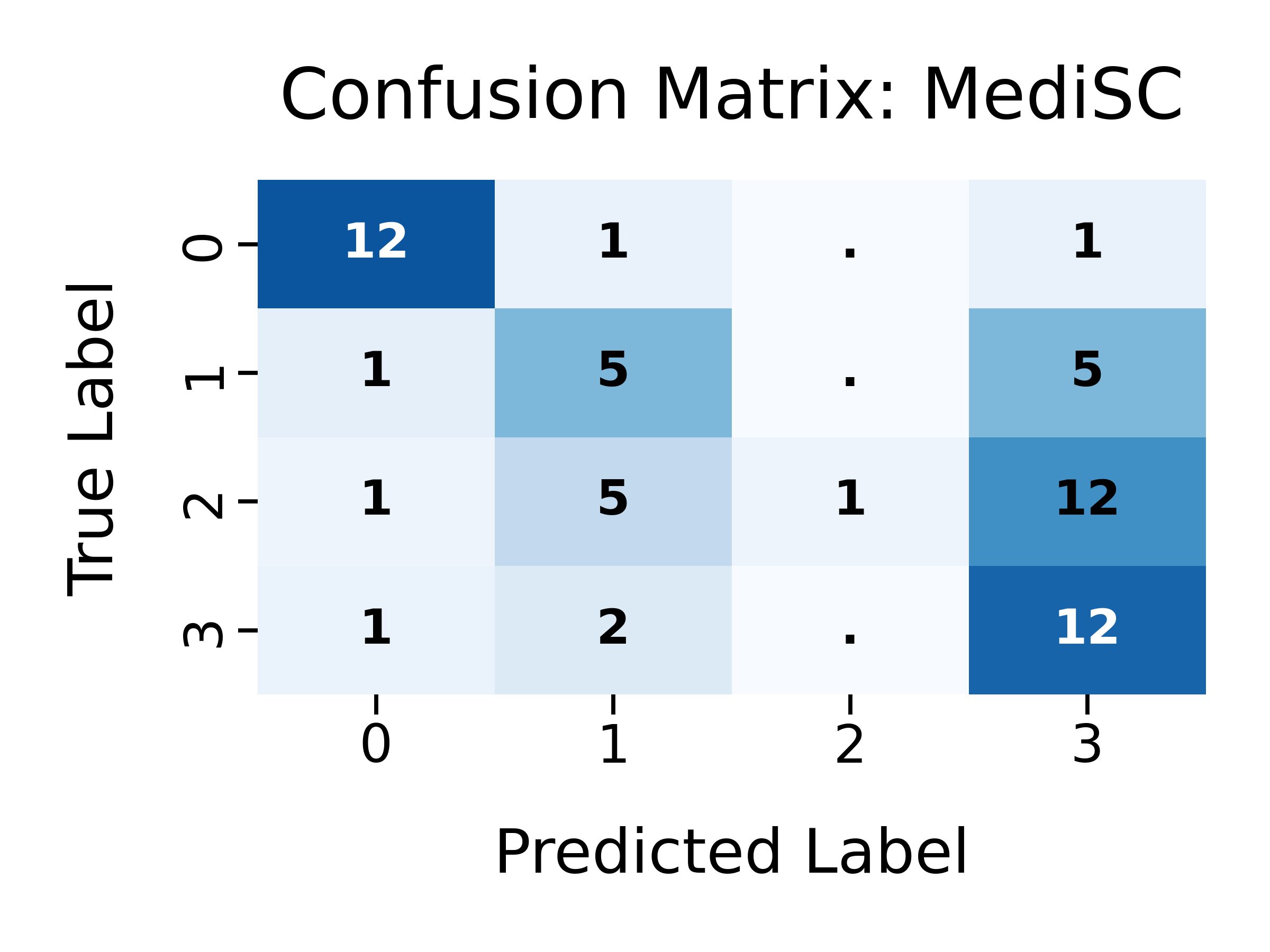}
    \end{subfigure}\hfill
    \begin{subfigure}[t]{0.33\textwidth}
        \centering
        \includegraphics[width=\linewidth]{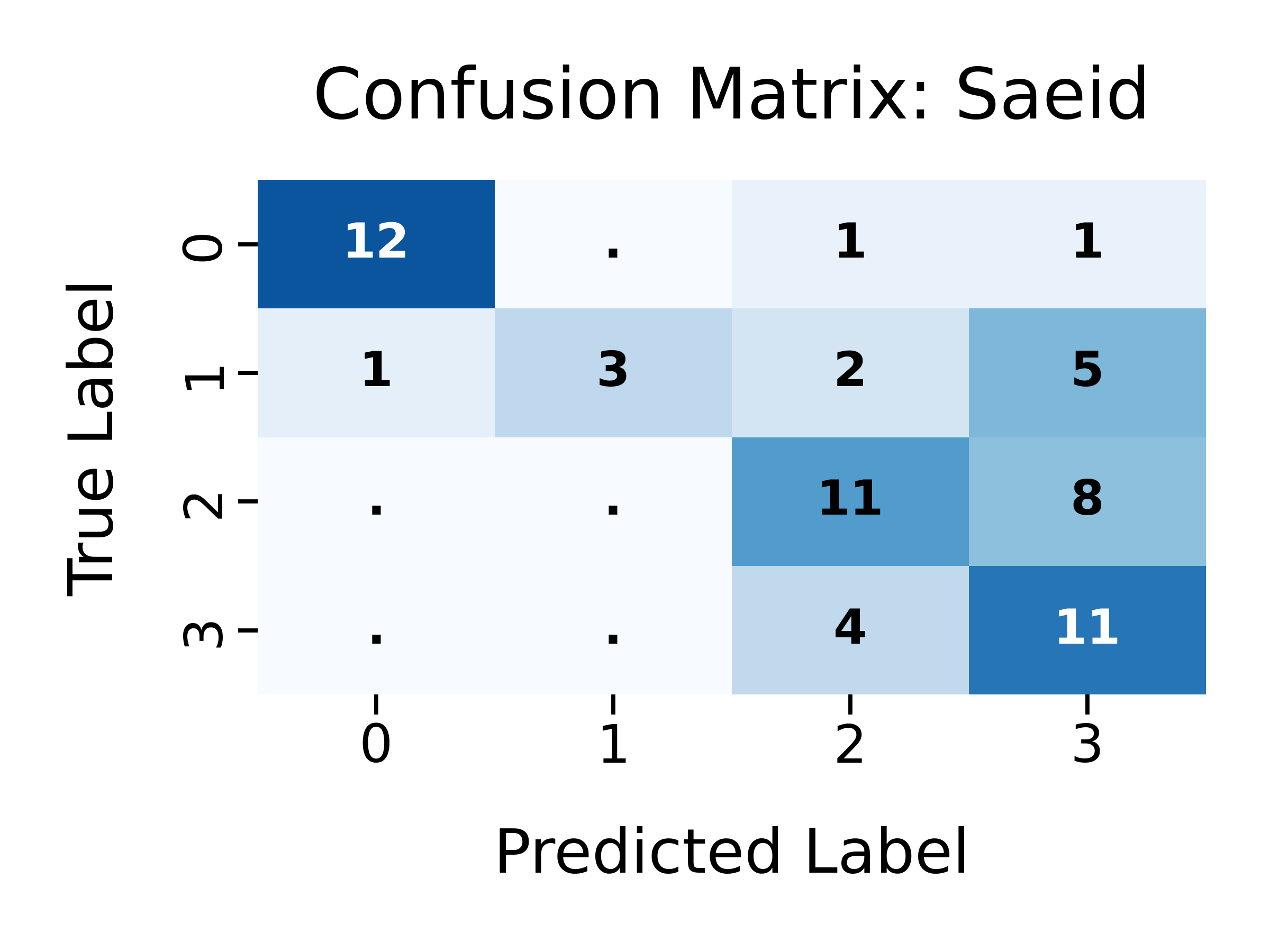}
    \end{subfigure}

    \caption{Task 1 confusion matrices of all teams for the 2025 Challenge.}
    \label{fig:cfm25_task1}
\end{figure*}

Confusion matrices for the 2025 Challenge Task 1 are seen in Figure~\ref{fig:cfm25_task1}.

\printcredits

\bibliographystyle{cas-model2-names}

\bibliography{cas-refs}



\end{document}